\documentclass[Afour,sageh,times]{sagej}

\usepackage{moreverb,url}
\usepackage{amsfonts}
\usepackage{amsmath}
\usepackage{multirow}
\usepackage{todonotes}
\usepackage{bm}
\usepackage{algpseudocode}
\usepackage{algorithm}

\usepackage{subfigure}
\usepackage{empheq}
\usepackage[colorlinks,bookmarksopen,bookmarksnumbered,citecolor=red,urlcolor=red]{hyperref}
\usepackage{siunitx}

\usepackage[acronym,shortcuts]{glossaries}
\newacronym{pdf}{PDF}{Probability Density Function}
\newacronym{ttgo}{TTGO}{Tensor Train for Global Optimization}

\usepackage[perpage]{footmisc}

\DeclareMathOperator*{\argmax}{arg\,max}
\DeclareMathOperator*{\argmin}{arg\,min}
\newcommand{\mc}{\mathcal}
\newcommand{\mb}{\mathbb}

\newcommand\BibTeX{{\rmfamily B\kern-.05em \textsc{i\kern-.025em b}\kern-.08em
T\kern-.1667em\lower.7ex\hbox{E}\kern-.125emX}}

\def\volumeyear{2016}
\begin{document}

\title{Tensor Train for Global Optimization Problems in Robotics
}

\author{Suhan Shetty\affilnum{1,2}, Teguh Lembono\affilnum{1,2}, Tobias Loew\affilnum{1,2},  and Sylvain Calinon\affilnum{1,2} \\
\affiliation{\affilnum{1}Idiap Research Institute, Martigny, Switzerland \\
\affilnum{2} \'Ecole Polytechnique Fed\'erale de Lausanne (EPFL),  Switzerland.}
\email{name.surname@idiap.ch}}

\begin{abstract}
The convergence of many numerical optimization techniques is highly dependent on the initial guess given to the solver. To address this issue, we propose a novel approach that utilizes tensor methods to initialize existing optimization solvers near global optima. Our method does not require access to a database of good solutions. We first transform the cost function, which depends on both task parameters and optimization variables, into a probability density function. Unlike existing approaches, the joint probability distribution of the task parameters and optimization variables is approximated using the Tensor Train model, which enables efficient conditioning and sampling. We treat the task parameters as random variables, and for a given task, we generate samples for decision variables from the conditional distribution to initialize the optimization solver. Our method can produce multiple solutions (when they exist) faster than existing methods. We first evaluate the approach on benchmark functions for numerical optimization that are hard to solve using gradient-based optimization solvers with a naive initialization. The results show that the proposed method can generate samples close to global optima and from multiple modes. We then demonstrate the generality and relevance of our framework to robotics by applying it to inverse kinematics with obstacles and motion planning problems with a 7-DoF manipulator.
\end{abstract}

\maketitle
\section{Introduction}
\label{intro}

Numerical optimization has been one of the major tools for solving a variety of robotics problems including inverse kinematics and motion planning, and control. In this framework, the robotics task to be accomplished is formulated as the minimization of a cost function. Although we ideally seek a solution that incurs the least cost (i.e., global optima), any solution which has a cost comparable to the global optima is usually sufficient. It is essential in robotics applications that a feasible solution is found fast. In practice, the optimization problems in robotics involve non-convex cost functions and the existing optimization techniques often can not quickly find a feasible solution.

There are stochastic procedures, often called evolutionary strategies (e.g., CMA-ES~\citep{hansen2003reducing}, Genetic Algorithm~\citep{whitley1994genetic}, Simulated Annealing~\citep{rutenbar1989simulated}), that can find the global optima of non-convex functions. However, such techniques are too slow for most robotics applications. On the other hand, Newton-type optimization techniques are fast in general---a desirable feature for robotics applications. Hence, most of the existing numerical optimization techniques used in robotics are variants of Newton-type optimization techniques. However, such techniques are iterative in nature and require a good initial guess that determines the solution quality and the time required to find a solution.

 \newpage
 
Finding good techniques to initialize a Newton-type optimizer is an ongoing area of research. A common approach is to first build a database of \emph{optimal} solutions in the offline phase for all possible robotics tasks that are of interest in a given application ~\citep{stolle2006policies,mansard2018,lembono2019}. Then, to solve a given task in the online phase, an approximate solution is retrieved from the database to initialize the optimization solver. While this method is simple and widely applicable, building a useful database can be difficult, especially for challenging problems where the solver struggles to find even a feasible solution without a good initialization. Additionally, predicting an initial guess from the database can be problematic when the optimization problem is multimodal, meaning that a task may have multiple solutions.  Standard function approximation tools such as Gaussian Process Regression (GPR)~\citep{rasmusse:book:2006} and Multilayer Perceptron (MLP) will average the different modalities, resulting in a poor initialization. 

While the multimodal issue can potentially be overcome by keeping only one solution mode in the database, it is not ideal. Firstly, having multimodal solutions available can be useful. For example, in many applications, before the solution is executed on the robot, it needs the approval of an expert (or the user). In such scenarios,  multiple solutions for the given task are desirable so that the user has enough options. Secondly, it is often difficult to separate the different modes, making it impossible to keep only one solution mode in the database.

In this article, we propose a novel approach to produce approximate solutions to a given optimization problem that we call Tensor Train for Global Optimization (TTGO). This approach combines several different techniques, mainly: Tensor Train (TT) decomposition for function approximation~\citep{oseledets2011_tt}, sampling from TT model~\citep{dolgov2020_sampling}, and numerical optimization using cross approximation technique~\citep{Sozykin2022_ttopt}. In contrast to the database approach, we first transform the cost function to an unnormalized probability density function and then approximate the density function using TT decomposition~\citep{oseledets2011_tt}, a technique from multilinear algebra. TT models, as shown recently by~\cite{dolgov2020_sampling}, allow fast procedures to generate exact samples from the density model. Furthermore, we extend this approach to generate samples from a conditioned TT model with controlled priority for high-density regions (which in turn correspond to the low-cost regions) that can then be used as approximate solutions. This approach allows us to obtain a richer set of solutions, especially for multimodal problems (namely, problems with multiple solutions). As it does not use any gradient information, it is also less susceptible to getting stuck at local optima. 

We view the cost function as a function of both the \emph{task parameters} (e.g., desired end-effector pose to reach an object) and the \emph{optimization variables} (e.g., valid robot configurations for inverse kinematics or joint angle trajectories for motion planning). Minimizing the cost function can be reformulated as maximizing a probability density function. Existing methods, such as ~\cite{osa2020multimodal,osa2021motion,pignat2020variational}, use Variational Inference to approximate the probability density function by treating the task parameters as constant, making the cost function a function of only the optimization variables. In contrast, TTGO can handle diverse task parameters by approximating the joint probability distribution of both the task parameters and optimization variables. By exploiting the structure between the two, which often exhibit a \emph{low-rank structure}, the TT model can compactly approximate the density function. After training the model, during the online execution, we can quickly  condition it on specific task parameters and use the  samples from the conditioned model as approximate solutions.

Our method does not rely on an external database of good solutions, nor does it require a separate regression model to predict approximate solutions from a database. Instead, we construct our database in an unsupervised manner during the offline phase, using only the definition of the cost function. This database is compactly represented in TT format which allows for the efficient retrieval of approximate solutions during the online phase, in order of milliseconds. In the case of a multimodal problem, the retrieved solutions will come from multiple modes.

In summary, our contributions are as follows:
\begin{itemize}
    \item We propose a principled approach called TTGO (Tensor Train for Global Optimization) to obtain approximate solutions to a given optimization problem with boundary constraints on decision variables. The approximate solutions are close to the global optima and can then be used to initialize local optimization methods, such as gradient-based solvers, for further refinement.
    \item Our approach builds an implicit database in Tensor Train (TT) format for diverse tasks by only using the definition of the cost function in an unsupervised manner, i.e., without requiring any gradient information or another solver. When the underlying optimization problem is multimodal, our approach can find multiple solutions that correspond to a given task.
    \item The approach is first demonstrated on some benchmark optimization functions to show that it can find global optima and multiple solutions robustly. We show the relevance of the approach to robotics problems by applying it to inverse kinematics with obstacles and motion planning problems with a 7-DoF manipulator where the solution was obtained in a few milliseconds.
\end{itemize}

In this article, we introduce the generic version of TTGO, which includes a TT model for diverse task parameters. However, TTGO can also be used to solve a single task, where the TT model represents the probability distribution of only the optimization variables, and the offline training time is significantly reduced compared to the generic version. In these cases, TTGO's computational time and solution quality are comparable to evolutionary techniques like CMA-ES or GA. However, TTGO has the added advantage of generating multiple solutions even for a single task.

Additionally, even though we have outlined the procedure for functions of continuous variables, TTGO is capable of managing both continuous and discrete variables effortlessly. In situations where both types of variables are present, any required fine-tuning can be made solely for the continuous variables. As a result, it can be utilized as a supplement or substitute for mixed integer programming.

Finally, our approach paves the way for the determination of function optima in the TT model, which facilitates the use of the TT model in various robotics applications, such as Learning from Demonstration~\citep{Pignat22IJRR}, as outlined in Section~\ref{discussion}.

The article is organized as follows. In Section~\ref{related_work}, we provide a literature survey on initializing numerical optimization, multimodal optimization, and tensor methods. Section \ref{background} explains the necessary background on Tensor Train modeling that is used in this paper. Then, in Section \ref{methodology}, we describe the TTGO method proposed in this paper. Section \ref{experiments} presents the evaluation of our algorithm. We first test it on benchmark functions for numerical optimization and then apply it to inverse kinematics with obstacles and motion planning problems with manipulators. In Section \ref{discussion}  and \ref{conclusion}, we conclude the paper by discussing how our approach could lead to new ways of solving a variety of problems in robotics. 

\begin{figure}[t]
    \centering
    \subfigure[]{%
        \includegraphics[width=0.5\textwidth]{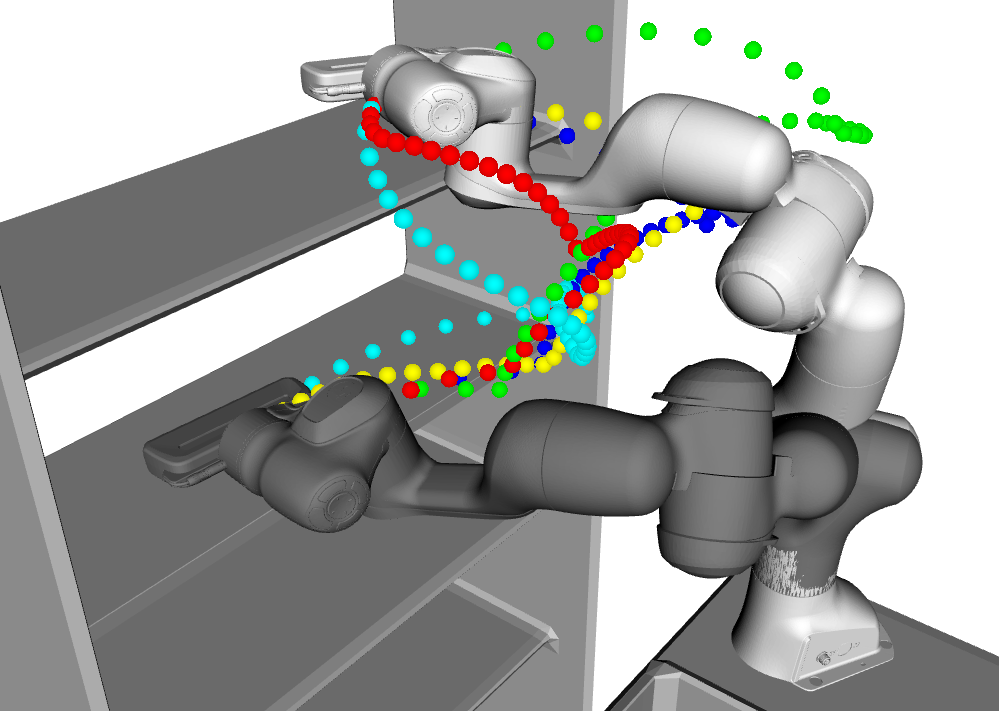}
        }
    \caption{Solutions from TTGO for motion planning of a manipulator from a given initial configuration (white) to a final configuration (dark). The obtained joint angle trajectories result in different paths for the end.effector which are highlighted by dotted curves in different colors. The multimodality is clearly visible from these solutions.}
    \label{fig:panda_no_task}
\end{figure}

\section{Related work}
\label{related_work}
This work intersects with several research directions. Firstly, we target robotics applications that are formulated as optimization problems. Our framework provides a way to predict a good initialization for the optimization solver. At the same time, it also provides a principled way to obtain multiple solutions of a given optimization problem. Finally, the proposed framework relies on tensor methods. We discuss each topic briefly in this section.

\subsection{Optimization in Robotics}
Many problems in robotics are formulated as optimization problems. For example, recent work in motion planning relies on trajectory optimization to plan the robot motion (e.g., CHOMP~\citep{Ratliff2009}, STOMP~\citep{kalakrishnan2011stomp}, TrajOpt~\citep{schulman2014motion}, GPMP~\citep{mukadam2018continuous}). Inverse kinematics for high dimensional robots is usually formulated as nonlinear least squares optimization~\citep{sugihara2011solvability} or Quadratic Programming (QP)~\citep{escande2010fast}. The optimization framework offers a convenient way to transfer the high-level requirement (e.g., energy efficiency, maintaining orientation) to cost functions or constraints. Furthermore, the availability of off-the-shelf optimization solvers and tools for automatic gradient computations allow researchers to focus more on the problem formulation.

However, most of the solvers used in robotics are local optimizers whose performance depend highly on the initialization, especially since most robotics problems are highly non-convex. Even state-of-the-art solvers such as TrajOpt can fail on a simple problem with poor initialization~\citep{lembono2019}. The initialization determines both the convergence speed, the solution quality, and the success rate of the solver. This motivates further research on how to predict good initialization for a given optimization problem.

\subsection{Predicting good initialization}
\label{memory_of_motion}
A majority of works that attempt to predict good initialization rely on a database approach, often called trajectory library~\citep{stolle2006policies} or memory of motion~\citep{mansard2018} for motion planning. In \cite{Hauser2016LearningTP}, they provided a data driven framework for solving globally optimal collision-free inverse kinematics (IK) problems.  The idea is to first build a database of precomputed solutions offline. This database can be constructed from expert demonstrations~\citep{stolle2007transfer}, using the optimization solver itself~\citep{jetchev2009trajectory}, or using the combination of a global planner and the optimization solver~\citep{dantec2021whole}. Once the database is constructed, we can predict a good initial guess (i.e., a \emph{warm start}) for a given task by formulating it as a regression problem that maps the task to the initial guess.  While the formulation is easy to implement, especially since there are many function approximators easily available, the database approach suffers from two main issues: non-convexity and multimodality. 


Firstly, the database approach requires computing good solutions to be stored in the database. With the complexity of general robotics problems, computing the solutions is often not trivial. Assuming we have access to a suitable database and are prepared to train function approximators, we must take into account the fact that many robotics problems have multiple solutions, making it challenging for most function approximators to accurately approximate this one-to-many mapping. Often, such approximators tend to average the different solutions, resulting in poor predictions. 



\subsection{Multimodal Optimization}
\label{survey_mmto}
In problems requiring multiple solutions, common heuristics involve initializing from a uniform distribution. SMTO~\citep{osa2020multimodal} was developed for multimodal trajectory optimization, transforming the task of minimizing a cost function into identifying the modes of a corresponding probability density function (PDF). It approximates the PDF using a Gaussian mixture model (GMM) through Variational Inference. However, SMTO faces practical challenges as it relies heavily on GMM for function approximation.

To address some of the limitations in approximating a PDF with GMM including scalability and the need for samples from the PDF, SMTO employs variational inference and importance sampling from a proposal distribution. The approach is primarily suitable for trajectory optimization problems with finite homotopic solutions. In contrast, LSMO~\citep{osa2021motion} explores infinite homotopic solutions for trajectory optimization using a neural network, albeit with high computational demands for online operations. Our method offers a more versatile solution by distributing computation offline and online, efficiently handling both finite and uncountably many solutions.

\subsection{Tensor Methods}

Tensor factorization techniques (also called Tensor Networks) are extensions of matrix factorization techniques into multidimensional arrays (i.e., tensors). These techniques approximate a given tensor compactly using a set of lower-dimensional arrays (called factors). In addition to the compact representation, they allow efficient algebraic operations to be performed on them. Popular tensor factorization techniques include CP/PARAFAC decomposition, Tucker decomposition, Hierarchical Tucker decomposition, and Tensor Train (TT), see \cite{lathauwer2017_survey,kolda2019_survey} for general surveys, and see \cite{rabanser2017_tensors_survey_ml,cichocki2015_SP} for applications in machine learning and signal processing. Tensor factorization techniques have also been used in robotics to solve control problems that were previously considered to be intractable~\citep{Shetty21_ergodic,horowitz2014_tensor_control,gorodetsky2015_tt}.

Tensor Train (TT) decomposition, also known as Matrix Product States (MPS), provides a good balance between expressive power and efficiency of the representation, and it is equipped with robust algorithms to find the decomposition~\citep{kressner2013_tt_survey}. 
TT decomposition has been used to solve problems involving high-dimensional integration of multivariate functions in~\cite{dolgov2020_integration,Shetty21_ergodic}. \cite{dolgov2020_sampling} used it to approximate probability density functions, with a fast procedure to sample from a probability distribution represented using the TT format. TT decomposition has also been used for data-driven density modeling (or generative modeling) \citep{Han2018UnsupervisedGM_densityEstimation,Stokes2019ProbabilisticMW_densityEstimation,miller2021tensor,Novikov2021_densityEstimation}.

The studies conducted in \cite{zheltkov2019global,Sozykin2022_ttopt} have shown that the TT decomposition can serve as an effective means for gradient-free optimization and can compete with top-performing global optimization algorithms, such as CMA-ES and GA. However, the use of TT decomposition for global optimization in these approaches resembles evolutionary strategies in that they tackle one optimization problem at a time and can only yield one solution, which is too slow for robotics applications.


\section{Background}
\label{background}

In this section, we first describe what a tensor is (Section~\ref{tensors}) and how a multivariate function can be approximated using a tensor (Section~\ref{tensors_as_discrete_analogue}).  We describe how to overcome the curse of dimensionality issues in tensor approximation by relying on TT decomposition that allows efficient computation and storage of the tensor. 
Then, we describe how TT can be used to handle a probability density function (PDF) (Section~\ref{tt_distribution}).

For details about how TT decomposition is analogous to matrix decomposition and how it allows function approximation as separation of variables, we refer to Section~\ref{separation_variables}.

\subsection{Tensors}
\label{tensors}
A tensor is a multidimensional array and as such, it is a higher-dimensional generalization of vectors and matrices. A vector can be considered as a first-order tensor and a matrix as a second-order tensor. The order of a tensor, therefore, refers to the number of dimensions (or modes) of the multidimensional array. 

The shape of a $d$-th order tensor $\bm{\mc{P}} \in\mb{R}^{n_{1}\times\cdots \times n_{d}}$ is defined by a tuple of integers $\bm{n}=(n_1,\ldots,n_d)$. We define the index set $\mc{I}$ of the tensor $\bm{\mc{P}}$  to be a set $\mc{I} = \{\bm{i}=(i_1,\ldots,i_d), i_k \in \{1,\ldots,n_k\}, k \in \{1,\ldots,d\}\}$. This is used to uniquely identify the elements of the tensor. We denote the $\bm{i}$-th element of the tensor $\bm{\mc{P}}$ by $\bm{\mc{P}}_{\bm{i}}$. 

A \emph{fiber} is the higher-order analogue of matrix row and column which is a vector obtained by fixing every index but one. Similarly, a \emph{slice} of a tensor is a matrix obtained by fixing every index but two.

\subsection{Tensors as Discrete Analogue of a Function}
\label{tensors_as_discrete_analogue}

In many applications, tensors arise naturally from the discretization of
multivariate functions defined on a rectangular domain. 
Consider a function $P:\Omega_{\bm{x}} \subset \mb{R}^d \rightarrow \mb{R}$ with a rectangular domain $\Omega_{\bm{x}}= \times_{k=1}^d \Omega_{x_k}$, i.e., a Cartesian product of the intervals of each dimension. Various discretization approaches to specify these intervals can be considered. Unless stated otherwise, we discretize the intervals uniformly. 
We discretize each bounded interval $\Omega_{x_k}  \subset \mb{R}$ into $n_k$ number of elements. $\mc{X} =  \{ \bm{x}=(x^{i_1}_1,\ldots,x^{i_d}_d): x^{i_k}_k \in \Omega_{x_k} ,  i_k \in \{1,\ldots, n_k\} \}$ represents the discretization set and  the corresponding index set is defined as $\mc{I}_{\mc{X}} = \{ \bm{i} = (i_1,\ldots,i_d): i_k \in \{1,\ldots, n_k\}, k \in \{1,\ldots, d\} \}$. We have a canonical bijective discretization map that maps the indices to the tensor elements, i.e,  $X: \mc{I}_{\mc{X}} \rightarrow \mc{X}$ defined as $X(\bm{i}) = (x^{i_1}_1,\ldots, x^{i_d}_d), \forall \bm{i}=(i_1,\ldots,i_d) \in \mc{I}_{\mc{X}}$. With such a discretization, we can obtain a tensor $\bm{\mc{P}}$, a discrete analogue of the function $P$, by evaluating the function at the discretization points given by $\mc{X}$, i.e., $\bm{\mc{P}_i} = P(X(\bm{i})), \bm{i}\in \mc{I}_{\mc{X}}$. To simplify the notation,  we overload the terminology and define $\bm{\mc{P}_x} =\bm{\mc{P}}_{X^{-1}(\bm{x})}, \forall \bm{x} \in \mc{X} $. Note that given a discrete analogue $\bm{\mc{P}}$ of a function $P$, we can approximate the value $P(\bm{x})$ for any $\bm{x} \in \Omega_{\bm{x}}$ by interpolating between certain nodes of the tensor $\bm{\mc{P}}$.

For a high-dimensional function, naively approximating it using a tensor is intractable due to the complexity of both the computation and the storage of the tensor ($\mc{O}(n^{d})$ where $n$ is the maximum number of discretization and $d$ is the dimension of the function and hence the order of the tensor). Tensor factorization solves the storage issue by representing a tensor with its factors that have a smaller number of elements. While many factorization techniques still require the computation of the whole tensor, one particular factorization technique called \emph{cross approximation} allows us to directly compute the factors by using only a function that can evaluate the value of the tensor given an index, hence solving the computation issue.

\subsection{Tensor Train Decomposition}
\label{tt_decomposition}

\begin{figure*}[t!]
    \centering
     \includegraphics[width=0.8\linewidth]{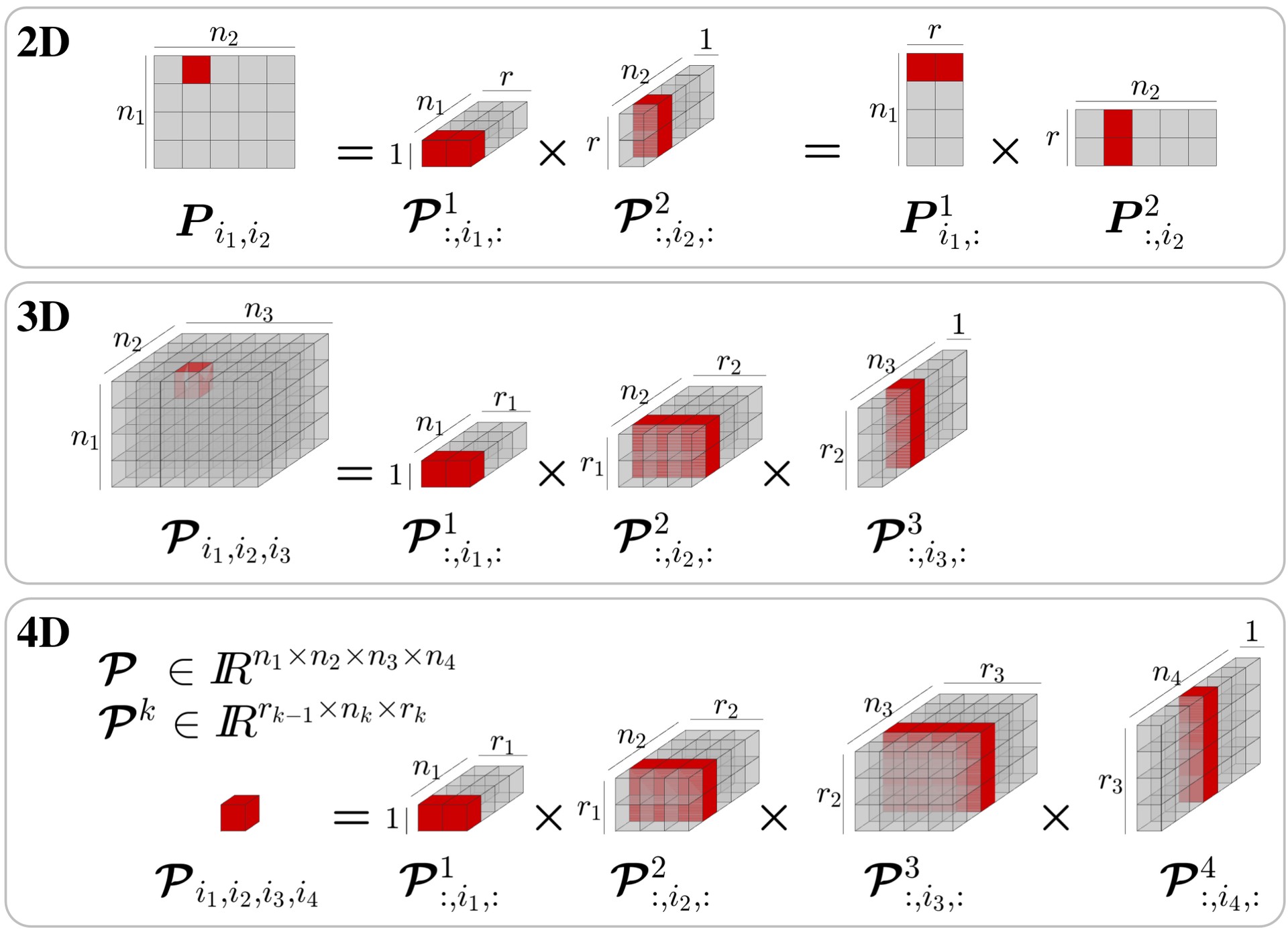}
     \caption{TT decomposition is an extension of matrix decomposition techniques to higher dimensional arrays. With a matrix decomposition, we can access an element of the original matrix by multiplying appropriate rows or columns of the factors. Similarly, an element of a tensor in TT format can be accessed by multiplying the selected slices (matrices represented in red color) of the core tensors (factors). The figure depicts examples for a 2nd order, 3rd order, and a 4th order tensor.}
     \label{fig:tt_format}
\end{figure*}

Similar to matrix factorization described in Section~\ref{separation_variables}, tensor factorization allows us to represent a tensor by its factors. Among the popular factorization techniques, we concentrate in this work on the Tensor Train (TT) decomposition. TT decomposition encodes a given tensor compactly using a set of third-order tensors called \textit{cores}. A $d$-th order tensor $\bm{\mc{P}} \in \mb{R}^{n_1\times\cdots \times n_d}$ in TT format is represented using a tuple of $d$ third-order tensors
$(\bm{\mc{P}}^1,\ldots,\bm{\mc{P}}^d)$. The dimension of the cores are given as $\bm{\mc{P}}^1\in \mb{R}^{1 \times n_1 \times r_{1}}, \bm{\mc{P}}^k\in \mb{R}^{r_{k-1} \times n_k \times r_{k} }$,  $k \in \{2,\ldots,d{-}1\}$,  and
$\bm{\mc{P}}^d\in \mb{R}^{ r_{d-1}  \times n_{d} \times 1}$ with $r_0 = r_d = 1$. As shown in Figure~\ref{fig:tt_format}, the $\bm{i}$-th element of the tensor in this format, with  $\bm{i} \in \mc{I} =\{(i_{1},\ldots,i_{d})\colon i_k \in \{1,\ldots,n_k\}, k \in \{1,\ldots,d\} \}$, is simply given by multiplying matrix slices from the cores:
\begin{equation}
    \label{eq:tt_rep}
    \bm{\mc{P}_i} = \bm{\mc{P}}^1_{:,i_1,:}\bm{\mc{P}}^2_{:,i_2,:}\cdot\cdot\cdot \bm{\mc{P}}^d_{:,i_d,:},
\end{equation}
where $\bm{\mc{P}}^k_{:,i_k,:} \in \mb{R}^{r_{k-1} \times r_k}$ represents the $i_{k}$-th frontal slice (a matrix) of the third-order tensor $\bm{\mc{P}}^k$. The dimensions of the cores are such that the above matrix multiplication yields a scalar.
The \textit{TT-rank} of the tensor in TT representation is then defined as the tuple $\bm{r}=(r_{1},r_{2},\ldots,r_{d-1})$. We call $r$ = $\max{(r_1,\ldots,r_{d-1})}$ as the \textit{maximal rank}.  For any given tensor, there always exists a TT decomposition \eqref{eq:tt_rep} ~\citep{oseledets2011_tt}.

Similarly to the 2D case~\eqref{eq:matrix_separability}, we can also obtain a continuous approximation of the function $P$
as
\begin{equation}
    \label{eq:seperability_matrix}
    P(x_1,\ldots,x_d) \approx \bm{P}^1(x_1) \cdots \bm{P}^d(x_d),
\end{equation}
where $\bm{P}^k(x_k), k \in \{1,\dots,d\}$ is obtained by interpolating each of the core, analogously to the matrix example in~\eqref{eq:matrix_interpolation} (see Appendix~\ref{tensor_core_interpolation_appendix} for more detail). We overload the terminology again to define the continuous TT representation as
\begin{equation*}
    \bm{\mc{P}_x} = P(x_1,\ldots,x_d), \quad \forall \bm{x} \in \Omega_{\bm{x}}.
\end{equation*}

Due to its structure, the TT representation offers several advantages for storage and computation. Let $n = \max(n_{1},\ldots,n_{d})$. Then, the number of elements in the TT representation is $\mc{O}(ndr^{2})$ as compared to $\mc{O}(n^{d})$ elements in the original tensor. For a small $r$ and a large $d$, the representation is thus very efficient. As explained in Section~\ref{separation_variables}, the existence of a low-rank structure (i.e., a small $r$) of a given tensor is closely related to the separability of the underlying multivariate function. Although separability of functions is not a very well understood concept, it is known that smoothness and symmetry of functions often induces better separability of the functions. By \textit{better}, we mean fewer low-dimensional functions in the sum of products representation. The degree of \textit{smoothness} can be formally defined using the properties of higher-order derivatives, however, roughly speaking, it implies the degree of variation of the function across its domain. For example, a probability density function in the form of a Gaussian Mixture Model (GMM) is considered to become less smooth as the number of mixture components (i.e., multi-modality) increases or the variance of the component Gaussians decreases (i.e., sharper peaks). 

\subsection{TT-Cross}
\label{tt_cross}

 The popular methods to find the TT decomposition of a tensor are TT-SVD ~\citep{oseledets2011_tt}, TT-DMRG~\citep{dolgov2020_integration}, and TT-cross~\citep{savostyanov2011_ttcross2}. TT-SVD and TT-DMRG, like matrix SVD, require the full tensor in memory to find the decomposition, and hence they are infeasible for higher-order tensors. TT-cross approximation (TT-cross) is an extension of the cross approximation technique for matrix decomposition explained in Section \ref{matrix_cross_approx} for obtaining the TT decomposition of a tensor. It is appealing for many practical problems as it approximates the given tensor with a controlled accuracy, by evaluating only a small number of its elements and without having to compute and store the entire tensor in the memory. The method needs to compute only certain fibers of the original tensor at a time and hence works in a black-box fashion. We refer the readers to \cite{oseledets2010_ttcross1,savostyanov2011_ttcross2} for more details.
 
Suppose we have a function $P$ and its discrete analogue $\bm{\mc{P}}$ (a tensor). Given the desired accuracy for the approximation $\epsilon$, TT-cross returns an approximate tensor in TT format $\hat{\bm{\mc{P}}} = \text{TT-cross}(P,\epsilon)$ to the tensor $\bm{\mc{P}}$ by querying only a portion of its elements ($\mc{O}(ndr^2)$ evaluations instead of $\mc{O}(n^d)$). The maximal TT-rank $r$ is determined by the algorithm depending on the $\epsilon$ specified. The model is very efficient if the rank $r$ of the tensor is low, which is typically the case in many engineering applications, including robotics.  Thus, TT-cross avoids the need to compute and store explicitly the original tensor, which may not be possible for higher-order tensors. It only requires computing the function $P$ that can return the elements of the tensor $\bm{\mc{P}}$ at various query points, i.e.,  the fibers of the tensor $\bm{\mc{P}}$.

\subsection{TT Distribution} 
\label{tt_distribution}
 Suppose we use the tensor $\bm{\mc{P}}$ in TT format to approximate an unnormalized probability density function $P$ within the discretization set $\mc{X}$ of the domain $\Omega_{\bm{x}}$. We can then construct the corresponding probability distribution that we call TT distribution,
\begin{equation}
    \label{eq:tt_distribution}
    \text{Pr}(\bm{x}) = \frac{|\bm{\mc{P}_x}|}{Z}, \quad \bm{x} \in \Omega_{\bm{x}},
\end{equation}
where $Z$ is the corresponding normalization constant. Alternatively, we could also define the TT distribution to be  $\text{Pr}(\bm{x}) = \frac{\bm{\mc{P}_x}^2}{Z}$. All the techniques, such as conditional sampling and prioritized sampling, used in this paper can also be adapted to this distribution. However, for simplicity of presentation, we do not consider it here.

Due to the separable structure of the TT model, we can get the exact samples from the TT distribution in an efficient manner without requiring to compute the normalization factor $Z$. In Section~\ref{tt_sample_appendix}, we provide details about sampling from the above distribution for the discrete case $\bm{x} \in \mc{\bm{X}}$.

\subsection{Conditional TT Model and Distribution}
\label{tt_condition}

Suppose we want to fix a subset of variables in $\bm{x}$ and find the corresponding conditional distribution of the remaining variables.  
Without loss of generality, let $\bm{x}$ be segmented as $\bm{x}=(\bm{x}_1,\bm{x}_2) \in \Omega_{\bm{x}} = \Omega_{\bm{x}_1} \times \Omega_{\bm{x}_2}$ with $\bm{x}_1 \in \Omega_{\bm{x}_1} \subset \mb{R}^{d_1}$, $\bm{x_2} \in \Omega_{\bm{x}_2} \subset \mb{R}^{d_2}$. i.e., $\bm{x}_1$ corresponds to the first $d_1$ variables in $\bm{x}$.  We are interested in finding the conditional distribution $\text{Pr}(\bm{x}_2 | \bm{x}_1)$ of the TT distribution given in \eqref{eq:tt_distribution}. 

Suppose $\bm{x}_1$ takes a particular value $\bm{x}_t$. We can obtain $\text{Pr}(\bm{x}_2 | \bm{x}_1=\bm{x}_t)$  by defining a conditional TT model $\bm{\mc{P}}^{\bm{x}_1=\bm{x}_t}$ using TT model $\bm{\mc{P}}$ as $$\bm{\mc{P}}^{\bm{x}_t}_{\bm{x}_2} = \bm{\mc{P}}_{(\bm{x}_t,\bm{x}_2)} \forall \bm{x_2} \in \Omega_{\bm{x}_2}.$$

\noindent In other words, the TT cores of $\bm{\mc{P}}^{\bm{x}_1=\bm{x}_t}$ are then given by

\begin{equation}
    \label{eq:conditioned_tt_model}
    {(\bm{\mc{P}}^{\bm{x}_t})}^k =  \begin{cases}
      \bm{\mc{P}}^{k}_{:,\bm{x}_{t_k},:}, & \ k \in \{1,\ldots,d_1 \} \\
      \bm{\mc{P}}^{k}, &  k \in \{d_1+1,\ldots,d_1+d_2\}. \\
    \end{cases}
\end{equation}

\noindent Given the above-defined conditional TT model, we can obtain the conditional distribution as
\begin{equation}
    \label{eq:tt_distribution_cond}
    \text{Pr}(\bm{x}_2|\bm{x}_1=\bm{x}_t) = \frac{|\bm{\mc{P}}_{\bm{x}_2}^{\bm{x}_t}|}{Z_1}, \forall \bm{x}_2 \in \Omega_{\bm{x}_2}.
\end{equation}

\noindent Given  $\bm{x}_1=\bm{x}_t$, we can sample $\bm{x}_2$ from this distribution using Algorithm \ref{alg:tt_cd} with the conditional TT model $\bm{\mc{P}}^{\bm{x}_1=\bm{x}_t}$.
\section{Methodology}
\label{methodology}

\subsection{Problem Definition}

Cost functions in robotics applications typically rely on two types of variables: \emph{task parameters} and \emph{decision variables}. Task parameters are constant for a given optimization problem and describe the range of tasks that may arise in a particular robotic application. For example, in an inverse kinematics (IK) problem with obstacles, the task parameters could be the desired end-effector pose to reach an object, while the decision variables are the variables being optimized (e.g., the robot configuration or joint angles). In most cases, we can anticipate the possible range of task parameters, such as the robot workspace for IK. Ideally, we can solve the optimization problem offline numerous times for the complete range of task parameters and leverage this knowledge to expedite online optimization for new tasks.

It is important to note that in robotics, the cost function is frequently a piecewise smooth function that incorporates a specific structure (i.e., low-rank structure explained in Section \ref{background}) among the cost function variables. For instance, similarities among task parameters correspond to similarities among solutions to the optimization problem.  By capturing this structure, we can compactly model the relationships among variables rather than relying on database approaches that store each data point. Although such a structure is prevalent in many robotics applications, it has not been extensively utilized.

The proposed \gls{ttgo} framework exploits such a structure to gather experience in the offline phase for faster optimization in the online phase. As discussed in Section~\ref{related_work}, the common approach using database and function approximators does not work well when the optimization problems are highly non-convex with many poor local optima and the solutions are multimodal. Our framework provides a principled solution to these two problems. The following section presents the approach in detail.

\subsection{Overview of the Proposed Approach}

Given an optimization problem, TTGO predicts approximate solution(s) that can be refined using an optimization solver. The refinement can be done using standard Newton-type solvers such as SLSQP or L-BFGS-B, so we focus our discussion on the problem of predicting a good approximation of the solution(s). 

Our approach involves first transforming the cost function into an unnormalized Probability Density Function (PDF) and then approximating it with a surrogate probability model, specifically a TT distribution. We regard the cost function as a function of both the optimization variables and the task parameters that define the optimization problem. The surrogate model, therefore, approximates the joint distribution of both the task parameters and optimization variables. During online execution, when a task parameter is specified, we condition the surrogate model on the corresponding parameter. Subsequently, we can sample from this conditional distribution to obtain approximations of the solutions related to the specified task parameters. In cases where the underlying PDF is multimodal, the samples will be derived from multiple modes. These samples represent viable candidates for the solutions. By evaluating the corresponding cost functions, we can then choose the best sample(s) and select the sample(s) with the lowest cost if multiple solutions are needed. In the second stage, we refine these proposed optima using an appropriate optimization technique, such as Newton-type solvers if the objective function is differentiable.

The feasibility of such an approach depends on the properties of the surrogate probability model, namely:
\begin{itemize}
    \item The surrogate probability model should be able to approximate a wide range of probability density functions we encounter in robotics by using only the cost function definition.
    \item Conditioning and sampling from the surrogate probability model determine the speed of the online execution and hence it should be fast.
\end{itemize}

The first requirement comes from the fact that we do not usually have access to the samples from the true probability distribution; we only have the definition of the density function (corresponding to the cost function) that can return the value of the function at a query point.

In our approach, we propose to use the TT distribution (Section \ref{tt_distribution}) as the surrogate model that satisfies the above requirements. The TT model defining the TT distribution corresponds to the discrete analogue of the given unnormalized PDF, and it can be obtained efficiently using TT-Cross algorithm (Section~\ref{tt_cross}). The efficiency is in terms of the number of evaluations of the target function to be modeled, the memory requirement, and the speed of computation. The resultant TT distribution allows fast sampling procedures (see Section~\ref{tt_sample_appendix}). Since we use the samples from the TT distribution as the solution candidates, we are often mainly interested in samples from the high-density regions (i.e., the low-cost regions). This can be accomplished using the prioritized sampling procedures for TT distribution (see Section~\ref{prioritized_sampling}).

In the following section, we provide the mathematical formulation of the approach. 

\subsection{Mathematical Formulation}

Let $\bm{x}_1 \in \Omega_{\bm{x}_1}$ be the task parameter, $\bm{x}_2 \in \Omega_{\bm{x}_2}$ be the decision variables and $\bm{x}=(\bm{x}_1,\bm{x}_2)$. Let $C(\bm{x}_1,\bm{x}_2)$ be a nonegative cost function. Given the task parameter $\bm{x}_1=\bm{x}_t$, we consider the continuous optimization problem in which we want to minimize $C(\bm{x}_t, \bm{x}_2)$ w.r.t $\bm{x}_2$:
\begin{equation}
    \label{eq:opt_0}
    \begin{aligned}
        \bm{x}_2^* =& \argmin_{\bm{x}_2} C(\bm{x}_1,\bm{x}_2)  \\
        \textrm{s.t. } & \bm{x}_1=\bm{x}_t, \\  & \bm{x}_2 \in \Omega_{\bm{x}_2}. \\
    \end{aligned}
\end{equation}

 We assume that $\Omega_{\bm{x}_1} \in \mb{R}^{d_1}$, $\Omega_{\bm{x}_2} \in \mb{R}^{d_2}$ are both rectangular domain and let  $\Omega_{\bm{x}} = \Omega_{\bm{x}_1} \times \Omega_{\bm{x}_2} \subset \mb{R}^{d}$ with $d=d_1+d_2$. TTGO decomposes the procedure to solve such an optimization problem into two steps: 
\begin{enumerate}
    \item Predict an approximate solution $\bm{\hat{x}}_2^*$ that corresponds to the given $\bm{x}_1=\bm{x}_t$, then
    \item Improve the solution $\bm{\hat{x}}_2^*$ using a local search (e.g., Newton type optimization) to obtain the optimal solution $\bm{x}_2^*$.
\end{enumerate} 

To find the approximate solution(s) $\bm{\hat{x}}_2^*$, we first convert the above optimization problem of minimizing a cost function into maximizing an unnormalized probability density function $P(\bm{x}_1, \bm{x}_2)$ using a monotonically non-increasing transformation,

\begin{equation}
\label{eq:opt_1}
\begin{aligned}
    \bm{x_2^*}  =& \argmax_{\bm{x}_2} P(\bm{x}_t,\bm{x}_2)  \\
    \textrm{s.t. } & \bm{x}_1=\bm{x}_t,\\ & \bm{x}_2 \in \Omega_{\bm{x}_2}. \\   
\end{aligned}
\end{equation}

\noindent For example, we can define $P(\bm{x}) = e^{-\beta C(\bm{x})^2}$ with $\beta>0$. Without loss of generality, in the remainder of the paper we consider optimization problems to be of type~\eqref{eq:opt_1} with the objective function being the density function.

In this probabilistic view, the solution $\bm{x}_2^*$ corresponds to the mode, i.e., the point with the highest density, of the conditional distribution $P(\bm{x}_2|\bm{x}_1=\bm{x}_t)$. In general, however, we do not have an analytical formula of  $P(\bm{x}_2|\bm{x}_1=\bm{x}_t)$, and finding the mode is as difficult as solving the optimization problem in~\eqref{eq:opt_0}. TTGO overcomes this issue by first approximating the unnormalized PDF $P(\bm{x}_1,\bm{x}_2)$ using a TT model as the surrogate model to obtain the joint distribution $\text{Pr}(\bm{x}_1,\bm{x}_2)$. Given the task $\bm{x}_1=\bm{x}_t$, we condition the TT model to obtain the conditional distribution $\text{Pr}(\bm{x}_2|\bm{x}_1=\bm{x}_t)$. Finally, the TT model allows us to sample easily from its distribution to produce the approximate solution(s) $\bm{\hat{x}}_2^*$.

\subsubsection{Approximating the unnormalized PDF using TT model:}

As described in Section~\ref{background}, a TT model can approximate a multivariate function as its discrete analogue, i.e., by discretizing the function on a rectangular domain and storing the value in a tensor. For a high-dimensional function, however, it is intractable to construct and store the whole tensor. To avoid the curse of dimensionality, we rely on TT decomposition that allows us to store the tensor in a very compact form as TT cores. We use the TT-cross algorithm that allows us to compute the TT cores without having to construct the whole tensor, reducing the complexity of both the storage and the computation significantly. 


Given the unnormalized PDF $P(\bm{x}_1,\bm{x}_2)$, TTGO uses the TT-Cross algorithm (see Section \ref{tt_cross}) to compute its discrete analogue approximation, i.e., $\bm{\mc{P}}$, in the TT format. The construction of $\bm{\mc{P}}$ only requires the computation of $P(\bm{x}_1,\bm{x}_2)$ at selected points $(\bm{x}_1,\bm{x}_2)$ in the rectangular domain. Instead of computing every single possible value of $P$ in the discretized domain ($\mc{O}(n^d)$), the TT-Cross algorithm only requires $\mc{O}(ndr^2)$ cost function evaluations, where $n$ is the maximum number of discretization and $r$ is the maximum rank of the approximate tensor. More details on how to approximate the function using the TT decomposition are described in Section~\ref{background}. 

The tensor model $\bm{\mc{P}}$ is an approximation of the unnormalized PDF. We can construct the corresponding normalized TT distribution $\text{Pr}(\bm{x})$ with~\eqref{eq:tt_distribution}, which requires the normalization constant as per the definition. However, as described in Section~\ref{tt_sample_appendix}, we can sample from the TT distribution  without calculating the normalization constant. Hence, in practice we can generate the samples by working directly with the unnormalized density $\bm{\mc{P}}$.

\subsubsection{Conditioning TT Model:}

After approximating the joint distribution, we can condition it on the given task. Given the task parameter  $\bm{x}_1 =\bm{x}_t \in \Omega_{\bm{x}_1}$, we first condition the TT model $\bm{\mc{P}}$ to obtain $\bm{\mc{P}}^{\bm{x}_t}$. We then use it to construct the conditional TT distribution $\text{Pr}(\bm{x}_2|{\bm{x}_1=\bm{x}_t})$ as described in Section \ref{tt_condition}. This is the desired surrogate probability model for $P(\bm{x}_2|\bm{x}_1=\bm{x}_t)$.

\subsubsection{Prioritized Sampling}
\label{prioritized_sampling}
In Section~\ref{tt_sample_appendix}, we explain how to sample efficiently from a TT distribution and propose a sampling procedure to allow \emph{prioritized sampling}. In our applications, we do not necessarily want to sample from the whole distribution, but instead focus on obtaining samples from the high-density regions (e.g., when we only want to find the modes of the distribution) using \emph{prioritized sampling}.
This is described in the Algorithm~\ref{alg:tt_cd} in the Appendix. 

 The sampling procedure consists of repeated sampling of each dimension separately from a multinomial distribution. We provide a sampling parameter $\alpha \in (0,1)$ that can be chosen to adjust the sampling priority. When $\alpha=0$, the samples will be generated from the whole distribution (i.e., exact sampling), including from the low-density region (albeit with a lower probability). Higher $\alpha$ will focus the sampling around the area with higher density. This is ideal for robotics applications, as some applications require a very good initial solution for fast optimization (in that case, $\alpha$ is set near to one to obtain the best possible solution) while some others prefer the diversity of the solutions (by setting a small $\alpha$). As the sampling procedures can be done in parallel, we can generate many samples and select the best few samples according to their cost function values as the solution candidates $\hat{\bm{x}}_2^*$. 

\subsubsection{Fine-tuning the solution:}
The solution candidates $\hat{\bm{x}}_2^*$ obtained from the TT model can be further refined using local search methods to obtain the optimal solution $\bm{x}_2^*$. If all the decision variables are continuous and the cost function is differentiable, we can use $\hat{\bm{x}}_2^*$ as a warm-start for gradient-based optimization techniques to find the nearest local optima. If some of the decision variables are discrete, we can fix the discrete variables and only optimize the continuous ones. In this paper, we used SLSQP
for the refinement and we only deal with continuous variables. 

\subsection{Hierarchically Finding the TT Model}
When the objective function includes multiple objectives, we can approximate the corresponding PDF in the TT model using TT-cross in an efficient manner by exploiting the algebra associated with the TT. Suppose the cost function to be minimized is $C(\bm{x}) = C_a(\bm{x}_a) + C_b(\bm{x}_b)$,   (or $C(\bm{x}) = C_a(\bm{x}_a) C_b(\bm{x}_b)$) where $\bm{x} = \bm{x}_a \cup \bm{x}_b $. For example, $C_a$ could be the cost for obstacle avoidance which is only a function of joint angles and $C_b$ could be the cost for target reaching which is a function of joint angles and position of the target. Minimizing this cost function corresponds to maximizing the PDF $P(\bm{x}) =  P_a(\bm{x}_a) + P_b(\bm{x}_b)$. We can find the TT model $\mc{\bm{P}}$ corresponding to $P$ using $\mc{\bm{P}}_a$ and $\mc{\bm{P}}_b$ which are TT models corresponding to $P_a$ and $P_b$ respectively and they are often more smoother and easier to model using TT-cross. Then, we can quickly compute $\mc{\bm{P}} = \mc{\bm{P}}_a + \mc{\bm{P}}_b$ as an addition operation over tensor in TT format is efficient. For more information about algebraic operations over TT, we refer to \cite{cichocki2018_ttops}. This offers an efficient way to model a target PDF in TT format by separately modeling multiple individual components which are often favorable to compute in terms of dimensionality and smoothness.  


\begin{table}[t!]
\label{ttgo_algo}
\hrulefill
\caption*{TTGO Algorithm}
\hrulefill

\begin{enumerate}
    \item Training Phase (Offline):
    \begin{enumerate}
    \item Given: 
    \begin{itemize}
        \item Cost function $C(\bm{x}_1,\bm{x}_2)$,
        \item Rectangular domain $\Omega_{\bm{x}} = \Omega_{\bm{x}_1} \times \Omega_{\bm{x}_2}$
    \end{itemize}
    \item Transform the cost function into an unnormalized PDF $P(\bm{x}_1,\bm{x}_2)$.
    \item Discretize the domain $\Omega_x$ into  $\mathcal{X} = \mathcal{X}_1 \times \mathcal{X}_2$.
    \item Using TT-Cross, construct the TT-Model $\bm{\mathcal{P}}$ as the discrete analogue of $P(\bm{x})$ with discretization set $\mathcal{X}$.
    \end{enumerate}
    \item Inference Phase (Online):
    \begin{enumerate}
    \item Given: The task-parameter $\bm{x}_1=\bm{x}_t\in \Omega_{\bm{x}_1}$, the desired number of solutions $K$.
    \item Construct the conditional TT Model $\bm{\mc{P}}^{\bm{x}_t}$ from $\bm{\mathcal{P}}$ (see Section \ref{tt_condition}).
    \item Generate $N$ samples $\{ \bm{x}_2^l \}_{l=1}^N$ with the sampling parameter $\alpha \in (0,1)$ from the TT distribution $\text{Pr}(\bm{x}_2|\bm{x}_1=\bm{x}_t) = \frac{ |\bm{\mc{P}}^{\bm{x}_t}_{\bm{x}_2}|}{Z}$ (Algorithm \ref{alg:tt_cd}).
    \item Evaluate the cost function at these samples and choose the best-$K$ samples as approximation for the optima $\{ \bm{\hat{x}}_2^{*^l} \}_{l=1}^K$.
    \item Fine-tune the approximate solutions using gradient-based approaches on $C(\bm{x}_t,\bm{x}_2)$ to obtain the optima $\{ \bm{x}_2^{*^l} \}_{l=1}^K$.
    \end{enumerate}
\end{enumerate}

\hrulefill
\end{table}

\section{Experiments}
\label{experiments}

In this section, we evaluate the performance of the proposed algorithm with several applications. A PyTorch-based implementation of TTGO and the accompanying videos are available at \href{https://sites.google.com/view/ttgo/home}{https://sites.google.com/view/ttgo/home}. The implementation also provides the benefits of approximating a target PDF in TT format as compared to GMM or Neural Networks (NN).
 
We evaluated it on challenging benchmark functions such as the Rosenbrock function, Himmelblau function, and Gaussian Mixture Models, which are difficult for gradient-based optimization techniques. The experimental analysis provided in Appendix \ref{benchmark_appendix} demonstrates the proposed method can consistently find global optima, and multiple solutions when they exist, and it can adapt to task parameters that influence the locations of global optima. Additionally, the prioritized sampling approach proposed in the paper is evaluated, with small $\alpha$ values generating samples that cover a wide region around many local optima and $\alpha$ values close to one producing samples close to global optima.

 We then apply it to inverse kinematics with obstacles and motion planning problems. 
Besides qualitatively observing the solutions, we also perform quantitative analyses to evaluate the quality of the approximate solutions produced by TTGO. We consider three different metrics:
\begin{itemize}
    \item $c_i$, the initial cost value of the approximate solutions.
    \item $c_f$, the cost value after refinement.
    \item \textbf{Success}, the percentage of samples that converge to a good solution, i.e., with the cost value below a given threshold.
\end{itemize}

To compare the performance of the proposed approach with random initialization, we initialize the solver with random samples generated from the \emph{uniform} distribution across the entire domain. To observe the effect of prioritized sampling, we also use TTGO with various values of $\alpha$. The performance evaluation involves generating 100 random test cases within the task space. For each test case, we generate $N$ samples using both the TT and uniform distribution methods and select the best sample based on the initial cost value as the approximate solution. We vary the number of samples $N$ from 1 to 1000. The SLSQP solver is used to optimize the sample with respect to the given cost function. We then evaluate the initial cost $c_i$, the final cost after refinement $c_f$, and the convergence status for each method. The average performance of both methods across all test cases is computed. The results for the robotics applications are summarized in Table~\ref{tab: ik}-\ref{tab: motion_via} and are discussed in the corresponding sections.

\begin{figure*}[t]
    \centering
    \subfigure[]{%
        \includegraphics[height=0.22\textwidth]{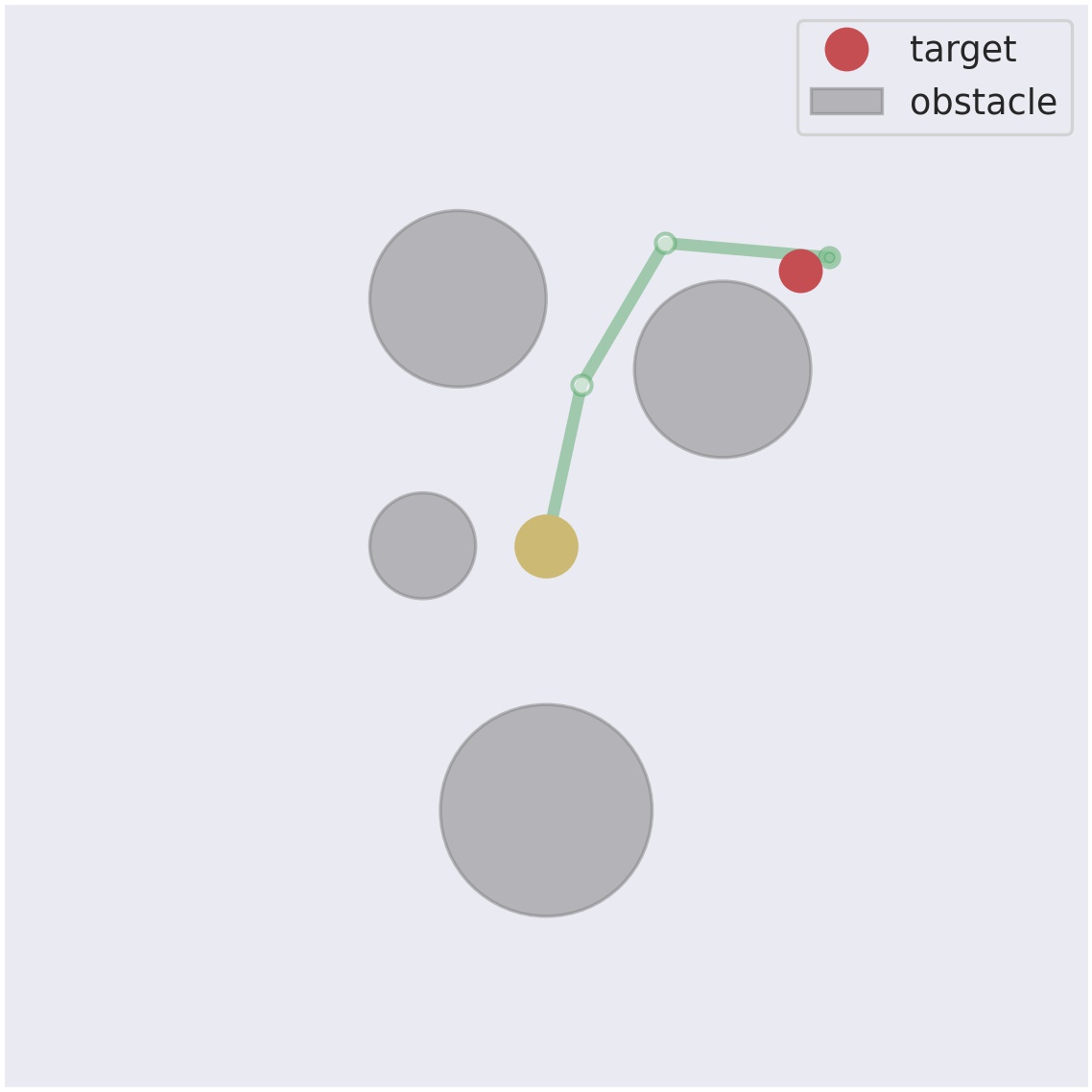}
        }
    \hfill
    \subfigure[]{%
        \includegraphics[height=0.22\textwidth]{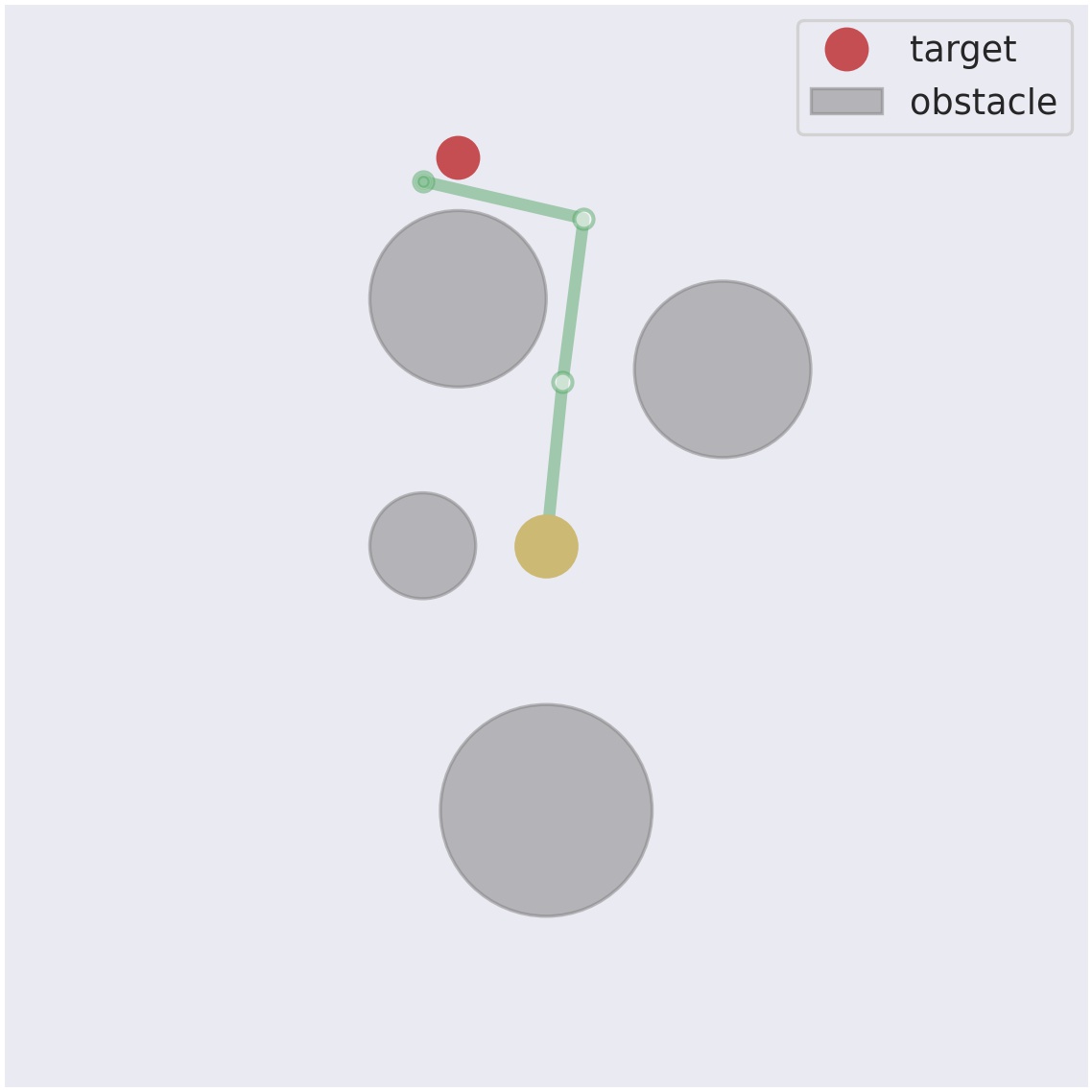}
        }
    \hfill
    \subfigure[]{%
        \includegraphics[height=0.22\textwidth]{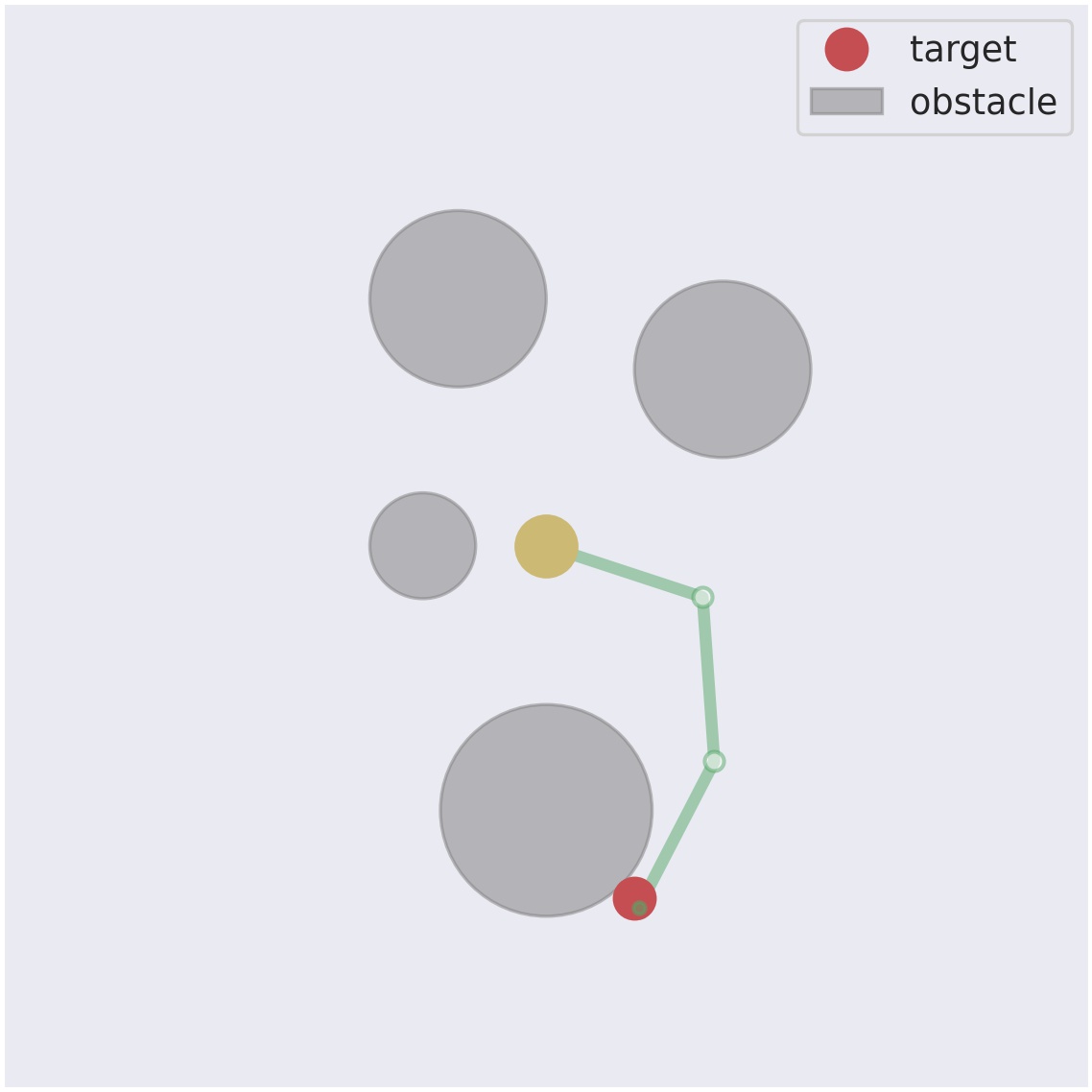}
        }
    \hfill
    \subfigure[]{%
        \includegraphics[height=0.22\textwidth]{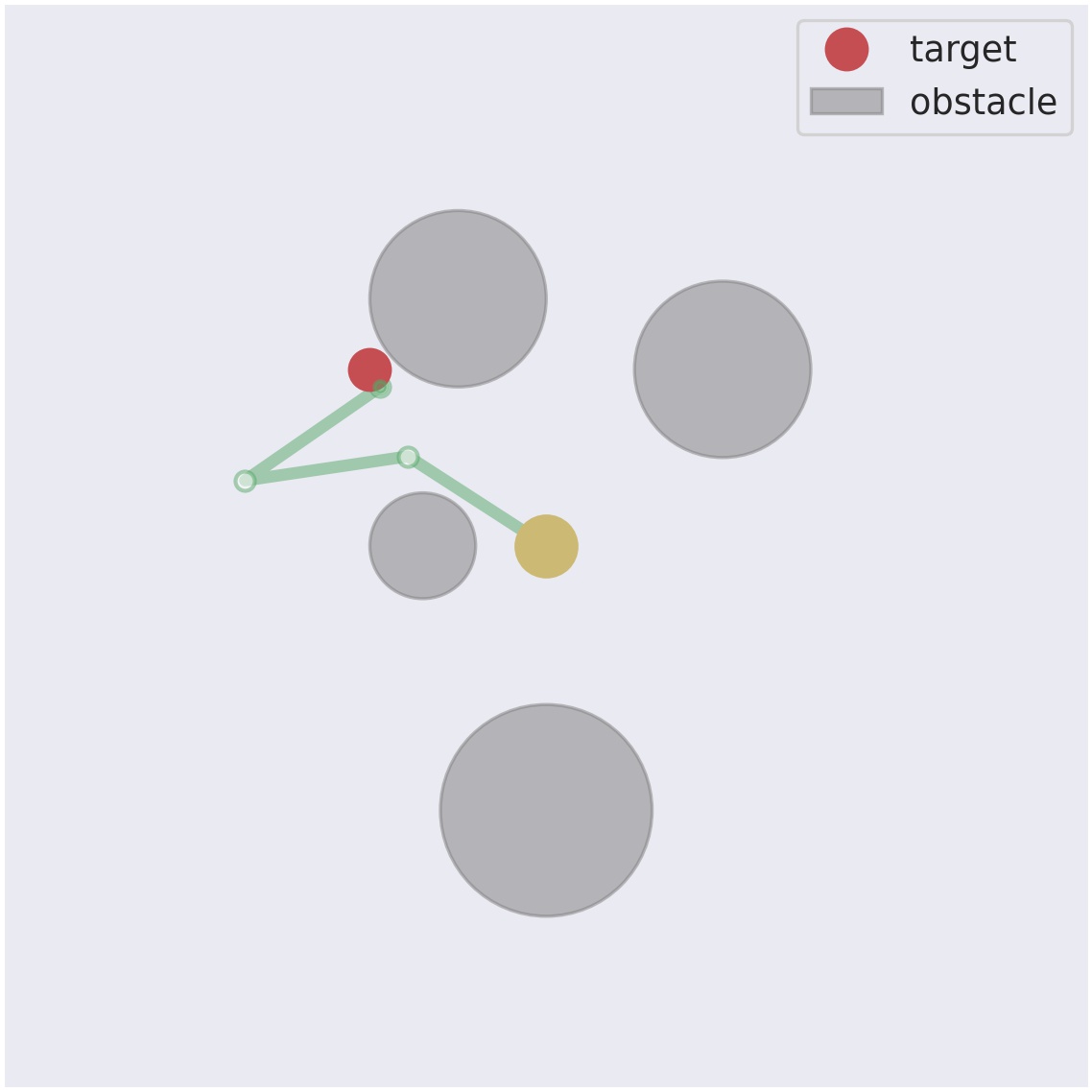}
        }
    
    \caption{A single sample taken from a conditional TT distribution with $\alpha=1$ for inverse kinematics of a $3$-link planar manipulator in the presence of obstacles (gray spheres). The yellow circle and the green segments depict the base and the links of the robot, respectively. The target end-effector positions are shown in red. The samples are very close to the targets and collision-free, even without refinement.}
    \label{fig:planar_ik_single}
\end{figure*}

\begin{figure*}[t]
    \centering
    \subfigure[]{%
        \includegraphics[height=0.22\textwidth]{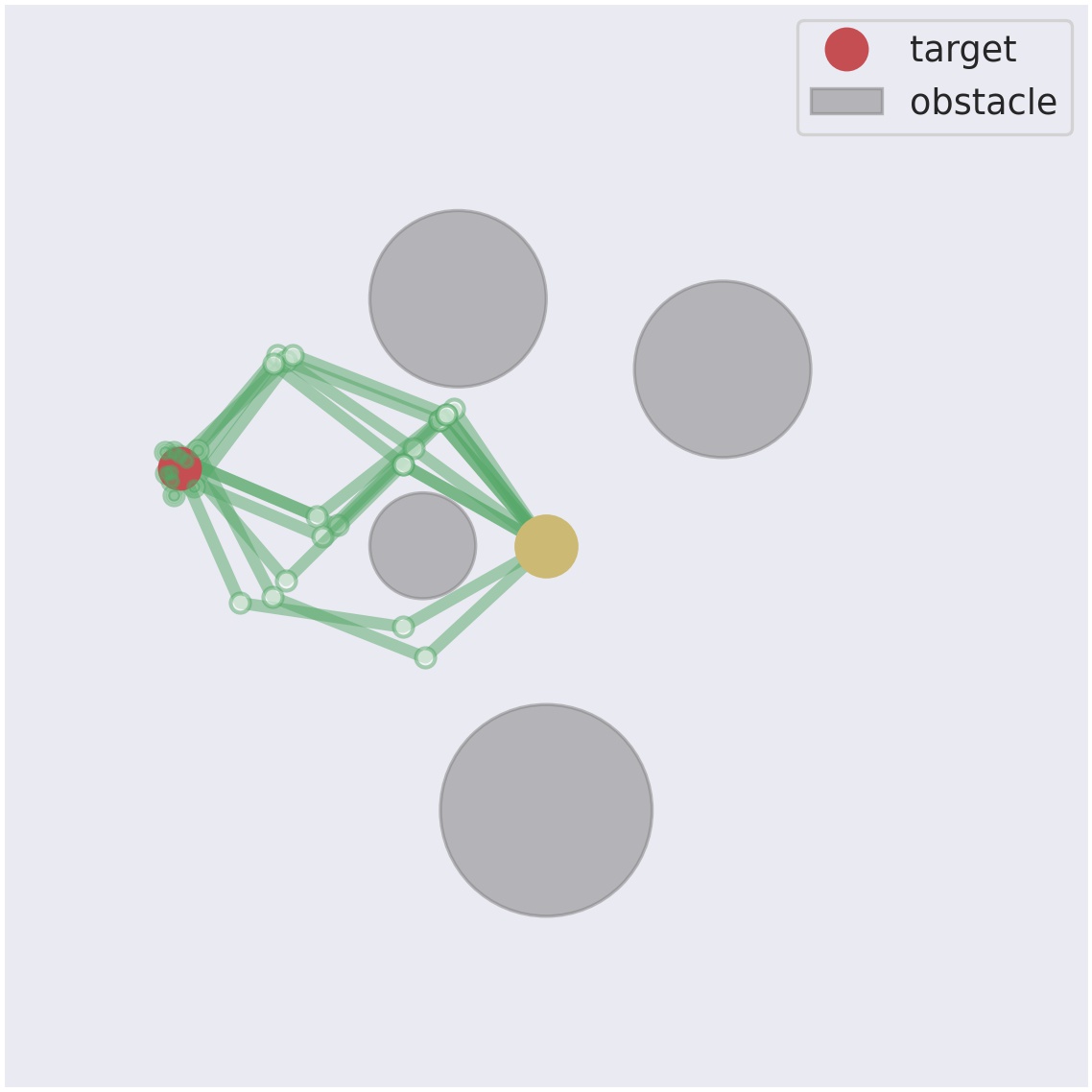}
        }
    \hfill
    \subfigure[]{%
        \includegraphics[height=0.22\textwidth]{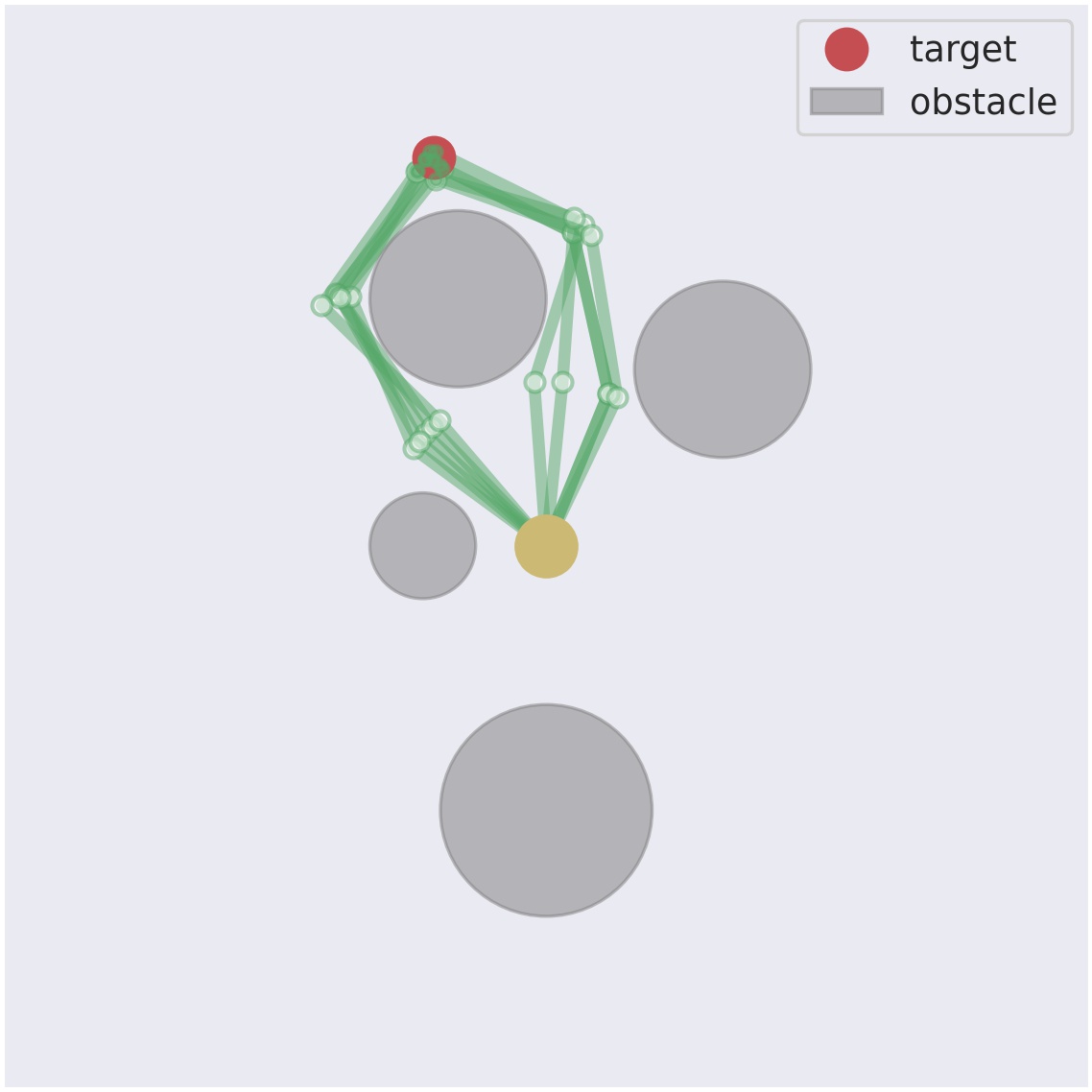}
        }
    \hfill
    \subfigure[]{%
        \includegraphics[height=0.22\textwidth]{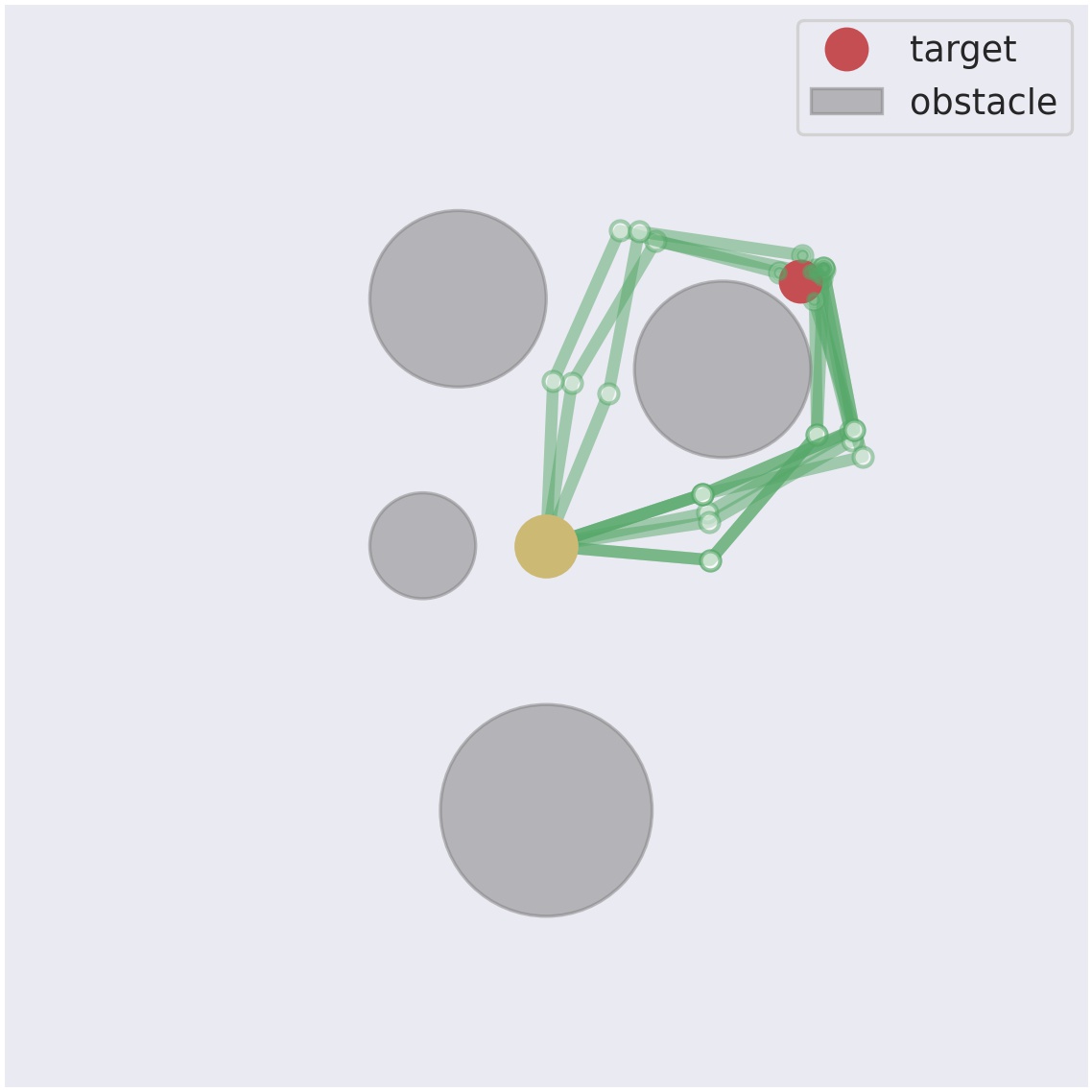}
        }
    \hfill
    \subfigure[]{%
        \includegraphics[height=0.22\textwidth]{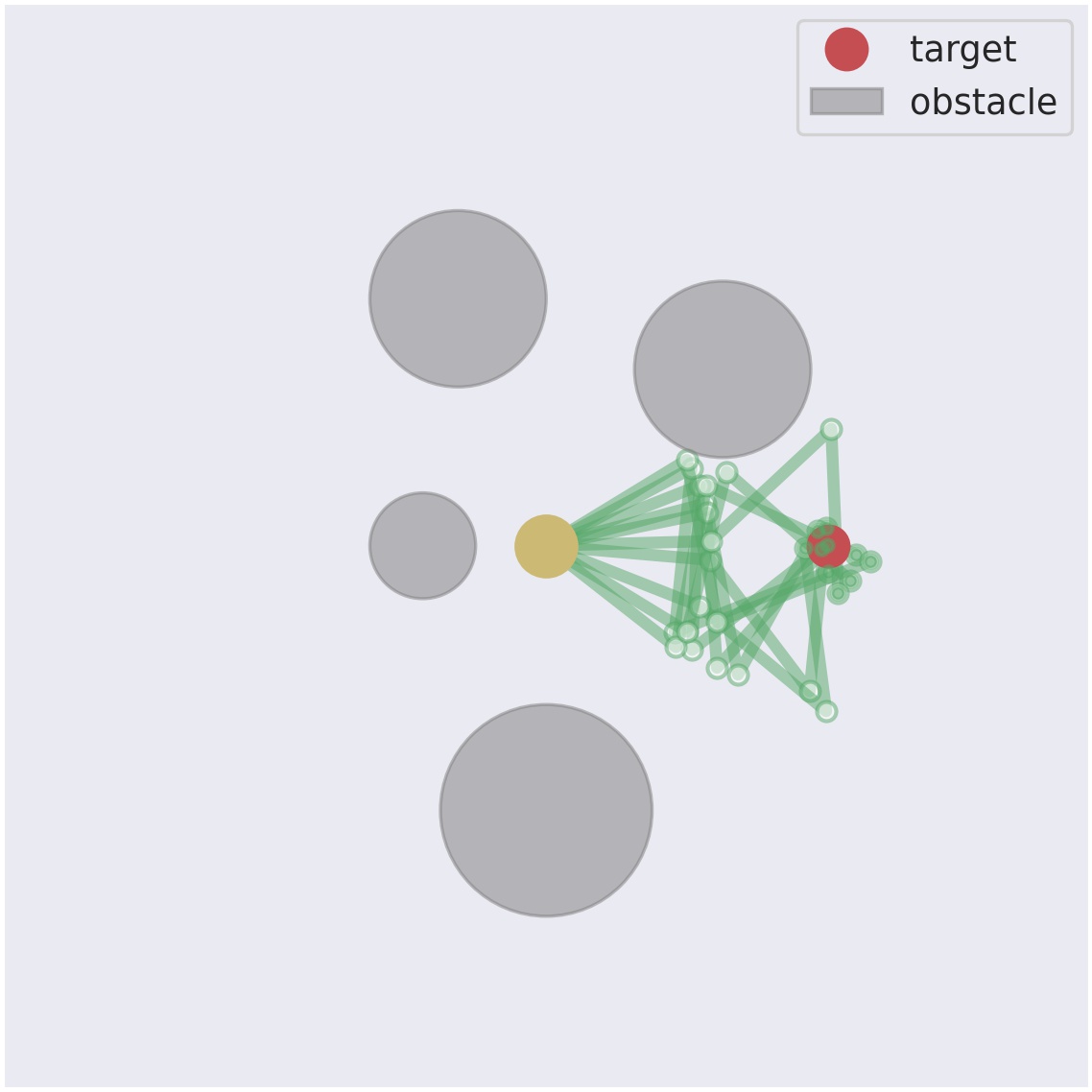}
        }
    
    \caption{Best $10$ out of $50$ samples taken from a conditional TT distribution with $\alpha=0.8$ for inverse kinematics of a $3$-link planar manipulator in the presence of obstacles. The samples are already close enough to the optima even without refinement and the multimodality of the solutions is clearly visible.}
    \label{fig:planar_ik_multi}
\end{figure*}

\begin{figure}[t]
    \centering
    \subfigure[]{%
        \includegraphics[width=0.22\textwidth]{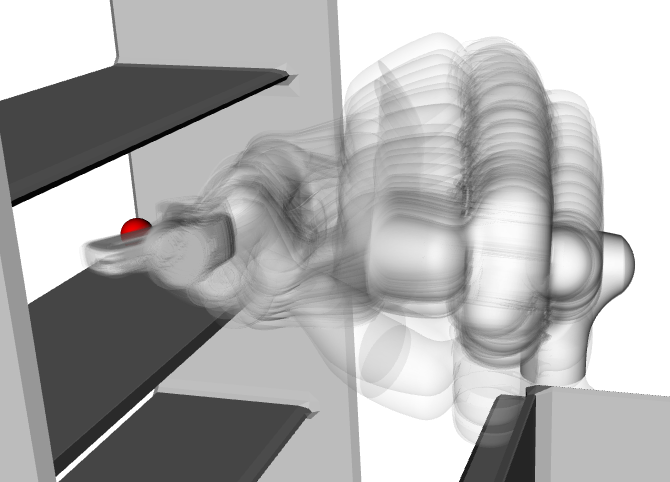}
        }
    \hfill
    \subfigure[]{%
        \includegraphics[width=0.22\textwidth]{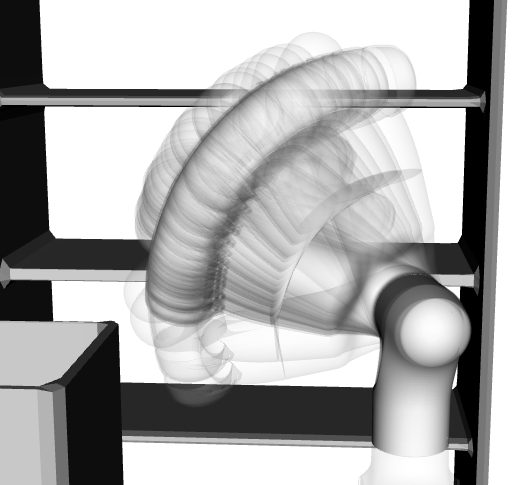}
        }
    \\
    \subfigure[]{%
        \includegraphics[width=0.22\textwidth]{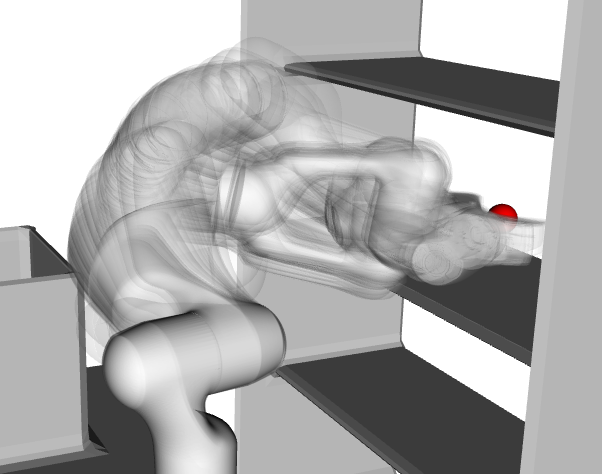}
        }
    \hfill
    \subfigure[]{%
        \includegraphics[width=0.22\textwidth]{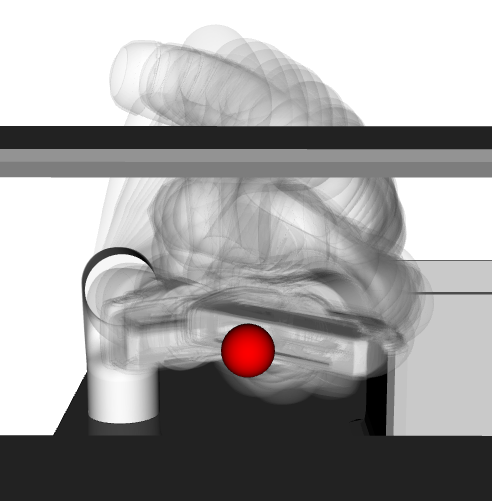}
        }
    \caption{The samples taken from a conditional TT distribution for the IK of a Franka Emika manipulator in the presence of obstacles, after refinement. We can see that there is a continuous set of solutions due to the additional degrees of freedom.}
    \label{fig:panda_ik_1}
\end{figure}

\subsection{Inverse Kinematics problems}
\label{ik}
We consider here the optimization formulation of numerical Inverse Kinematics (IK). The task parameters $\bm{x}_1$ then correspond to the desired end-effector pose, while the decision variables $\bm{x}_2$ are the joint angles. We use approximately $n_2=50$ discretization points for each of the joint angles ($\sim \ang{5}$) and approximately $n_1=200$ discretization points ($\sim 0.5$cm) for each task parameter.  $\Omega_{\bm{x}_1} \subset \mathbb{R}^{3}$ is the rectangular space that includes the robot workspace. $\Omega_{\bm{x}_2} = \times_{k=1}^{d_2} [\theta_{\min_k}, \theta_{\max_k}]$, $\Omega_{\bm{x}_2} \subset \mathbb{R}^{d_2}$ where $[\theta_{\min_k}, \theta_{\max_k}]$ represents the joint angle limits for the $k$-th joint.  

We consider two IK problems: 6-DoF IK to clearly demonstrate the multimodal solutions and 7-DoF IK with obstacle cost to consider the infinite solution space. In both cases, we transform the cost function into a density function as $P(\bm{x})= \exp(-C(\bm{x})^2)$. 

\subsubsection{Inverse Kinematics for 6-DoF Robot:}
\label{ik_6dof}

A 6-DoF robot has a finite number of joint angle configurations that correspond to a given end-effector pose. In this section, we consider the 6-DoF Universal Robot that can have up to 8 IK solutions. While there is an analytical solution for such robots, it is a nice case study to illustrate the capability of TTGO to approximate multimodal distributions in a robotics problem where the modes are very distinct from one another. We constrain the end-effector orientation to a specific value (i.e., facing upward without any free axis of rotation), and set the end-effector position as the task parameter. Hence, $\bm{x}_1 \in \Omega_{\bm{x}_1}\subset \mathbb{R}^3$ while $\bm{x}_2 \in \Omega_{\bm{x}_2} \subset \mathbb{R}^6$, so $d = 9$, where $\Omega_{\bm{x}_1}$ is the rectangular domain enclosing the workspace of the manipulator.

\begin{figure*}[t]
    \centering
    \subfigure[]{%
        \includegraphics[height=0.22\textwidth]{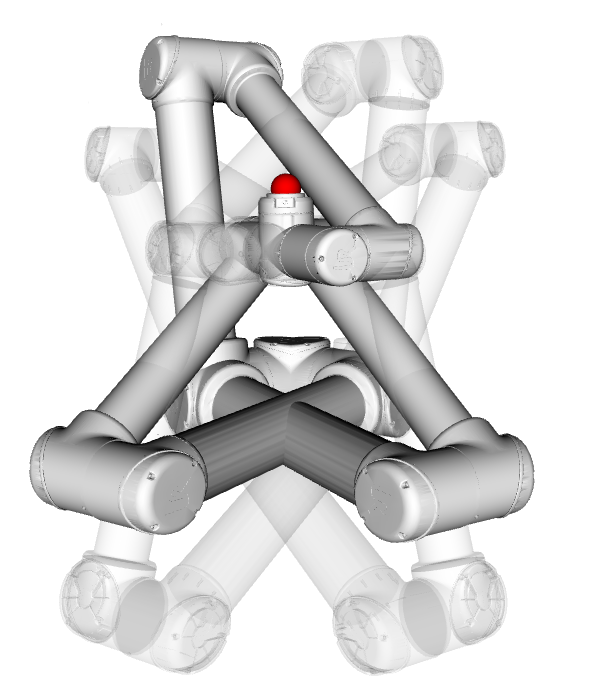}
        }
    \hfill
    \subfigure[]{%
        \includegraphics[height=0.22\textwidth]{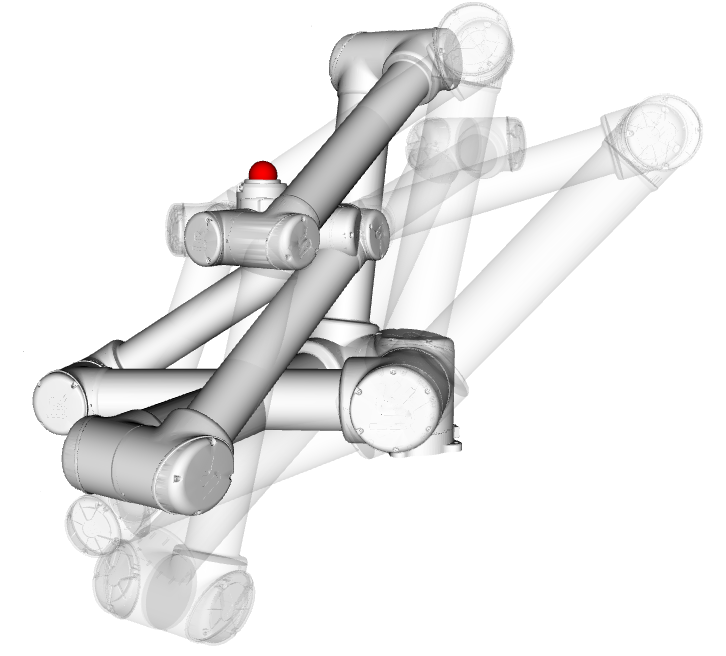}
        }
    \hfill
    \subfigure[]{%
        \includegraphics[height=0.22\textwidth]{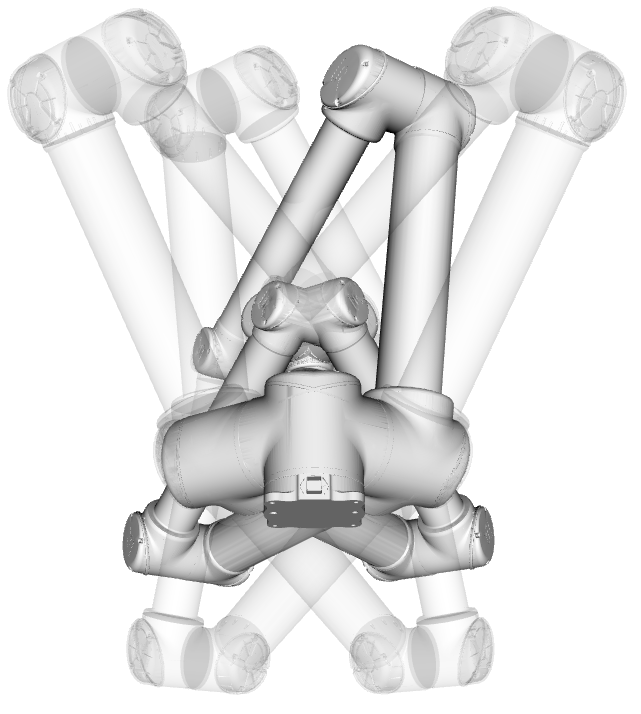}
        }
    \hfill
    \subfigure[]{%
        \includegraphics[height=0.22\textwidth]{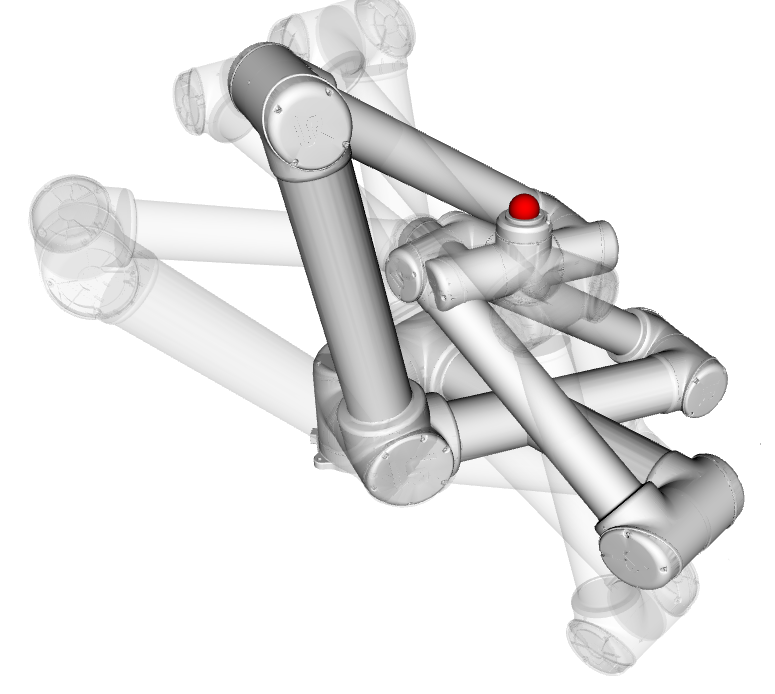}
        }
    \caption{8 IK solutions of the UR10 robot for a given pose from TTGO samples after refinement, shown from four different views. 5 of the solutions are drawn transparently to provide better visualization. The desired end-effector position is shown in red.}
    \label{fig:ur10_ik}
\end{figure*}

We observe that TTGO is able to retrieve most of the 8 IK solutions for a given end-effector pose. Figure~\ref{fig:ur10_ik} shows the refined samples from TTGO by conditioning the TT distribution on a desired end-effector position. This validates our claim that TTGO is able to approximate multimodal solutions even for a complex distribution. 

\subsubsection{Inverse Kinematics for 7-DoF Robot with Obstacle Cost:}
\label{ik_7dof}

A 7-DoF robot can have an infinite number of joint angle configurations for a given end-effector pose, unlike a 6-DoF robot. It can also have multiple solution modes similar to the 6-DoF robot. To ensure a collision-free solution, we introduce an obstacle cost to the optimization formulation, using the same collision cost as in CHOMP~\citep{Ratliff2009}. This collision cost uses a precomputed Signed-distance Function (SDF) to compute the distance between each point on the robot link and the nearest obstacle. When there are obstacles, numerical IK typically involves generating multiple solutions and checking for collision until a collision-free configuration is found. However, in cluttered environments, this approach may have a low success rate, requiring the user to generate many IK solutions before finding a collision-free one. By adding an obstacle cost to optimize for collision-free configurations directly, the non-convexity of the problem increases significantly, leading the solver to get stuck at poor local optima, especially with a high weight on the obstacle cost. Therefore, it is an interesting case study to demonstrate how TTGO can avoid poor local optima and find robust solutions in this challenging scenario.

We first test the IK with obstacle cost for a 3-DoF planar robot to provide some intuition on the effectiveness of TTGO. Figure~\ref{fig:planar_ik_single} and Figure~\ref{fig:planar_ik_multi} show some samples from TTGO conditioned on the target end-effector position (shown in red).
By setting $\alpha=1$, we focus the sampling around the mode of the distribution, enabling us to obtain a very good solution even with only 1 sample (Figure~\ref{fig:planar_ik_single}). As we decrease  $\alpha$ to $0.8$ and retrieve more samples, we can see that multiple solutions can be obtained easily (Figure~\ref{fig:planar_ik_multi}). Note that even without the refinement step, all samples reach the goal closely while being collision-free.

We then apply the formulation on the 7-DoF Franka Emika robot, where the collision environment is set to be a table, a box, and a shelf.
The task parameters correspond to the end-effector position in the shelf, while the gripper is constrained to be oriented horizontally with one free DoF around the vertical axis. Hence, $\bm{x}_1 \in \mathbb{R}^3$ while $\bm{x}_2 \in \mathbb{R}^7$, so $d = 10$. The number of parameters of the TT cores is $1.4 \times 10^7$ whereas the original tensor $\bm{\mc{P}}$ has $1 \times 10^{18}$ parameters. TT-cross found the tensor in TT-format using only $2 \times 10^8$ evaluations of the function $P$. For this application, a rank of $60$ already produces satisfactory performance.

Figure~\ref{fig:panda_ik_1} shows samples generated from a TT distribution on a given end-effector position after refinement. Note that unlike in the 6 DoF case, we can see here a continuous set of IK solutions due to the additional degrees of freedom. We also note that distinctly different modes of solutions can also be observed in this case, as can be seen in the accompanying video. 

The results are reported in Table \ref{tab: ik}. We can see that TTGO consistently outperforms uniform sampling by a wide margin across the three metrics. The initial cost values of TTGO samples are much lower than uniform samples, and after refinement, they converge to smaller cost values on average. The success rates of TTGO samples are also much higher. Furthermore, from qualitative analysis, the approximate solutions of TTGO are very close to the optimized solution. It is especially important to note that the best out of 1000 uniform samples (bottom right corner of the table) is still worse than a single sample from TTGO with $\alpha>0.75$ (top left corner). 

We can see the effect of prioritized sampling by comparing the performance of different values of $\alpha$. In general, using higher values of $\alpha$ improves the performance, as we concentrate the samples around the high-density region. TTGO samples with $\alpha=0.9$ have impressive performance with 94\% success rates even by using only one sample per test case. However, higher $\alpha$ means less diversity of solutions, so a trade-off between solution quality and diversity needs to be considered when choosing the value of $\alpha$. Note that even with $\alpha=0$ we still obtain a very good performance by using as few as 10 samples. 

\begin{table*}[!t]
    \centering
	\renewcommand{\arraystretch}{1.3} 
	\scriptsize
	\caption*{\textbf{Tables 1--3.} The performance measures for three different applications with the Franka Emika manipulator. We compare the performance of TTGO for initializing a given gradient-based solver (namely, SLSQP) against initialization from uniform distribution. The three performance metrics are the cost at the initialization ($c_i$), the cost after optimization ($c_f$) using the solver and the success rate. The criteria for success is that $c_f \leq 0.25$. We compute the average of each of these measures over $100$ randomly chosen test cases. Each of the target points are chosen so that they are sufficiently away from the surface of the obstacle but they are not guaranteed to be feasible. }
    \caption{Inverse kinematics of the Franka Emika robot}
	\begin{tabular}{c|c|ccc|ccc|ccc|cccc|cccc}
		\hline
		\multirow{3}{*}{Method} & \multirow{3}{*}{$\alpha$} & \multicolumn{12}{c}{\textbf{\# Samples}} \\
		\cline{3-14}
		& & \multicolumn{3}{c|}{$1$} & \multicolumn{3}{c|}{$10$} & \multicolumn{3}{c|}{$100$} &  \multicolumn{3}{c}{$1000$}\\
		\cline{3-5} \cline{6-8} \cline{9-11} \cline{12-14}
		& & $\text{c}_\text{i}$ & $\text{c}_\text{f}$  & \textbf{Success}  & $\text{c}_\text{i}$ & $\text{c}_\text{f}$ & \textbf{Success}  & $\text{c}_\text{i}$ & $\text{c}_\text{f}$ &  \textbf{Success} &  $\text{c}_\text{i}$ & $\text{c}_\text{f}$ & \textbf{Success} \\
		\hline
		\multirow{4}{*}{TTGO} 
                    & $0.9$ & $1.04$ & $0.01$ & $94.00$\% & $0.55$ & $0.02$ & $98.00$\%  & $0.37$ & $0.02$ & $98.00$\%  & $0.26$ & $0.02$ & $99.00$\%  \\ & $0.75$ & $1.52$ & $0.07$ & $84.00$\%  & $0.65$ & $0.02$ & $95.00$\%  & $0.37$ & $0.02$ & $95.00$\%  & $0.24$ & $0.03$ & $97.00$\%  \\ & $0.5$ & $2.01$ & $0.08$ & $88.00$\%  & $0.85$ & $0.04$ & $93.00$\%  & $0.43$ & $0.04$ & $93.00$\%  & $0.28$ & $0.01$ & $98.00$\%  \\ & $0$ & $2.88$ & $0.17$ & $71.00$\%  & $1.23$ & $0.05$ & $91.00$\%  & $0.68$ & $0.05$ & $91.00$\%  & $0.39$ & $0.04$ & $96.00$\% \\		
		\cline{1-6} \cline{7-10} \cline{11-14} \cline{15-18}		
		\multirow{1}{*}{Uniform} 
        &- & $8.42$ & $1.22$ & $37.75$\%  & $4.47$ & $0.91$ & $45.50$\%  & $2.56$ & $0.5$ & $59.25$\%  & $1.59$ & $0.24$ & $75.00$\%  \\
		\hline
	\end{tabular}
	\label{tab: ik}
\end{table*}

\vspace*{1 cm}

\begin{table*}
    \centering
	\renewcommand{\arraystretch}{1.3} 
	\scriptsize
	\caption{Target Reaching}
	\begin{tabular}{c|c|ccc|ccc|ccc|cccc|cccc}
		\hline
		\multirow{3}{*}{Method} & \multirow{3}{*}{$\alpha$} & \multicolumn{12}{c}{\textbf{\# Samples}} \\
		\cline{3-14}
		& & \multicolumn{3}{c|}{$1$} & \multicolumn{3}{c|}{$10$} & \multicolumn{3}{c|}{$100$} &  \multicolumn{3}{c}{$1000$}\\
		\cline{3-5} \cline{6-8} \cline{9-11} \cline{12-14}
		& & $\text{c}_\text{i}$ & $\text{c}_\text{f}$  & \textbf{Success}  & $\text{c}_\text{i}$ & $\text{c}_\text{f}$ & \textbf{Success}  & $\text{c}_\text{i}$ & $\text{c}_\text{f}$ &  \textbf{Success} &  $\text{c}_\text{i}$ & $\text{c}_\text{f}$ & \textbf{Success} \\
		\hline
		\multirow{4}{*}{TTGO} 
          & $0.9$ & $3.99$ & $0.17$ & $62.00$\%  & $1.1$ & $0.09$ & $86.00$\%  & $0.71$ & $0.1$ & $86.00$\%  & $0.58$ & $0.09$ & $88.00$\%  \\ & $0.75$ & $5.63$ & $0.21$ & $53.00$\%  & $1.29$ & $0.14$ & $72.00$\%  & $0.78$ & $0.1$ & $86.00$\%  & $0.56$ & $0.1$ & $83.00$\%  \\ & $0.5$ & $4.53$ & $0.17$ & $50.00$\%  & $1.54$ & $0.14$ & $64.00$\%  & $0.96$ & $0.11$ & $83.00$\%  & $0.62$ & $0.1$ & $84.00$\%  \\ & $0$ & $6.7$ & $0.31$ & $46.00$\%  & $2.06$ & $0.18$ & $60.00$\%  & $1.3$ & $0.12$ & $82.0$ & $0.84$ & $0.12$ & $86.00$\%  \\ 
		\cline{1-6} \cline{7-10} \cline{11-14} \cline{15-18}		
		\multirow{1}{*}{Uniform} 
          &- & $13.85$ & $1.34$ & $19.25$\%  & $4.79$ & $0.91$ & $28.75$\%  & $3.02$ & $0.68$ & $41.00$\%  & $2.06$ & $0.45$ & $53.50$\%  \\ 
		\bottomrule[0.12em]
	\end{tabular}
	\label{tab: motion_reaching}
\end{table*}

\begin{table*}
    \centering
	\renewcommand{\arraystretch}{1.3} 
	\scriptsize
	\caption{Pick-and-Place}
	\begin{tabular}{c|c|ccc|ccc|ccc|cccc|cccc}
		\hline
		\multirow{3}{*}{Method} & \multirow{3}{*}{$\alpha$} & \multicolumn{12}{c}{\textbf{\# Samples}} \\
		\cline{3-14}
		& & \multicolumn{3}{c|}{$1$} & \multicolumn{3}{c|}{$10$} & \multicolumn{3}{c|}{$100$} &  \multicolumn{3}{c}{$1000$}\\
		\cline{3-5} \cline{6-8} \cline{9-11} \cline{12-14}
		& & $\text{c}_\text{i}$ & $\text{c}_\text{f}$  & \textbf{Success}  & $\text{c}_\text{i}$ & $\text{c}_\text{f}$ & \textbf{Success}  & $\text{c}_\text{i}$ & $\text{c}_\text{f}$ &  \textbf{Success} &  $\text{c}_\text{i}$ & $\text{c}_\text{f}$ & \textbf{Success} \\
		\hline
		\multirow{4}{*}{TTGO} 
                		
          & $0.9$ & $2.41$ & $0.16$ & $70.00$\%  & $1.41$ & $0.15$ & $81.00$\%  & $1.05$ & $0.15$ & $79.00$\%  & $0.87$ & $0.14$ & $89.00$\%  \\ & $0.75$ & $3.25$ & $0.17$ & $66.00$\%  & $1.71$ & $0.17$ & $66.00$\%  & $1.31$ & $0.14$ & $84.00$\%  & $1.01$ & $0.15$ & $78.00$\%  \\ & $0.5$ & $4.31$ & $0.26$ & $54.00$\%  & $2.33$ & $0.19$ & $62.00$\%  & $1.66$ & $0.17$ & $77.00$\%  & $1.29$ & $0.18$ & $76.00$\%  \\ & $0$ & $6.2$ & $0.27$ & $48.00$\%  & $2.98$ & $0.23$ & $48.00$\%  & $2.17$ & $0.21$ & $58.00$\%  & $1.61$ & $0.18$ & $71.00$\%  \\
		\cline{1-6} \cline{7-10} \cline{11-14} \cline{15-18}		
		\multirow{1}{*}{Uniform} 
         &- & $9.64$ & $0.78$ & $23.75$\%  & $5.23$ & $0.63$ & $30.25$\%  & $3.95$ & $0.49$ & $39.5$\%  & $3.07$ & $0.39$ & $44.25$\%  \\
		\bottomrule[0.12em]
	\end{tabular}
	\label{tab: motion_via}
\end{table*}

\subsection{Motion Planning of Manipulators}
\label{motion_plan}
In this section, we explore the use of our framework in the motion planning of the Franka Emika robot to generate obstacle-free robot motions. This problem is highly dimensional, as a robot with $m$ degrees of freedom and $T$ time intervals results in optimization variables $\bm{x}_1$ with $mT$ dimensions. To address this issue, we adopt a trajectory representation using movement primitives with basis functions, as commonly done in learning from demonstration~\citep{paraschos2013probabilistic,Calinon19MM}. The optimization variables in this representation consist of the superposition weights of the basis functions, which is much smaller than the number of configurations. Moreover, our formulation of movement primitives guarantees that the motion always starts from the initial configuration and ends at the given final configuration. In the case of a goal in the task space, we need to first determine the corresponding final configuration, which can be done using inverse kinematics. In our motion planning formulation, we optimize both the final configuration and the weights of the basis functions jointly.

The cost function in our motion planning formulation includes the reaching cost, the joint limit cost, the smoothness cost, and the obstacle cost, which is the same cost used in inverse kinematics. It is important to note that if we want to ensure that the solution avoids small obstacles, the number of time discretizations must be large, which can result in more than $700$ dimensions for motion planning with a Franka Emika robot. The use of the obstacle cost helps the solver directly optimize for a collision-free configuration, but it also significantly increases the non-convexity of the problem, making it susceptible to getting stuck in poor local optima, especially with a large weight on the obstacle cost. Details on the motion planning formulation can be found in Appendix~\ref{appendix_mp}.

We consider two different motion planning tasks as follows:
\begin{enumerate}
    \item \textbf{Target Reaching}: From the initial configuration $\bm{\theta}_0 \in \mb{R}^m$, reach a target location $\bm{p_d} \in \mb{R}^3$.
    \item \textbf{Pick-and-Place}: From the initial configuration $\bm{\theta}_0 \in \mb{R}^m$, reach two target locations $\bm{p_d}^1$ (picking location) and $\bm{p_d}^2$ (placing location) in sequence before returning to the initial configuration $\bm{\theta}_0$.
\end{enumerate}

For the target reaching problem, the task parameter is the target location $\bm{x}_1=\bm{p_d}$ and the decision variables $\bm{x}_2=(\bm{\theta}_1, \bm{w})$. Here, $\bm{\theta}_1 \in \Omega_{\bm{\theta}} \subset \mb{R}^{m}$  is the joint angle defining the final configuration and $\bm{w}=(\bm{w}^k)_{k=1}^{m} \in \mb{R}^{Jm}$, where $\bm{w}^k=(w^k_j)_{j=1}^J \in \bm{R}^J$ are the superposition weights of the basis functions representing the motion from $\bm{\theta}_0$ to $\bm{\theta}_1$. We use $J=2$ and $m=7$ for the 7-DoF Franka Emika manipulator, so the total number of dimensions for the reaching task is $d = 3 + 7 + 2\times7 = 24$.

For the pick-and-place problem, the task parameters are the two target locations (pick and place location): $\bm{x}_1=(\bm{p_d}^1,\bm{p_d}^2)$. The decision variables are $\bm{x}_2=(\bm{\theta}_1,\bm{\theta}_2,{}^{01}\bm{w},{}^{12}\bm{w},{}^{20}\bm{w})$, where $\bm{\theta}_1$ and $\bm{\theta}_2$ are the configurations corresponding to the two target points, $\bm{w} =({}^{01}\bm{w}^k,{}^{12}\bm{w}^k,{}^{20}\bm{w}^k)_{k=1}^m$ where ${}^{uv}\bm{w} \in \mb{R}^{Jm}$ are the weights of the basis functions representing the movement from the configuration $\bm{\theta}_u$ to $\bm{\theta}_v$. Hence, the total number of dimensions for the pick-and-place task is $d = 2\times3 + 2\times7 + 3\times2\times7 = 62$.

We use the transformation $P(\bm{x})=\exp(-C(\bm{x})^2)$. The target location $\bm{p_d}$ for target reaching and $\bm{p_d}^1$ in the pick-and-place problem are inside the shelf as in the IK problems (picking location). For the pick-and-place task, the second target location $\bm{p_d}^2$ is on the top of the box (drop location). We discretize each of the task parameters using $100$ points and the decision variables with $30$ points. We use radial basis functions with $J=2$, which we find sufficient for our applications. The bounds on the weights of basis function for a joint are the same as the joint limits i.e., $(w^k_{min},w^k_{max}) = (\theta_{{min}_k},\theta_{{max}_k})$.

\begin{figure}[t]
    \centering
    \subfigure[TTGO Task-1]{%
        \includegraphics[width=0.23\textwidth]{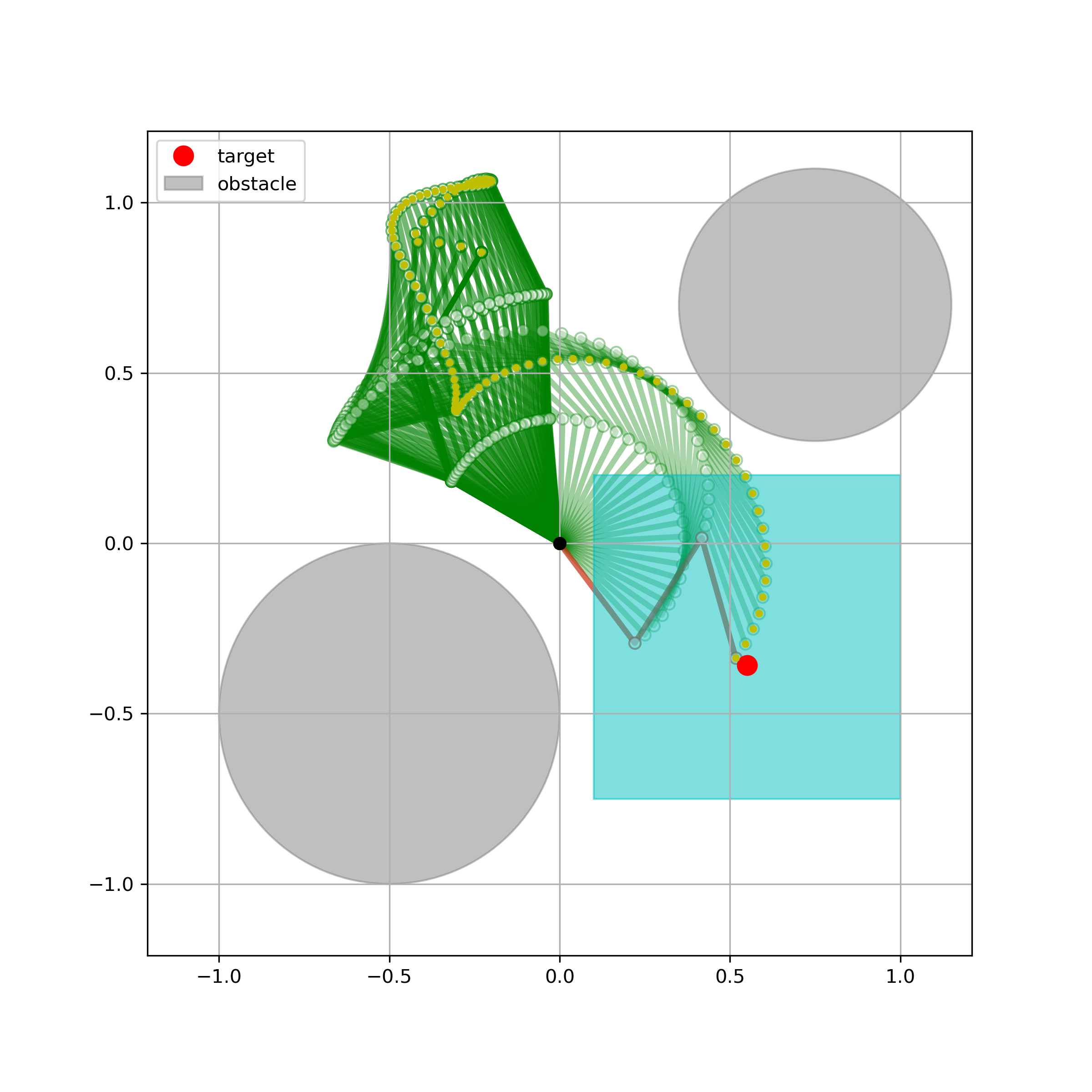}
        }
        \hfill
    \subfigure[TTGO Task-2]{%
        \includegraphics[width=0.23\textwidth]{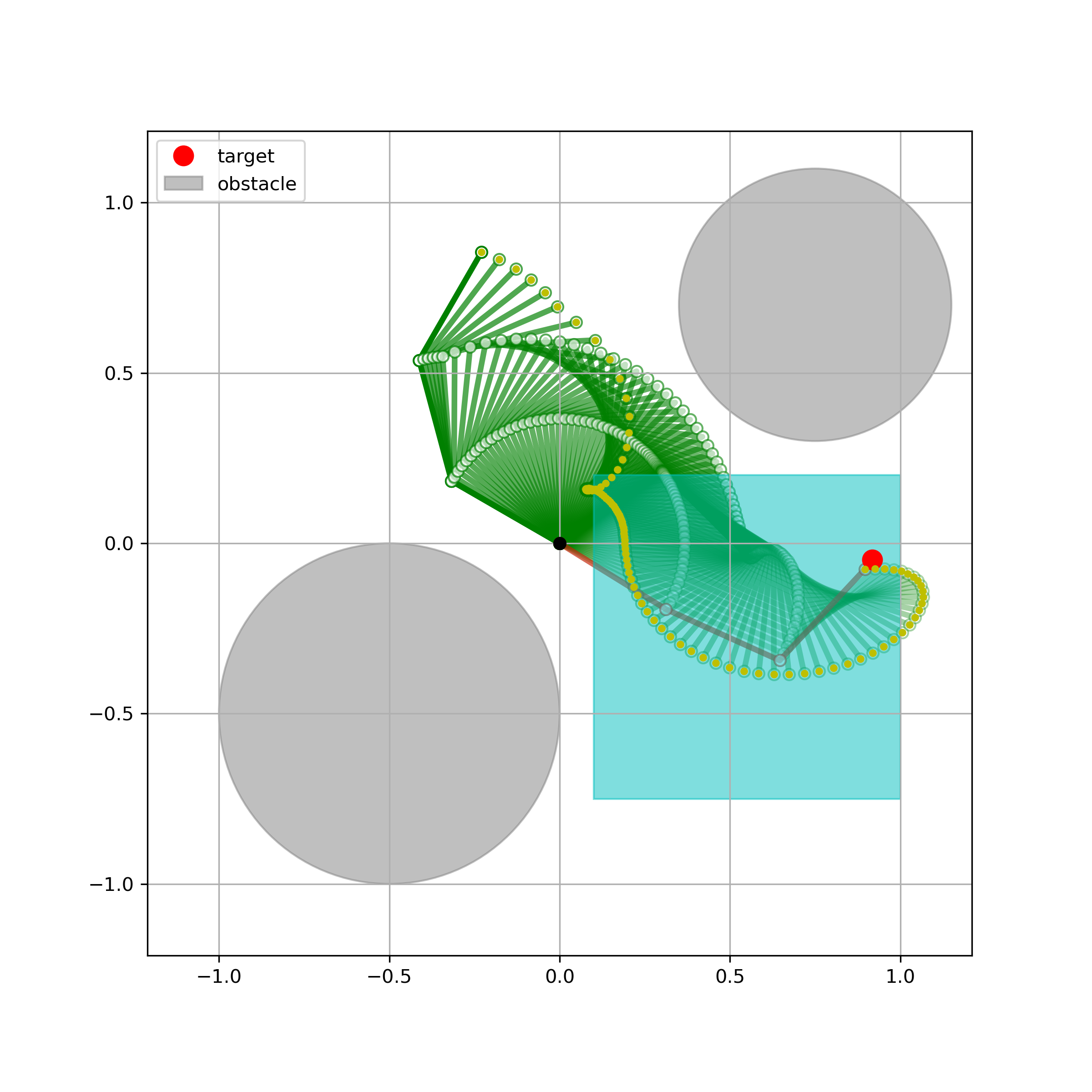}
        }
        \\
    \subfigure[Random Task-1]{%
        \includegraphics[width=0.23\textwidth]{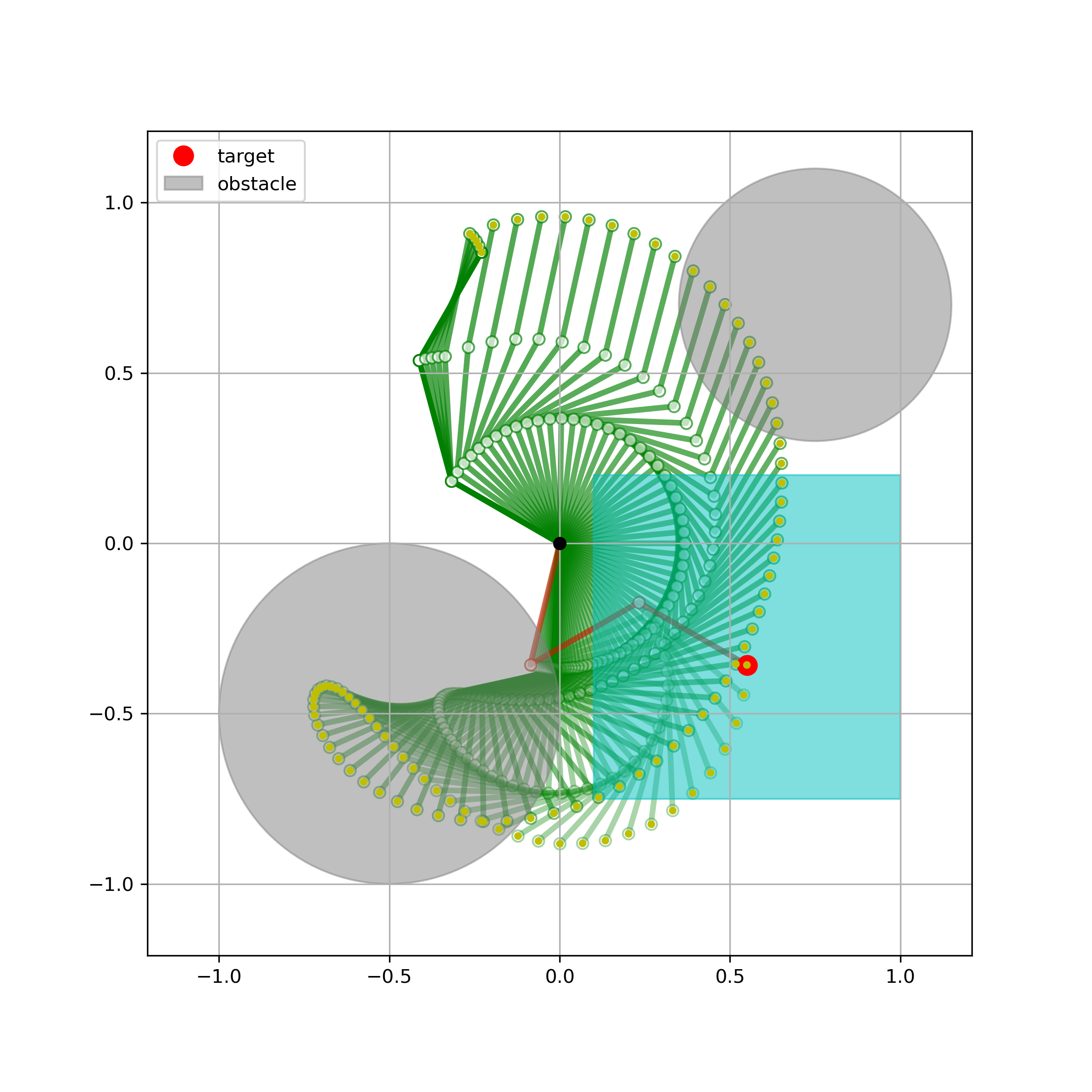}
        }
    \hfill
    \subfigure[Random Task-2]{%
        \includegraphics[width=0.23\textwidth]{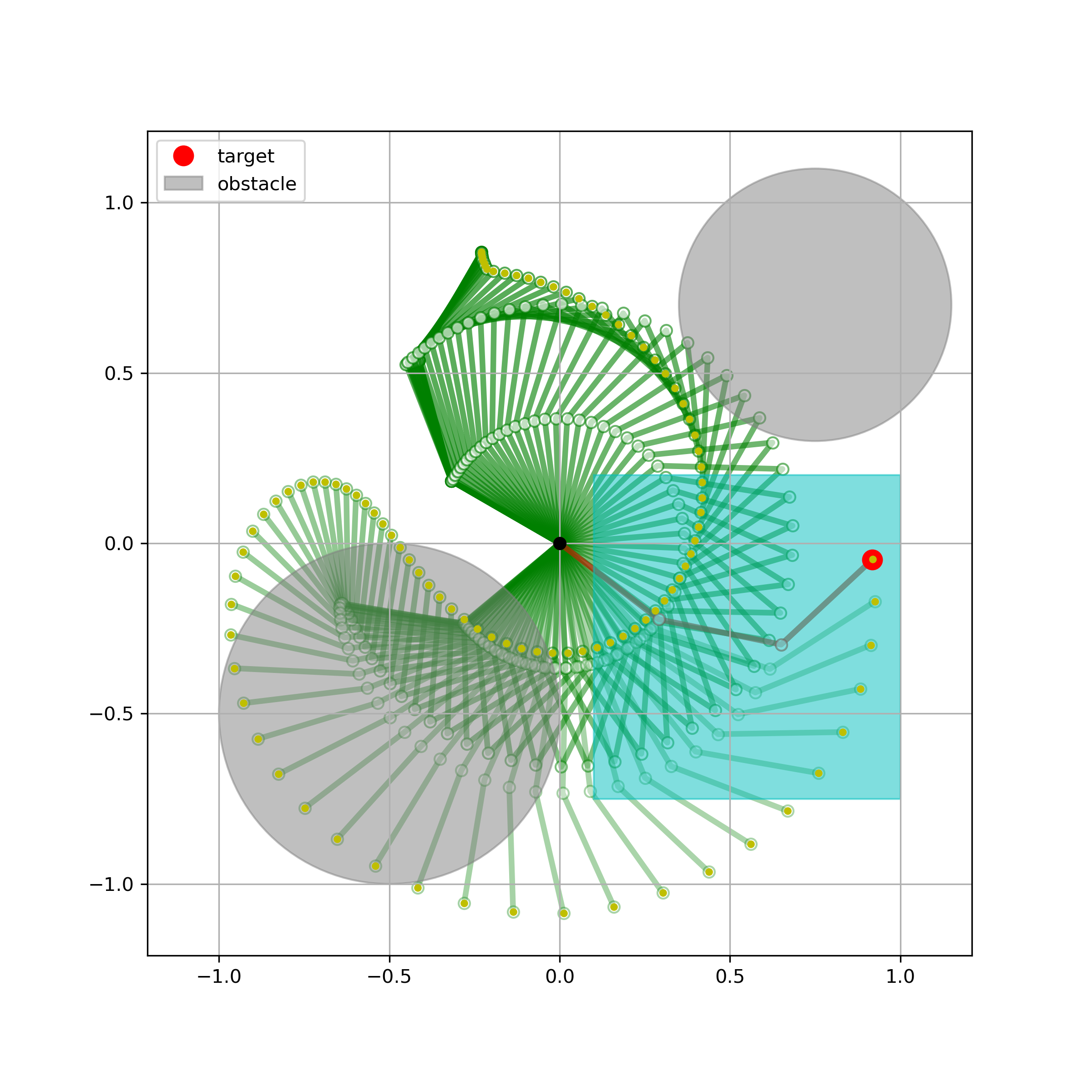}
        }
    \caption{Motion Planning of Planar Manipulators: The task is to reach a given target point in the square region depicted in cyan (task space) from a fixed initial configuration (dark green configuration). The final configuration and the joint angle trajectory to reach the target point are the decision variables. The approximate solutions from TTGO for two different tasks are given in (a) and (b) (before refinement). The solution obtained by a gradient-based solver with random initialization can result in poor local optima, as can be seen in (c) and (d).}
    \label{fig:planar_target_1}
\end{figure}

\begin{figure}[t]
    \centering
    \subfigure[]{%
        \includegraphics[width=0.22\textwidth]{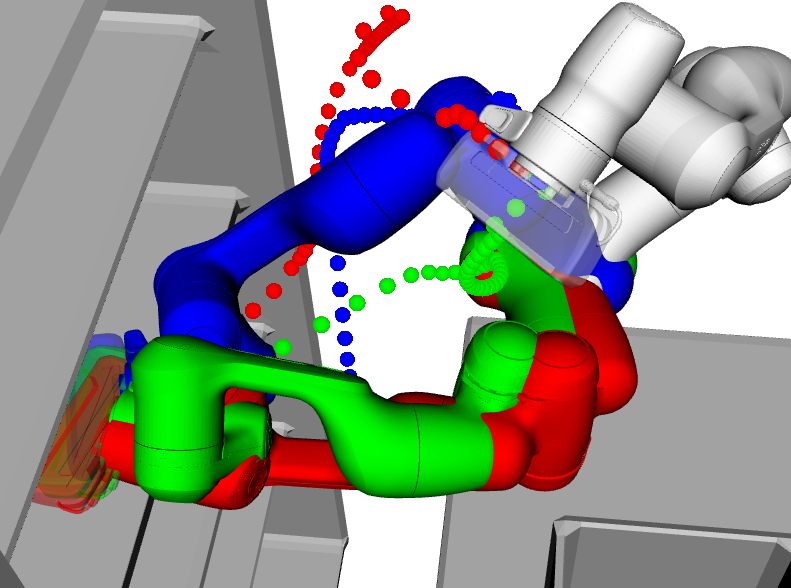}
        }
    \hfill
    \subfigure[]{%
        \includegraphics[width=0.22\textwidth]{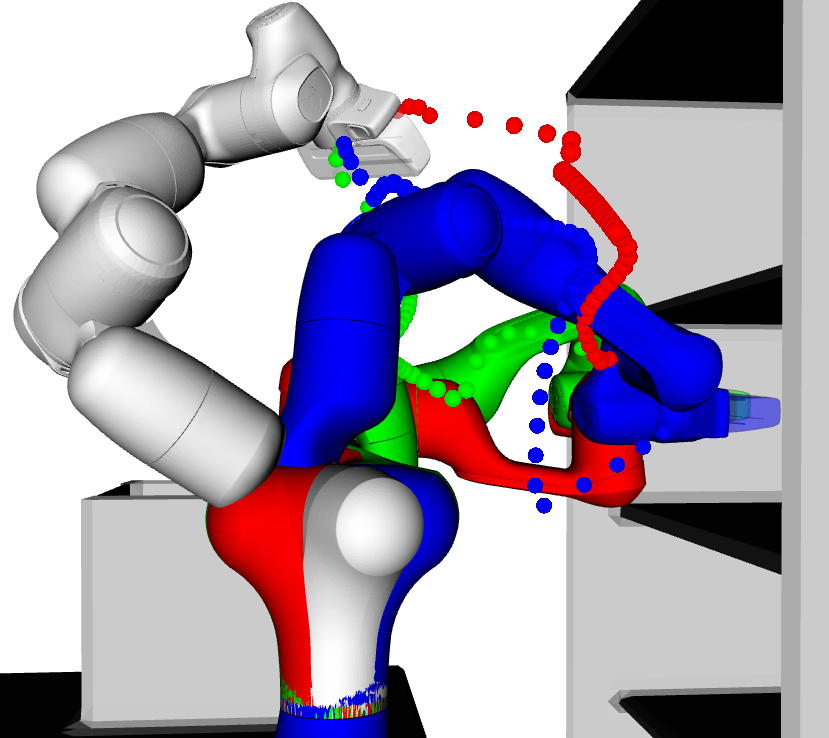}
        }
    \\
    \subfigure[]{%
        \includegraphics[width=0.22\textwidth]{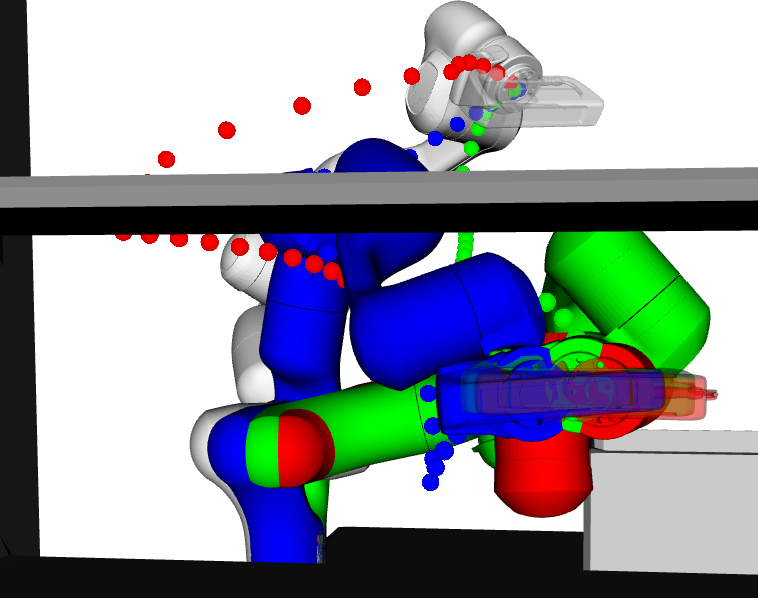}
        }
    \hfill
    \subfigure[]{%
        \includegraphics[width=0.22\textwidth]{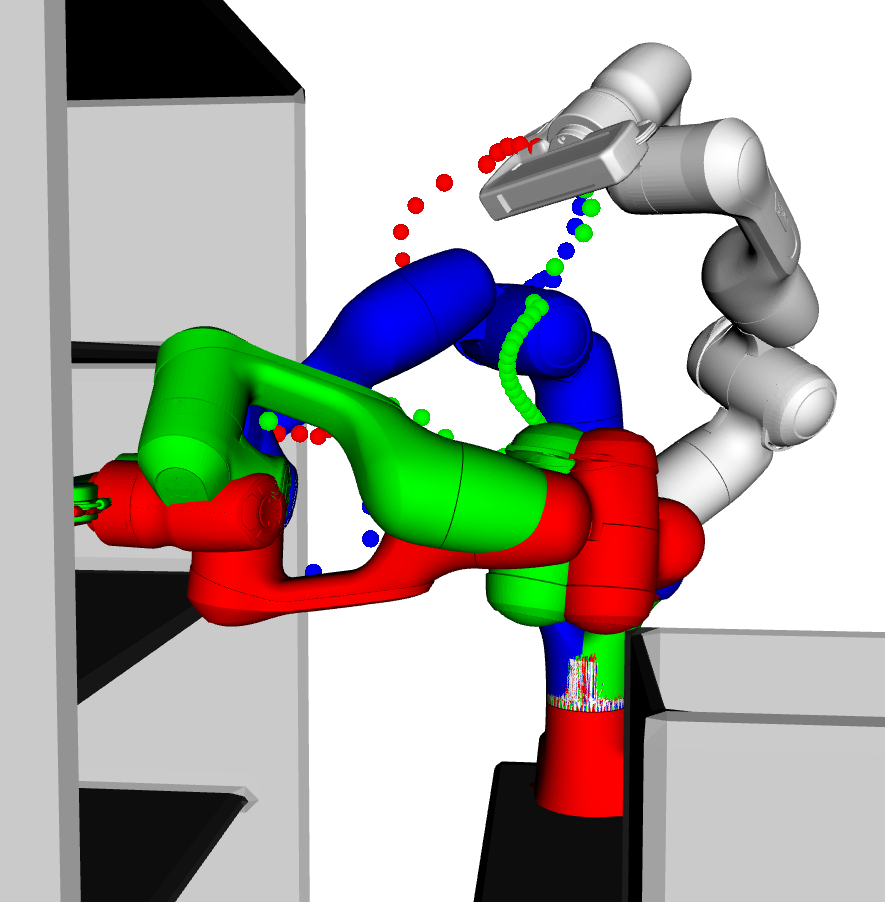}
        }
    \caption{Best $3$ out of $1000$ samples taken from a conditional TT distribution with $\alpha=0.75$ for the reaching task of a manipulator in the presence of obstacles, after refinement. The initial configuration is shown in white, while the final configuration is shown in red, green, and blue, for each solution. The end-effector path is shown by the dotted curves. The multimodality is clearly visible from these three solutions.}
    \label{fig:panda_mp_1}
\end{figure}

Figure~\ref{fig:planar_target_1} shows some examples of a reaching task for a 3-DoF planar manipulator. We can see here that the TTGO samples lead to good solutions, i.e., they avoid collisions while reaching the target quite accurately. In comparison, random sampling initialization often results in poor local optima, where the final solutions still have collisions even after the refinement. Figure~\ref{fig:panda_mp_1} shows the same reaching task for the Franka Emika robot, where the multimodality of the solutions is clearly visible. We also test the trajectory on the real robot setup as shown in Figure~\ref{fig:panda_reaching_real} and~\ref{fig:panda_pnp_real}. 

The results are presented in Table~\ref{tab: motion_reaching} and~\ref{tab: motion_via}. Similarly to the IK results, TTGO outperforms uniform sampling by a wide margin across all metrics. In reaching tasks and especially in pick-and-place tasks, uniform sampling performs quite badly in terms of success rates, since the tasks are much more difficult than the IK problem. Taking only 1 TTGO sample also does not produce satisfying performance here (i.e., $\sim 60-70 \%$) success rates, but using 10-100 samples already makes a good improvement. In pick-and-place tasks, since we consider the three different phases as a single optimization problem, it becomes quite complicated, and low values of $\alpha$ do not provide good success rates, but prioritized sampling with $\alpha=0.9$ manages to achieve $89\%$ success rates using 1000 TTGO samples. 

\begin{figure}[!ht]
    \centering
    \subfigure[]{\includegraphics[width=0.22\textwidth]{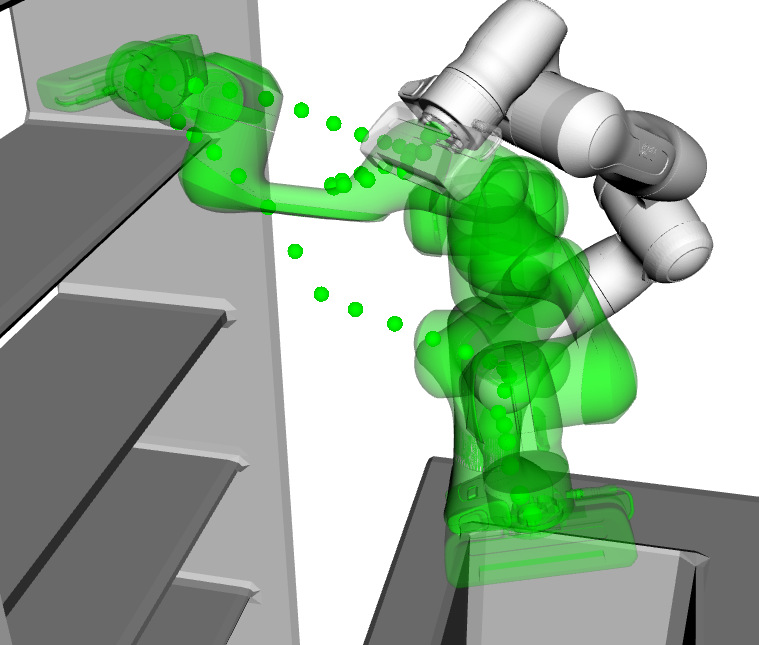}}
    \hfill
    \subfigure[]{\includegraphics[width=0.22\textwidth]{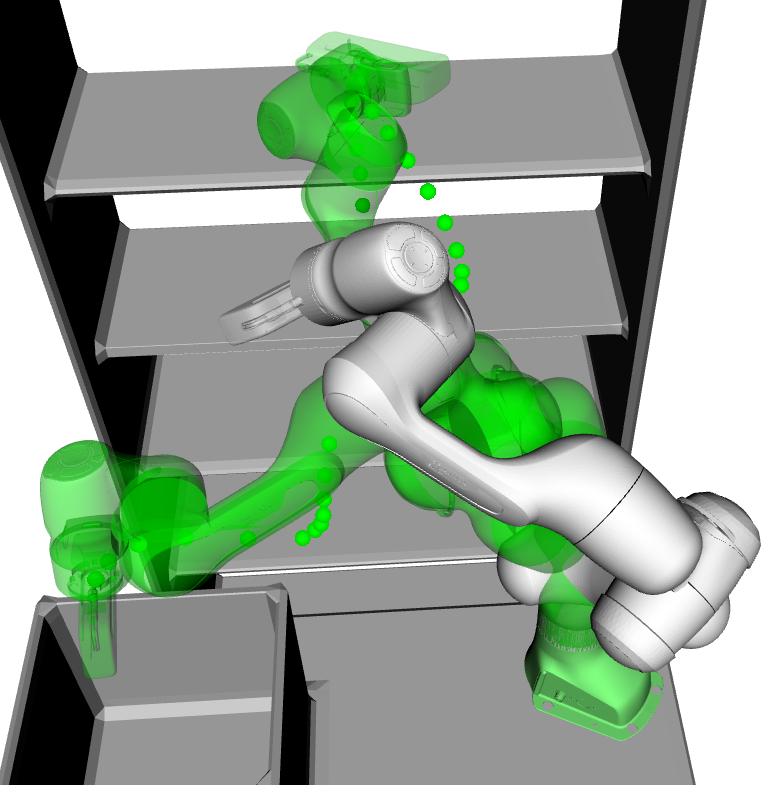}}
    \\
    \subfigure[]{\includegraphics[width=0.22\textwidth]{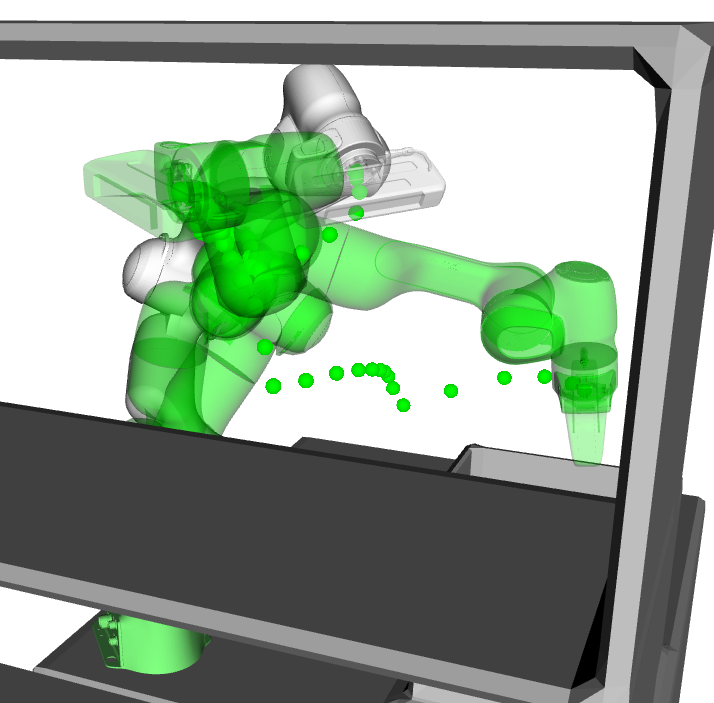}}
    \hfill
    \subfigure[]{\includegraphics[width=0.22\textwidth]{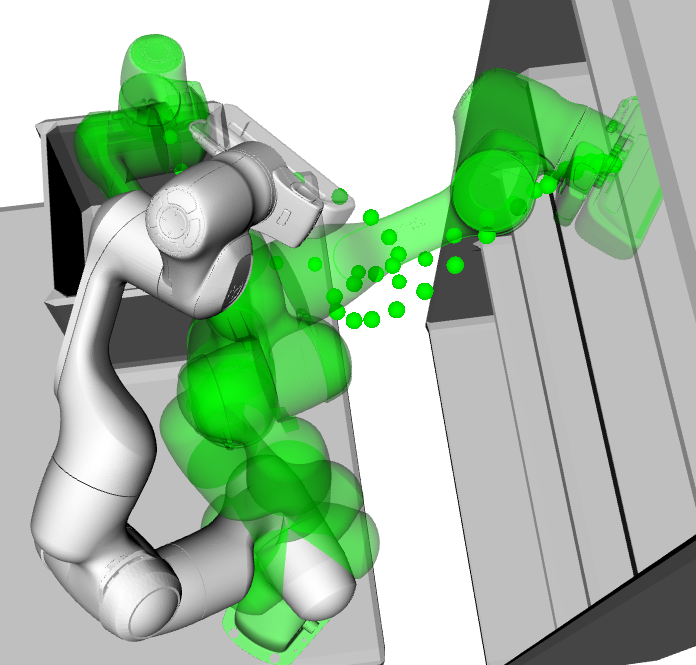}}
    \caption{A sample taken from a conditional TT distribution for the pick-and-place task, after refinement. (a) to (d) represent the same motion in different perspectives. In green, we see the picking configuration (from the shelf) and placing configuration (on the box), while the initial configuration is shown in white. The end-effector positions in the shelf and the box are the task parameters.}
    \label{fig:a_solution_for_a_}
\end{figure}

\begin{figure}[t]
    \centering
    \includegraphics[width=0.4\textwidth]{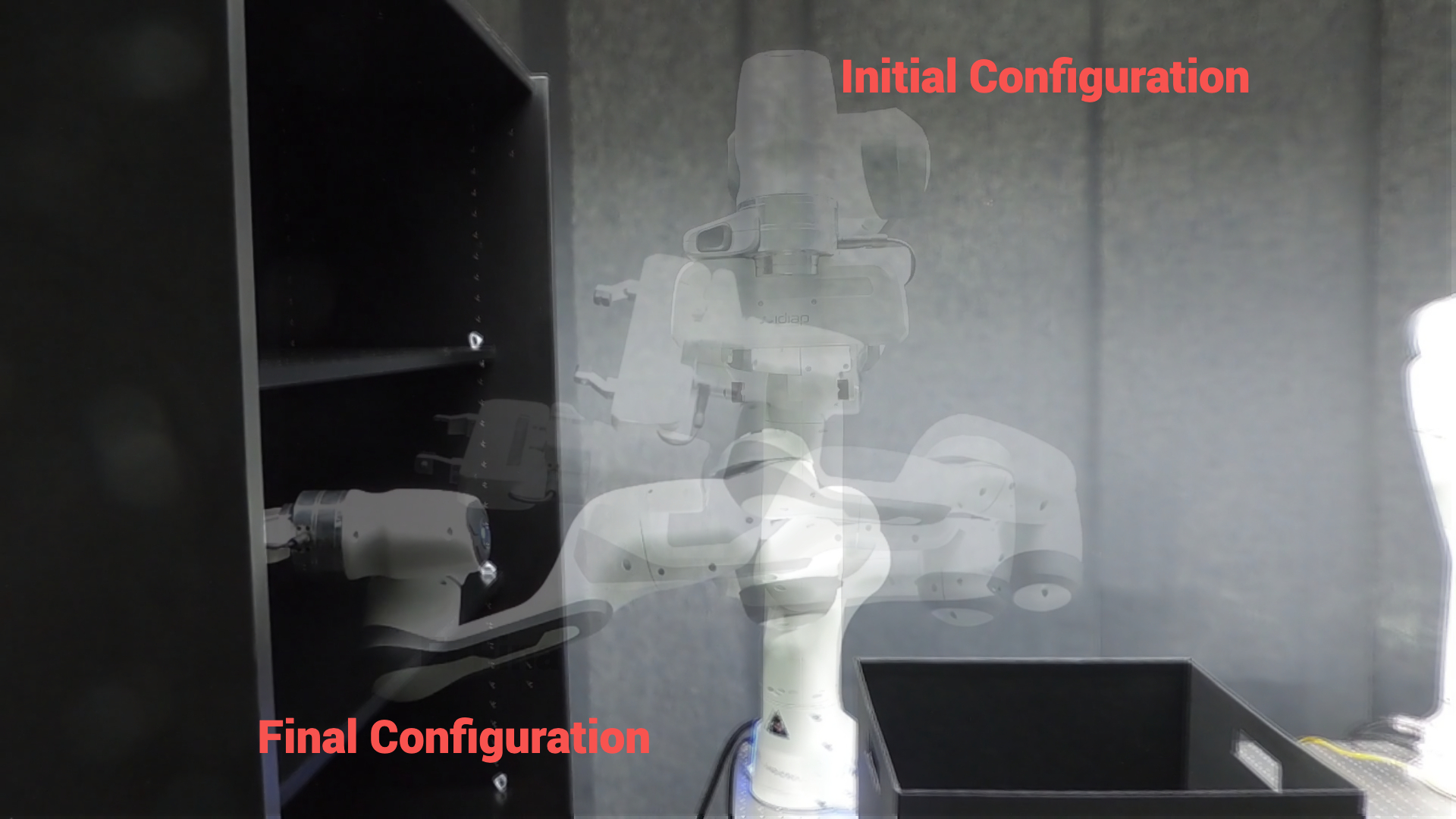}

    \caption{ The motion from the initial configuration to the final configuration in the real robot implementation of one of the TTGO solutions for the reaching task. }
    \label{fig:panda_reaching_real}
\end{figure}

\begin{figure}[t]
    \centering
    \includegraphics[width=0.4\textwidth]{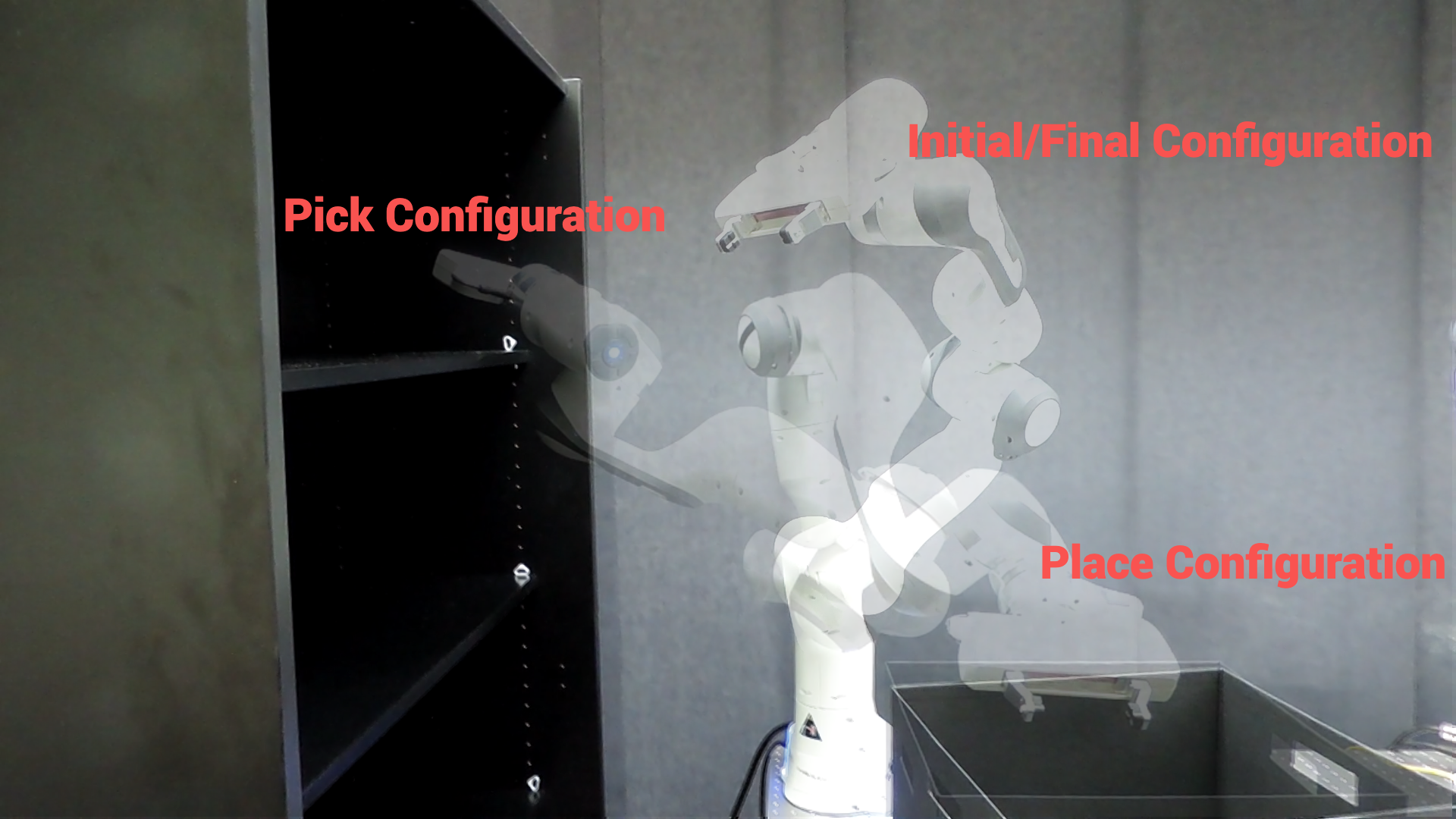}

    \caption{ Real robot implementation of one of the TTGO solutions for the pick-and-place task.  The motion from the initial configuration to the final configuration (same as the initial configuration in this case) via the  picking configuration and placing configuration is depicted.  }
    \label{fig:panda_pnp_real}
\end{figure}

\subsection{TTGO with Constant Task Parameters}
\label{contant_task}

Up to now, we have presented TTGO in its general form, where we take into account different task parameters during the training of the TT model. This approach allows us to rapidly generate approximate solutions for a specific task by conditioning the TT model. However, TTGO is also applicable when we only need to solve a single task. In this case, the TT model represents the probability distribution of only the optimization variables, and the training time is substantially shorter compared to the general case. We found that a maximum TT-rank of less than 5 works well for the applications examined in this study. In terms of computational time and solution quality, TTGO is comparable to evolutionary methods such as CMA-ES or GA, but with the added benefit of being able to provide multiple solutions.

 For such applications involving a single task parameter, our work is closely related to TTOpt~\citep{Sozykin2022_ttopt}. TTOpt is a gradient-free optimization technique based on TT-cross, which has been shown to perform comparably to evolutionary strategies. The goal of TTOpt is to maximize a reward function, which is similar to the probability density function in TTGO. TTOpt discretizes the reward function and assumes that the maximum element of the discrete approximation of the reward function closely approximates the true maximum. TT-cross is used in TTOpt to find the maximum of the discrete approximation. In this context, TT-cross is used not to build a TT approximation, but rather for its ability to identify the maximal elements of a tensor, which are likely to be in the maximum volume submatrix. This submatrix is found using the \textsc{maxvol} algorithm in TT-cross, and the maximal element of the submatrix is updated monotonically over iterations. During each iteration, the maximal element from the submatrix is stored in memory and updated until convergence. In contrast, TTGO first models the density function using TT-cross and then uses the resulting samples to approximate a solution, which is then refined using local search techniques, with the option of estimating multiple solutions.

To test the performance of TTGO as a single task optimizer, we have applied it to motion planning of both the 2-D planar robot and Franka Emika manipulator. We set the initial and the desired final configurations, and TTGO finds the trajectory to move to the final configuration while avoiding the obstacles. The joint angle trajectory is represented using the motion primitives as described in Appendix \ref{appendix_primitives}, thus the optimization variables are the weights of the basis functions. We used 2 radial basis functions for each joint.

\begin{figure*}[t]
    \centering
    \subfigure[]{%
        \includegraphics[height=0.22\textwidth]{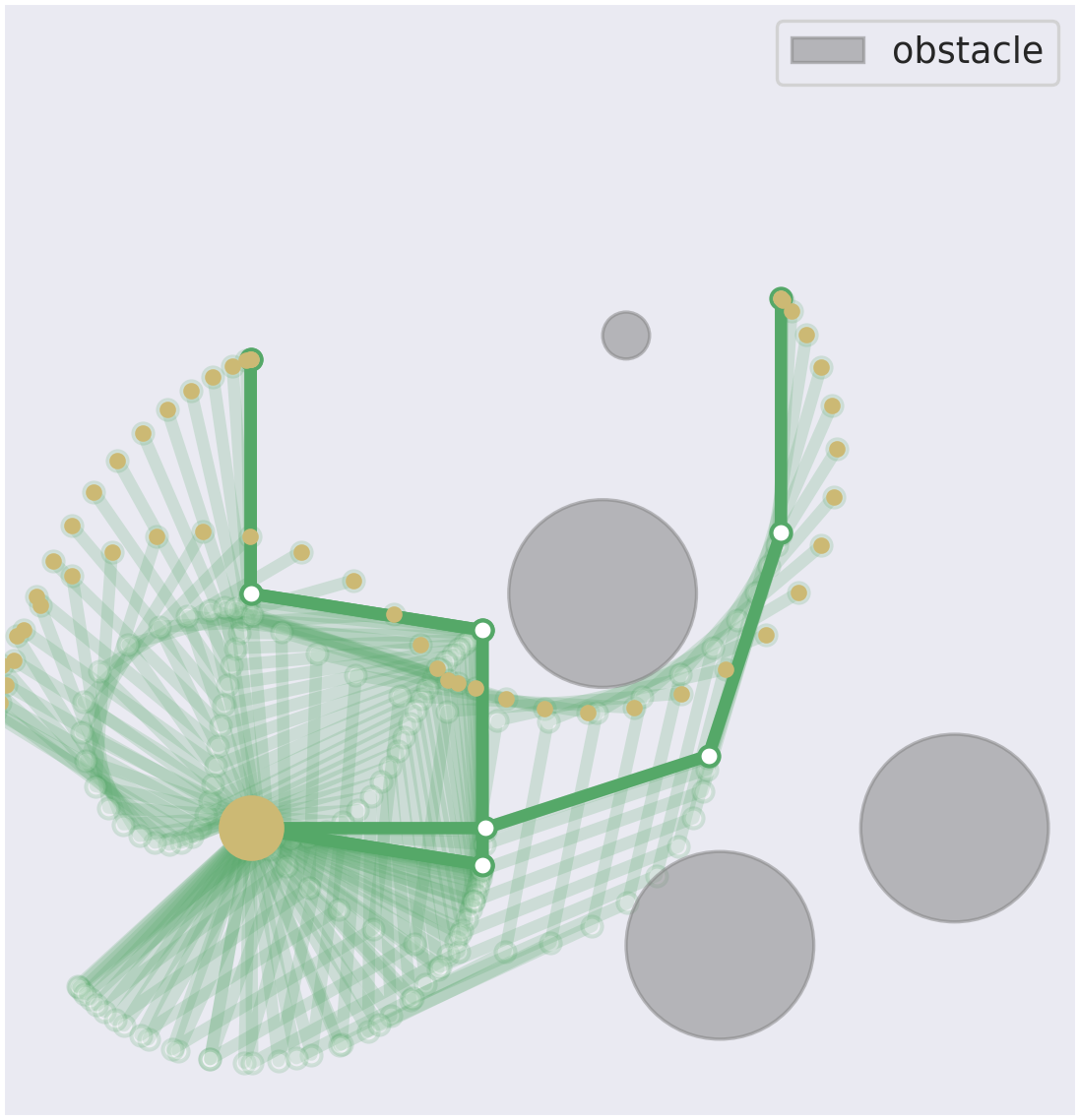}
        }
    \hfill
    \subfigure[]{%
        \includegraphics[height=0.22\textwidth]{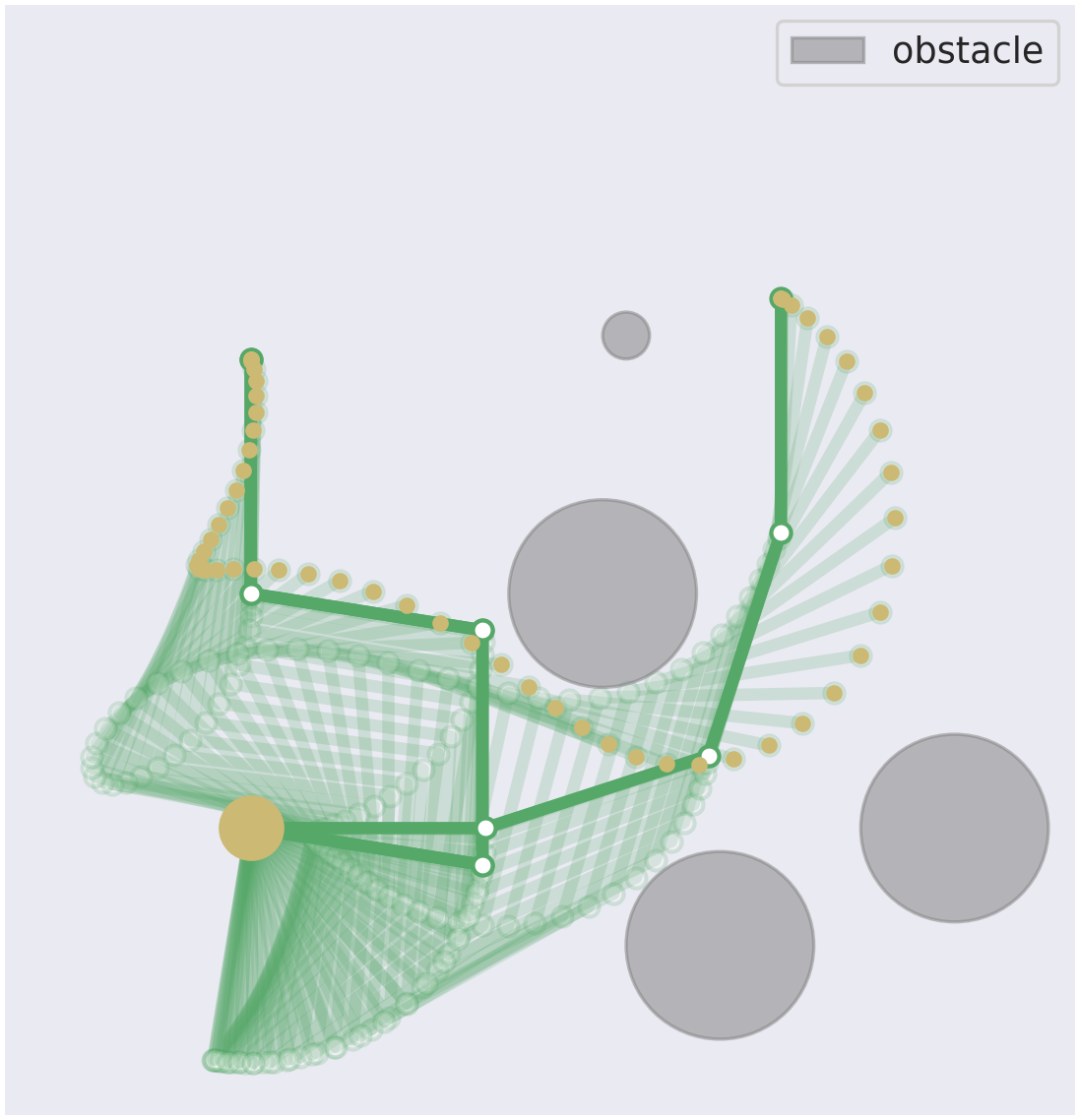}
        }
    \hfill
    \subfigure[]{%
        \includegraphics[height=0.22\textwidth]{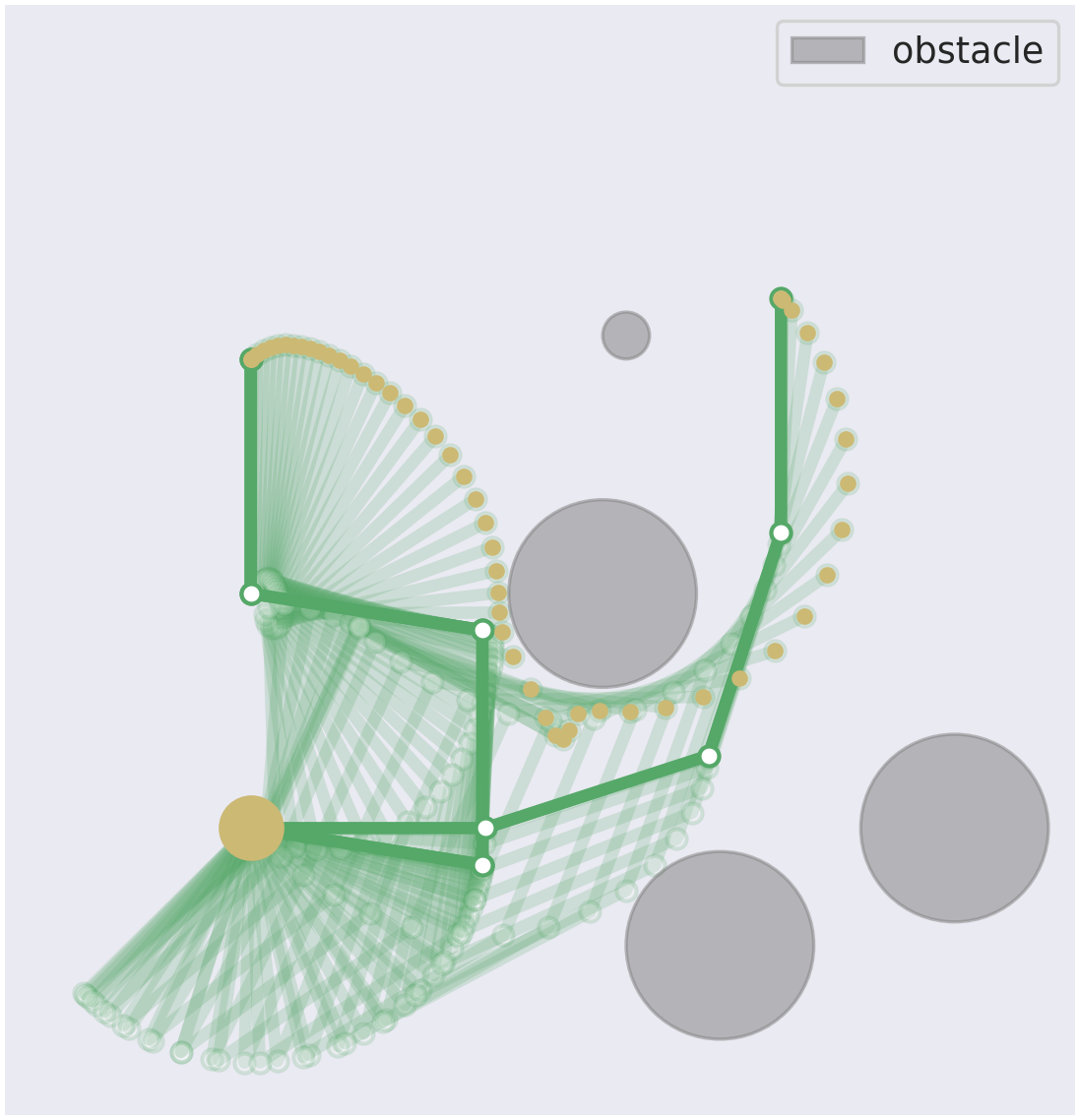}
        }
    \hfill
    \subfigure[]{%
        \includegraphics[height=0.22\textwidth]{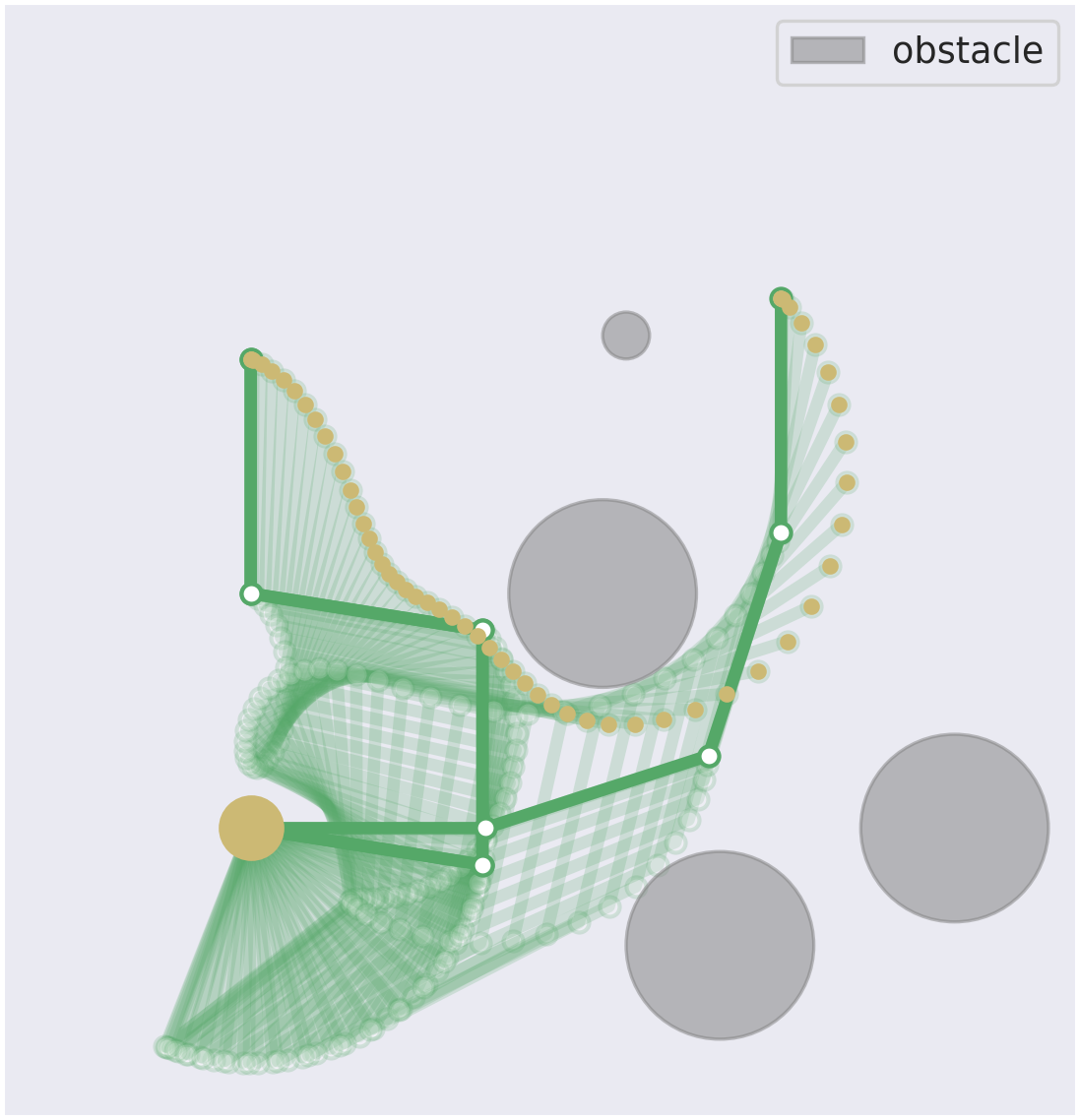}
        }
    
    \caption{Four different solutions obtained by TTGO for a  motion planning task with 4-link planar manipulator. The initial and final configuration are given (dark green) and the optimization variables are the weights of the basis functions (two basis functions per joint) that determine the joint angle trajectory.}
    \label{fig:reaching_noTask}
\end{figure*}

For the 2-D planar robot, we replicate the setting in Figure 7 of \citet{osa2020multimodal}, but we move the obstacle positions and add two more obstacles to increase the difficulty of the problem. With a fixed task parameter, the training of the TT model only takes less than 7 seconds, and we easily obtain multiple solutions. Figure~\ref{fig:reaching_noTask} shows four solutions obtained by TTGO after the refinement step. We can clearly see the multimodality of the solutions.

For the Franka Emika manipulator, we use the same setting as in Section~\ref{experiments}, i.e., with the shelf, table, and box as the collision objects. In addition, we add a cost to maintain the end-effector pose (horizontal) throughout the trajectory. The initial and final configurations are set such that both end-effector positions are located within the shelf, and they are computed using TTGO for IK, as explained in Section \ref{ik_7dof}. With this setting, we are able to obtain multiple solutions consistently for all possible scenarios (we test with different end-effector positions within the shelf) with 10 iterations of TT-cross and a maximal TT-rank of 5. With the fixed task parameter, it only takes under 5 seconds to obtain the solutions (includes TT modeling, sampling and fine tuning). Some  solutions for a given task are shown in Figure~\ref{fig:panda_no_task}. 

In comparison, SMTO~\citep{osa2020multimodal} takes $\sim 2$ minutes to solve the 2-D planar robot problem and the 7-DoF manipulator example (using their MATLAB code), whereas LSMO~\citep{osa2021motion} takes even longer, i.e., more than five minutes (according to their paper). For the 2-D example, SMTO fails to find any solution when we added more obstacles as in Figure~\ref{fig:reaching_noTask}, even after increasing the covariance by 100 times. This is because none of the initial samples from the proposal distribution is close to the feasible region. We also tried increasing the number of samples from 600 (standard value) to 2000, but it still cannot find any solution. Furthermore, adding the number of samples by $\sim 3$ times increases the computation time of SMTO by $\sim 3$ times, i.e., from $\sim 150s$ to $\sim 500s$.

\begin{figure}[!ht]
    \centering
        \subfigure[$d=7$]{%
            \includegraphics[width=0.45\linewidth]{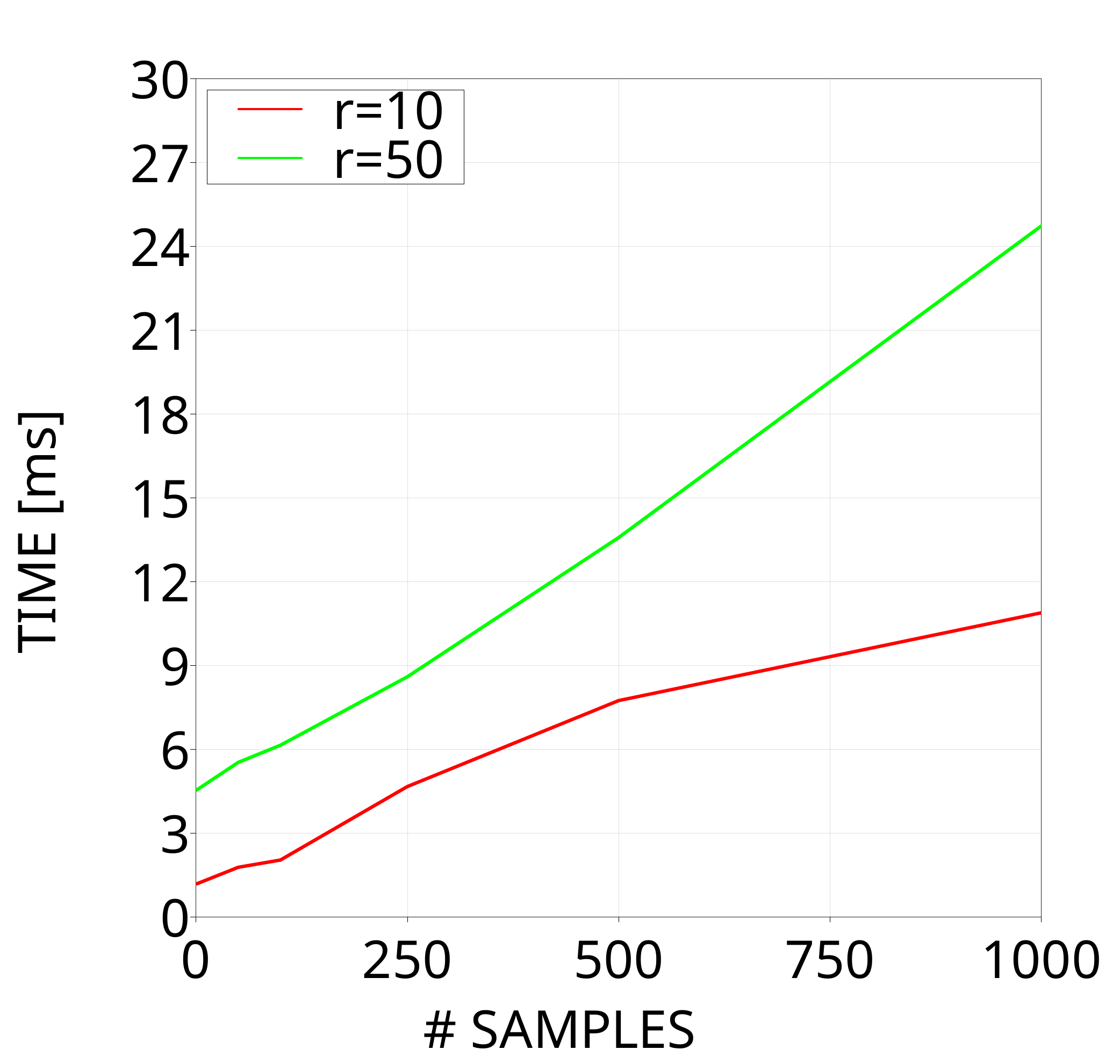}
        }
        \hfill
        \subfigure[$d=70$]{%
            \includegraphics[width=0.45\linewidth]{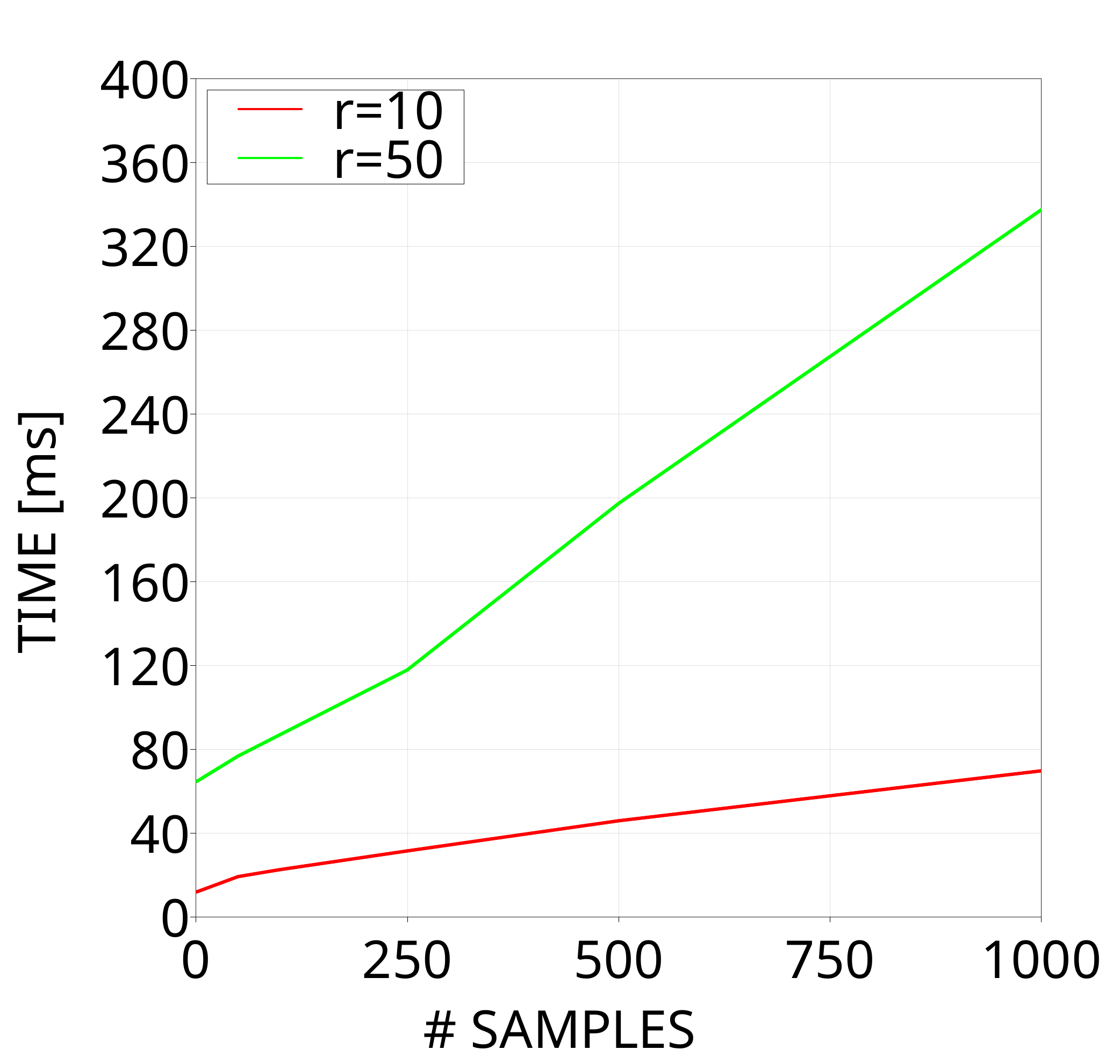}
        }
    \caption{Sampling Time: The sampling procedure has a computational complexity of $\mc{O}(ndr^2)$ and it is independent of the application. (a) and (b) show the computation time curves for two different values of $d$ with the size of each mode being $n=100$. For each figure, we show the sampling time for two different ranks as shown in red ($r=10$) and green ($r=50$).}
    \label{fig:sampling_times}
\end{figure}

\section{Discussion}
\label{discussion}

\subsection{Quality of the Approximation}
\label{quality_approx}

In this paper, we used a TT model to approximate an unnormalized PDF. The quality of the approximation highly depends on the TT-rank. If the approximation is good, the fine-tuning step is often not necessary. A nice property of TTGO that is derived from the TT-Cross method is that the model capacity can be incrementally augmented (i.e., non-parametric modeling). By increasing the number of iterations of TT-Cross and allowing a higher rank of the TT model, the approximation accuracy can be improved continuously. Furthermore, we can also use the continuous version of the TT model to allow continuous sampling. For initialization purposes, though, we found that the discrete version is enough, as the initialization does not have to be precise. 

When training the TT model, we can evaluate the quality of the approximation by picking a set of random indices, computing the value of the approximate function at those indices, and comparing it against the actual function value. This is an important evaluation for most applications that aim at finding an accurate low-rank TT decomposition of a given tensor across the whole domain. For our case, though, we are only interested in the maxima of the function, and we do not really care about the approximation accuracy in the low-density region, i.e., the region with high cost. Even if TT-Cross cannot find an accurate low-rank TT representation across the whole domain (e.g., due to non-smoothness), it can still capture the maximal elements robustly~\citep{Sozykin2022_ttopt, Goreinov2010_maxvol} as the interpolation in the TT-Cross algorithm is done using the high magnitude elements. In practice, we found that even when the approximation errors do not converge during the training, the resulting samples from the TT model are still very good as initialization.

\subsection{Comparison With Previous Work Using Variational Inference}

As described in Section~\ref{survey_mmto}, the work closest to our approach is SMTO~\citep{osa2020multimodal} that also transforms the cost function into an unnormalized PDF. SMTO uses Variational Inference to find the approximate model as a Gaussian Mixture Model (GMM) by minimizing the forward KL divergence. Its main limitation, however, is that it requires a good proposal distribution to generate the initial samples for training the model. These samples are used to find the initial GMM parameters, and subsequent iterations sample directly from the GMM. Hence, the initial samples have a large effect on the final solutions. When the initial samples do not cover some of the modes, subsequent iterations will have a very small chance of reaching those modes. We verified this by running the open-source codes provided by the author. Even for the 4-DoF manipulator example (Figure 7 in their paper), with the standard parameters given by the author, SMTO cannot find a single solution when the position of the obstacles are changed to increase the difficulty of the motion planning problem (e.g., by moving the large obstacle closer to the final configuration). It starts to find a solution only after we increased the covariance of the proposal distribution by 10-100 times the standard values, because the initial samples can then cover the region near the feasible solutions. Furthermore, when we added one more obstacle, SMTO failed to find any solution, even with the higher covariance and a larger number of samples. In comparison, we have shown in this article that TTGO can solve difficult optimization problems reliably while also providing multiple solutions. Their 4-DoF setup is in fact very similar to our planar manipulator example in Figure \ref{fig:planar_target_1}, and we have shown in Section~\ref{contant_task} that TTGO can consistently produce good solutions for different target locations. Since TTGO does not use any gradient information to find the TT model, it does not get stuck in poor local optima easily.

In~\citet{osa2021motion}, another method called LSMO was proposed to handle functions with an infinite set of solutions by learning the latent representation. As we showed in Appendix~\ref{benchmark_appendix} for sinusoidal and Rosenbrock functions, TTGO is naturally able to handle these kinds of distributions, even without any special consideration or change on the method. Unlike TTGO, SMTO and LSMO need to solve every single optimization problem from scratch. With the terminology used by TTGO, this corresponds to the task parameters being constant---a special case of the problem formulation considered so far in this paper. For such problems, since we only have a single task, the training phase in TTGO can be much faster by using a very low TT rank ($r <10$ almost always works for most optimization problems without task parameters) and fewer iterations of TT-Cross. The advantage of TTGO in such applications as compared to other global optimization approaches such as CMA-ES is that TTGO can provide multiple solutions. For example, we could find the optima of a 50D  mixture of Gaussians with $5$ components  and 30D Rosenbrock considered in Section \ref{benchmark_appendix} in less than 2 seconds. In this way, TTGO can be considered as a tool for global optimization that can offer multiple solutions. 

In this article, we proposed TTGO as a more generic tool. By anticipating and parameterizing the possible optimization problems using the task parameters, TTGO allows distribution of the computational effort into an offline and an online phase. In practice, this means that most of the computation time takes place during offline computation, while the online computation (conditioning on the TT model and sampling from it) only takes a few milliseconds. SMTO and LSMO, in comparison, take several minutes to solve a single motion planning problem for the 7-DoF manipulator case. Similarly, most trajectory optimization solvers (e.g., CHOMP, TrajOpt) and global optimization solvers (e.g., CMA-ES) can only solve a given optimization problem at each run. 



\subsection{Multimodality}
As we have shown in this paper, TTGO is able to generate samples from multiple modes consistently. Furthermore, continuing the iteration of TT-Cross will result in covering more modes as the rank of TT-model can be dynamically increased in the TT-Cross algorithm. However, unlike GMM, it is not easy to sample from only a specific mode, or to identify how many modes there are in a given problem. If we need to cluster the samples, standard clustering algorithms such as $k$-means clustering can be used.

\subsection{Further possible extensions}
\label{disc_problem_formulation}

As test cases for TTGO, we considered in this article two robotics problems that are commonly formulated as optimization problems with specific formulations of the IK and the motion planning problems. However, the proposed method is more general and can be applied to a variety of applications in robotics. For example, we can consider other choices of task parameters, e.g., by including the initial configuration, the end effector orientation, or the position of obstacle(s) as task parameters.  

TTGO approximates the joint distribution of both task parameters and decision variables, without distinguishing between the two types of variables internally. Although in this article, we conditioned the model on the given task parameters, we have the option to condition it on any subset of variables from the joint distribution. For instance, in the case of the IK problem, we can also condition it on one of the joints when we need to set a specific value for that joint. Similarly, in the pick-and-place task, we can choose to condition it only on the first target point, while treating the second target point as the decision variables. This means that we can generate optimal placement locations with respect to the cost functions and the first target point. This level of flexibility is not present in most other methods used in similar problems. Therefore, it would be interesting to explore this capability further and extend it to other robotics applications.

TTGO can be applied to other robotics problems as long as they can be expressed as optimization problems. For example, optimal control formulates the task of determining control commands as an optimization problem. Recent research has employed a database approach to warm start an optimal control solver (as explained in Section~\ref{memory_of_motion}), which could potentially benefit from the use of TTGO. It should be noted that control problems can be more demanding than planning problems since the cost function is more sensitive (i.e., slight changes in the control commands can result in significantly different state trajectories and cost values). Therefore, additional research is necessary to adapt TTGO to such problems. Moreover, certain applications, like task and motion planning~\citep{toussaint2018differentiable} or footstep planning for legged robots~\citep{deits2014footstep}, can be formulated as Mixed Integer Programming (MIP). Given that TTGO does not necessitate gradient information, combining discrete and continuous optimization can provide another compelling area for further research.

The choice of transformation used to obtain the probability function from the cost function plays an important role in TTGO. In the article, we used an  exponential function as the transformation function, however, a study on other possible transformation functions should be investigated in future work.  Moreover, in many robotics applications, the user has the flexibility to design the cost function. This will also play a role in TTGO, as smoother functions can be captured as a low rank TT model using TT-Cross with significantly lower computational cost. In the robotic applications considered in this article, we used the standard cost functions and it was non-smooth due to the cost on collision avoidance. However, a smoother cost function could still potentially be designed for such applications. This could improve the performance and the computation time reported in this article.

We used here an unsupervised approach for obtaining the TT model (and consequently the TT distribution, which captures low-cost solutions) using TT-Cross, which only requires the definition of the cost function. This approach is motivated by the fact that in various applications, it may not be feasible to access the samples (or solutions) that correspond to low cost for different task parameters. Nevertheless, if we have a repository of good solutions (i.e., optimal solutions for different possible task parameters), we can use an alternative approach with TTGO. Instead of using TT-Cross to obtain the TT model, we can use other modeling techniques like supervised learning or density estimation techniques, as described in \cite{Han2018UnsupervisedGM_densityEstimation, miller2021tensor, Novikov2021_densityEstimation}. These techniques can still capture multiple solutions, given the expressive power and generalization abilities of TT models, while also enabling quick retrieval of solutions, as explained in this article. However, obtaining a database of good solutions is a difficult challenge in robotics applications, as described in Section \ref{intro}.

Morover, TTGO has the potential to be utilized for Learning-from-Demonstration in robotics. One way to do this is by employing density modeling approaches \citep{Han2018UnsupervisedGM_densityEstimation, Novikov2021_densityEstimation} to create models of the demonstrations in TT format for different tasks. The techniques proposed in this article can then be used in the inference phase to generate a new solution for a new task.

\subsection{Limitations}
\label{limitations}
One of the major limitations of TTGO is to scale it to very high-dimensional problems. Although it has been tested up to 100 dimensions, many robotics problems involve an even greater number of dimensions. To address this issue, we utilized here motion primitive representations, which are effective for some trajectory planning applications. However, for other purposes, we may need to use nonlinear dimensionality reduction techniques such as autoencoders as a preprocessing step to determine task parameters and decision variables for TTGO. Another potential solution to this challenge is to explore the product-of-experts strategy, as presented in \cite{Pignat22IJRR}, which we plan to investigate in future work.
 
 Although constraints like joint limits can be handled naturally in TTGO, other constraints in the optimization problem need to be handled by imposing a penalty on the constraint violation in the cost function itself (i.e., formulated as soft constraints, similar to the problem formulation in evolutionary strategies and reinforcement learning). This may not be ideal for some applications in robotics that require hard constraints. However, the existing techniques for constrained optimization are mostly gradient-based, hence sensitive to initialization. Thus, we could still use TTGO for initializing such solvers. 
 
It should be noted that for TTGO to achieve fast offline computation time, it is necessary to process a batch of cost function evaluations in parallel. Without access to this parallelization capability, the time required to find the TT model using TT-Cross could become unacceptably long.

\section{Conclusion}
\label{conclusion}
In this article, we introduced TTGO, a novel framework for providing approximate solutions to optimization problems. Our evaluation of TTGO on challenging benchmark optimization functions and robotics applications (including inverse kinematics and motion planning) shows that it can produce high-quality solutions for difficult optimization problems that are often unsolvable with standard random initialization of solvers. Additionally, TTGO can provide multiple solutions from different modes, where applicable, and allows for adjustment of the sampling priority to either focus on obtaining the best solution or generating a diverse set of solutions. These features are highly beneficial in initializing optimization solvers for challenging robotics problems. Moreover, TTGO has the potential to be applied to other robotics tasks that can be formulated as optimization problems, such as task and motion planning or optimal control, and learning from demonstration. It could also serve as an alternative to mixed integer programming, which is commonly used in legged robotics and contact-rich manipulation. Future work will investigate these possibilities.

\section*{Acknowledgement}
This work was supported in part by the Swiss National Science Foundation through the LEARN-REAL project (https://learn-real.eu/, CHIST-ERA-17-ORMR-006) and by the European Commission's Horizon 2020 Programme through the MEMMO project (Memory of Motion, https://www.memmo-project.eu/, grant agreement 780684).
\bibliographystyle{SageH}
\bibliography{biblio}

\appendix
\section{Appendix}

\subsection{Separation of Variables using Matrix Factorization}
\label{separation_variables}

Consider a continuous 2D function 
\begin{equation}
    \label{eq:twoD_function}
    P(x_1,x_2)\colon \Omega_{\bm{x}} \subset \mb{R}^2 \rightarrow \mb{R}.
\end{equation}

\noindent Let $\Omega_{\bm{x}} = \Omega_{\bm{x}_1} \times \Omega_{\bm{x}_2}$ be the rectangular domain formed by the Cartesian product of intervals so that  $x_1 \in \Omega_{\bm{x}_1}$ and $x_2 \in \Omega_{\bm{x}_2}$. We can find a discrete analogue $\bm{P}$ of this function (which is a matrix in the 2D case) by evaluating the function on a grid-like discretization of the domain $\Omega_{\bm{x}}$. Let us discretize the interval $\Omega_{\bm{x}_1}$ and $\Omega_{\bm{x}_2}$ with $n_1$ and $n_2$ discretization points respectively. Let  $(x_1^1, \ldots, x_1^{n_1})$ and $(x_2^1, \ldots, x_2^{n_2})$ be the corresponding discretization points of the two intervals. The discretization set is then given by  $\mc{\bm{X}} = \{\bm{x}=(x_1^{i_1},x_2^{i_2}) \colon i_k \in \{1,\ldots,n_k \}, k \in \{1,2\} \}$ and corresponding index set is $\mc{I}_{\mc{X}} = \{ \bm{i}=(i_1,i_2)\colon i_k \in \{1,\ldots,n_k \}, k \in \{1,2\}  \}$. The corresponding discrete analogue is then given by the matrix defined as
\begin{equation}
    \label{eq: marix_analogue}
    \bm{P}_{i_1,i_2} = P(x_1^{i_1},x_2^{i_2}), \quad\forall (i_1,i_2) \in \mc{I}_{\mc{X}}.
\end{equation}

\noindent We can find a factorization of the matrix $\bm{P}$ to represent it using two factors $(\bm{P}^1, \bm{P}^2)$ with $\bm{P}^1 \in \mb{R}^{n_1 \times r}$ and $\bm{P}^2 \in \mb{R}^{r \times n_2}$  so that the elements of $\bm{P}$ can be approximated using the factors as 
\begin{equation}
    \label{eq: marix_decompose}
    \bm{P}_{i_1,i_2} \approx \bm{P}^1_{i_1,:} \; {\bm{P}^2_{:,i_2}}.
\end{equation}

The matrix factorization can be realized, for example, using  QR, SVD, or LU decompositions. Such a factorization offers several advantages: firstly, it can be used to represent the original matrix $\bm{P}$ compactly if the rank $r$ is low. Moreover, as we now show, it can be used to represent the function $P$ in a separable form. First, note that~\eqref{eq: marix_decompose} can only be used to evaluate the function $P$ at the discretized points in $\mc{\bm{X}}$. For a general $\bm{x}=(x_1,x_2) \in \Omega_{\bm{x}}$, we can use linear interpolation between the rows (or columns) and define the vector values functions
\begin{align}
    \begin{split}
    \bm{p}^1(x_1) &= \frac{x_1-x_1^{i_1}}{x_1^{i_1+1}-x_1^{i_1}}\bm{P}^1_{i_1+1,:}+\frac{x_1^{i_1+1}-x_1}{x_1^{i_1+1}-x_1^{i_1}}\bm{P}^1_{i_1,:}, \\
    \bm{p}^2(x_2) &= \frac{x_2-x_2^{i_2}}{x_2^{i_2+1}-x_2^{i_2}} \bm{P}^1_{:,i_2+1}+\frac{x_2^{i_2+1}-x_2}{x_2^{i_2+1}-x_2^{i_2}}\bm{P}^2_{:,i_2}, 
    \end{split}
    \label{eq:matrix_interpolation}
\end{align}
where $ x^{i_k}_k \leq x_k \leq x^{i_k+1}_k$, $\bm{p}^1(x_1) \colon \Omega_{\bm{x}_1} \subset \mb{R} \rightarrow \mb{R}^{1 \times r}$ and $\bm{p}^2(x_2) \colon \Omega_{\bm{x}_2} \subset \mb{R} \rightarrow \mb{R}^{r \times 1}$. Note that we could also use higher-order polynomial interpolation here.

Then, we have the approximation for the function $P$ in a separable form, 
\begin{align}
\begin{split}
    \label{eq:matrix_separability}
    P(x_1,x_2) &\approx \bm{p}^1(x_1) \bm{p}^2(x_2), \quad\forall (x_1,x_2) \in \Omega_{\bm{x}}\\
     &= \sum_{j=1}^r \bm{p}^1_j(x_1) \bm{p}^2_j(x_2).
\end{split}
\end{align}

Such a factorization of multivariate functions as a sum of product of univariate functions is an extremely powerful representation. For example, the integration of the multivariate function can be computed using integration of the univariate functions (factors)~\citep{dolgov2020_integration,Shetty21_ergodic}. If the multivariate function in hand is a probability density function, such separable representation also allows elegant sampling procedures (e.g., using conditional distribution sampling~\citep{dolgov2020_sampling}) which will be discussed in Section~\ref{tt_sample_appendix}.

In many engineering applications, we mostly deal with functions that have such separable forms. Moreover, we often have functions characterized by some smoothness improving separability. The degree of separability of the function $P$ determines a certain low-rank structure in the discrete analogue $\bm{P}$ of the function (often indicated by the number of sums in the sum of products of univariate functions representation). This implies that the rank $r$ of the factors would be low and thus the number of parameters to represent the factors is low.

The approximation accuracy of \eqref{eq:matrix_separability} also depends on the number of discretization points and on the decomposition technique that we use to find the factors. For the case of 2D functions, a common approach is to use matrix decomposition techniques such as QR, SVD or LU decomposition to find the factors. However, a standard implementation of these algorithms require the whole matrix $\bm{P}$ to be computed and stored in memory, and incurs a computational cost of $\mc{O}(n_1n_2)$. Although the resultant factors would require low memory for storage, if the discretization is very fine (i.e., $n_1$ and $n_2$ are very large numbers), computing and storing the matrix $\bm{P}$ becomes expensive and inefficient. 

A particular factorization technique called the \emph{cross approximation} (a.k.a. skeleton decomposition)~\citep{kishore2017_survey_matrix_lowrank} method avoids the above problem.
It can directly find the separable factors without having to compute and store the whole tensor in memory.  In the next section, we briefly explain the matrix cross approximation (a.k.a. skeleton decomposition) technique and some of its interesting features that are exploited in TTGO. 

\subsection{Matrix Cross Approximation}
\label{matrix_cross_approx}
\begin{figure}[t!]
    \centering
     \includegraphics[width=0.95\linewidth]{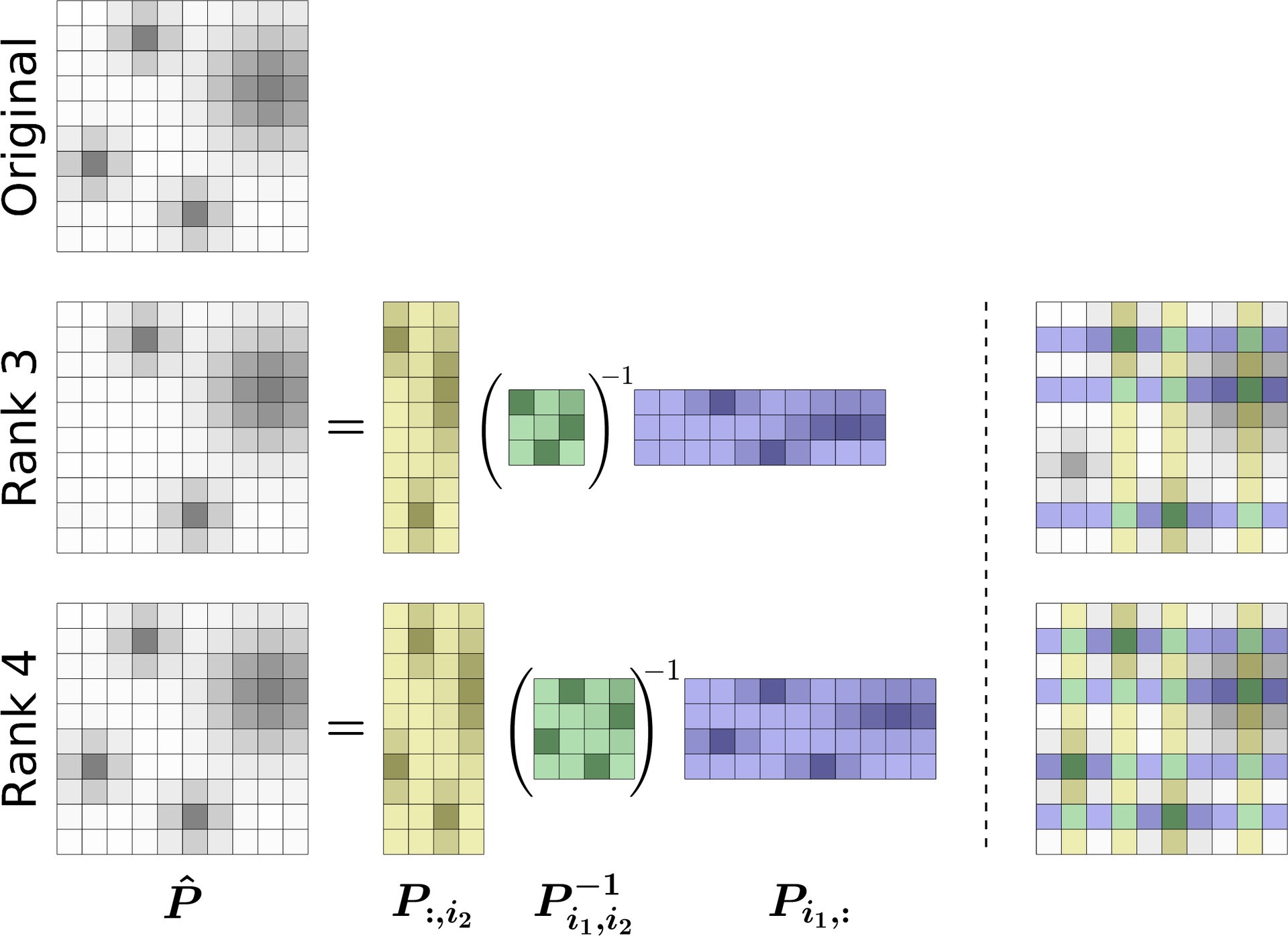}
     \caption{For a given matrix $\bm{P}$ (top-left), suppose we know $r$ independent columns indexed by $\bm{i}_2 = (i_{2,1},\ldots,i_{2,r})$, i.e., $\bm{P}_{:,\bm{i}_2} \in \mb{R}^{n_1 \times r}$ and $r$ independent rows indexed by $\bm{i}_1 = (i_{1,1},\ldots, i_{1,r})$, i.e., $\bm{P}_{\bm{i}_1,:} \in \mb{R}^{r \times n_2}$, with their intersection $\bm{P}_{\bm{i}_1,\bm{i}_2} \in \mb{R}^{r \times r}$ being nonsingular. Then, by skeleton decomposition we have $\bm{\hat{P}} = \bm{P}_{:,\bm{i}_2} \bm{P}_{\bm{i}_1,\bm{i}_2}^{-1} \bm{P}_{\bm{i}_1,:}$. If $\text{rank}(\bm{P})=r$, then $\bm{\hat{P}} = \bm{P}$ (bottom row). For $r<\text{rank}(\bm{P})$ we obtain a quasi-optimal approximation, $\bm{\hat{P}} \approx \bm{P}$ (middle row). The right figures show the rows and columns selected from the original matrix by the cross approximation algorithm to find the skeleton decomposition.}
     \label{fig:cur}
\end{figure}

Suppose we have a rank-$r$ matrix $\bm{P} \in \mb{R}^{n_1 \times n_2}$. Using cross-approximation (a.k.a.~CUR decomposition or skeleton decomposition), this matrix can be exactly recovered using $r$ independent rows (given by the index vector $\bm{i}_1 \subset \{1,\ldots,n_1 \}$) and $r$ independent columns (given by the index vector $\bm{i}_2 \subset \{1,\ldots,n_2\}$) of the matrix $\bm{P}$ as
$$ \bm{\hat{P}} = \bm{P}_{:,\bm{i}_2} \; \bm{P}^{-1}_{\bm{i}_1,\bm{i}_2} \; \bm{P}_{\bm{i}_1,:},$$
provided the intersection matrix $\bm{P}_{\bm{i}_1,\bm{i}_2}$ (called submatrix) is non-singular. Thus, the matrix $\bm{P}$, which has $n_1 n_2$ elements, can be reconstructed using only $(n_1+n_2-r)r$ of its elements (see Figure~\ref{fig:cur}).

Now suppose we have a noisy version of the matrix $\bm{P} = \bm{\Tilde{P}} + \bm{E}$ with $\|\bm{E}\| < \epsilon$ and $\bm{\Tilde{P}}$ is of low rank. For a sufficiently small $\epsilon$, $\text{rank}(\bm{\Tilde{P}})=r$ so that the matrix $\bm{P}$ can be approximated with a lower rank $r$ (i.e., $\text{rank}(\bm{P}) \approx r$). Then, the choice of the submatrix $\bm{P}_{\bm{i}_1,\bm{i}_2}$ (or index vectors $\bm{i}_1,\bm{i}_2$) for the cross approximation requires several considerations. The maximum volume principle can be used in choosing the submatrix which states that the submatrix with maximum absolute value of the determinant is the optimal choice. If $\bm{P}_{\bm{i}_1^*,\bm{i}_2^*}$ is chosen to have the maximum volume, then by skeleton decomposition we have an approximation of the matrix $\bm{P}$ given by $\bm{\hat{P}} = \bm{P}_{:,\bm{i}_2^*} \bm{P}^{-1}_{\bm{i}_1^*,\bm{i}_2^*} \bm{P}_{\bm{i}_1^*,:}$. This results in a quasi-optimal approximation: 
$$ \| \bm{P} - \bm{\hat{P}} \|_2 < (r+1)^2 \; \sigma_{r+1}(\bm{P}), $$
where $\sigma_{r+1}(\bm{P})$ is the $(r+1)$-th singular value of $\bm{P}$ (i.e., the approximation error in the best rank $r$ approximation in the spectral norm). Thus, we have an upper bound on the error incurred in the approximation which is slightly higher than the best rank $r$ approximation (Eckart–Young–Mirsky theorem). 

Finding the maximum volume submatrix is, however, an NP-hard problem. However,  many heuristic algorithms that work well exist in practice by using a submatrix with a sufficiently large volume, trading off the approximation accuracy for the computation speed. One of the widely used methods is the \textsc{maxvol} algorithm~\citep{Goreinov2010_maxvol} which can provide, given a tall matrix $\bm{P} \in \mb{R}^{r\times n_2}$ (or $\mb{R}^{n_1 \times r}$), the maximum volume submatrix $\bm{P}_{\bm{i}_1^*,\bm{i}_2^*} \in \mb{R}^{r \times r}$. The cross approximation algorithm uses the \textsc{maxvol} algorithm in an iterative fashion to find the skeleton decomposition as follows:
\begin{enumerate}
    \item \emph{Input}: $\bm{P} \in \mb{R}^{n_1 \times n_2}$, the approximation rank $r$ for the skeleton decomposition.
    \item Find the columns index set $\bm{i}_2^*$ and the row index set $\bm{i}_1^*$ corresponding to the maximum volume submatrix.
    \begin{enumerate}
        \item Randomly choose $r$ columns $\bm{i}_2$ of the matrix $\bm{P}$ and repeat the following until convergence:
        \begin{itemize}
        \item Use \textsc{maxvol} to find $r$ row indices $\bm{i}_1$ so that $\bm{P}_{\bm{i}_1,\bm{i}_2}$ is the submatrix with maximum volume in $\bm{P}_{:,\bm{i}_2}$.
        \item Use \textsc{maxvol} to find $r$ column indices $\bm{i}_2$ so that $\bm{P}_{\bm{i}_1,\bm{i}_2}$ is the submatrix with maximum volume in $\bm{P}_{\bm{i}_1,:}$.
        \end{itemize}
    \end{enumerate}
    \item \emph{Output}: Using the column index set $\bm{i}_2^*$ and the row-index set $\bm{i}_1^*$ corresponding to the maximum volume submatrix, we have the skeleton decomposition $\bm{\hat{P}} \approx \bm{P}_{:,\bm{i}_2^*} \bm{P}^{-1}_{\bm{i}_1^*,\bm{i}_2^*}
    \bm{P}_{\bm{i}_1^*,:}$.  
\end{enumerate}

In the above algorithm, during the iterations the matrices $\bm{P}_{:,\bm{i}_2}$ (or $\bm{P}_{\bm{i}_1,:}$) might be non-singular. Thus, a more practical implementation uses the $QR$ decomposition of these matrices and the \textsc{maxvol} algorithm is applied to the corresponding $Q$ factor to find the columns (or rows) of the submatrix. Furthermore, instead of a random choice in step (2.1), one can choose the $r$ columns from the multinomial distribution given by $p(i_2) = \frac{\|\bm{P}_{:,i_2}\|}{\|\bm{P}\|}, i_2 \in \{1,\ldots,n_1\}$ without sample replacement.

Note that, in the above algorithm, the input is only a function to evaluate the elements of the matrix $\bm{P}$ (i.e., we do not need the whole matrix $\bm{P}$ in computer memory). Some features of cross approximation algorithms are highlighted below:
\begin{itemize}
    \item The factors in a cross approximation method consist of elements of the actual data (rows and columns) of the original matrix and hence it improves interpretability. For example, SVD does projection onto the eigenvectors which could be abstract, whereas cross approximation does projection onto the vectors formed by rows and columns of the actual data of the matrix which are more meaningful.
    \item Since cross approximation algorithms follow the maximum volume principle, the factors are composed of \emph{high magnitude} elements of the original matrix with high probability~\citep{Goreinov2010_maxvol}. This is very useful for TTGO as we are interested in finding the maxima from a tensor (discrete analogue of a probability density function) and the skeleton decomposition preserves this information.
    \item Cross approximation algorithms directly find the factors without computing and storing the whole matrix. 
\end{itemize}

\subsection{Interpolation of Tensor Cores}
\label{tensor_core_interpolation_appendix}
Given the discrete analogue tensor $\bm{\mc{P}}$ of a function $P$, we can obtain the continuous approximation by interpolating the TT cores, in a similar way as in the matrix case in Section~\ref{separation_variables}. For example, we can use a linear interpolation for each core (i.e., between the matrix slices of the core) and define a matrix-valued function corresponding to each core $k \in \{1,\ldots, d\}$,
\begin{equation}
    \bm{P}^k(x_k) = \frac{x_k-x^{i_k}_k}{x^{i_k+1}_k-x^{i_k}_k}\bm{\mc{P}}^k_{:,i_k+1,:} +\frac{x^{i_k+1}_k-x_k}{x^{i_k+1}_k-x^{i_k}_k}\bm{\mc{P}}^k_{:,i_k,:},
\end{equation}
where $x^{i_k}_k\le x_k \le x^{i_k+1}_k$ and $\bm{P}^k: \Omega_{x_k} \subset \mb{R} \rightarrow \mb{R}^{r_{k-1} \times r_k}$ with $r_0=r_d=1$. This induces a continuous approximation of $P$ given by
\begin{equation}
    \label{eq:continuous_tt}
    P(x_1,\ldots,x_d) \approx \bm{P}^1(x_1) \cdots \bm{P}^d(x_d).
\end{equation}

\noindent Note that a higher-order polynomial interpolation can also be used if needed.

\subsection{Sampling from TT distribution}
\label{tt_sample_appendix}
Consider a probability distribution given by~\eqref{eq:tt_distribution}. For the simplicity of the presentation, we assume $Z=1$ as we will not require the normalization constant to be known for sampling from the above distribution. The distribution can be expressed as a product of conditional distributions
\begin{multline*}
        \text{Pr}(x_1,\ldots,x_d) = \text{Pr}_1(x_1)\text{Pr}_2(x_2|x_1) \cdots \\ \cdots \text{Pr}_d(x_d|x_1,\ldots,x_{d-1}),
\end{multline*}

\noindent where
\begin{equation*}
    \text{Pr}_k(x_k|x_1,\ldots,x_{k-1}) = \frac{\sigma_k(x_1,\ldots,x_k)}{\sigma_{k-1}(x_1,\ldots,x_{k-1})}    
\end{equation*}
is the conditional distribution defined using the marginals
\begin{equation*}
    \sigma_k(x_1,\ldots,x_k) = \sum_{x_{k+1}}\cdots \sum_{x_d} \text{Pr}(x_1,\ldots,x_d).    
\end{equation*}

Let $\sigma_0 = 1$. Now, using the above definitions, we can generate samples $\bm{x}\sim \text{Pr} $ by sampling from each of the conditional distributions in turn. Each conditional distribution is a function of only one variable, and in the discrete case it is a multinomial distribution, with
\begin{equation*}
    x_k \sim \text{Pr}_k(x_k|x_1,\ldots,x_{k-1}), \text{ } \forall k \in \{1,\ldots,d\}.
\end{equation*}
  
However, this is computationally intensive as sampling $x_k$ requires the conditional distribution $\text{Pr}_k$ which in turn requires the evaluation of the summation over several variables to find the marginal $\sigma_k$. It results in a computational cost that grows exponentially with the number of dimensions. This is where the TT format provides a nice solution by relying on the separability of the function. Let $\bm{\mc{P}}$ be the discrete analogue of the function $\text{Pr}$ (or $P$ as $Z=1$), a tensor in TT format,  with the discretization set $\mc{X}$. Let the TT model be given by the cores $(\bm{\mc{P}}^1,\ldots,\bm{\mc{P}}^d)$, then we have
\begin{align}
\footnotesize
\begin{split}
\sigma_k(x_1,&\ldots,x_k) = \sum_{x_{k+1}} \cdots \sum_{x_{d}} \text{Pr}(\bm{x}), \\
&\approx  \sum_{x_{k+1}}  \cdots \sum_{x_d} \bm{\mc{P}}_{\bm{x}} , \\
&= \sum_{x_{k+1}} \cdots \sum_{x_{d}} \bm{\mc{P}}^1_{:,x_1,:} \cdots \bm{\mc{P}}^k_{:,x_k,:} \bm{\mc{P}}^{k+1}_{:,x_{k+1},:} \cdots \bm{\mc{P}}^d_{:,x_d,:},\\
&=  \bm{\mc{P}}^1_{:,x_1,:} \cdots \bm{\mc{P}}^k_{:,x_k,:}\Big(\sum_{x_{k+1}} \bm{\mc{P}}^{k+1}_{:,x_{k+1},:}\Big) \cdots \Big(\sum_{x_{d}} \bm{\mc{P}}^d_{:,x_d,:}\Big),
\end{split}
\label{eq:marginal}
\end{align}
where $\sum_{x_k}\bm{\mc{P}}^{k}_{:,x_{k},:}$ is the summation of all the matrix slices of the third-order tensor (cores of TT). Thus, the TT-format reduces the complicated summation into one-dimensional summations. When the same summation terms appear over several conditionals $\text{Pr}_k$, we can use an algorithm called Tensor Train Conditional Distribution (TT-CD) sampling ~\citep{dolgov2020_sampling}, to efficiently get the samples from $\text{Pr}$. 

\subsubsection{Prioritized Sampling}
We can choose a hyperparameter $\alpha \in (0,1)$ to prioritize samples from higher-density regions in the distribution $\text{Pr}(\bm{x})$ given by \eqref{tt_distribution}. $\alpha=0$ leads to generating exact samples from the true TT distribution whereas $\alpha=1$ leads to sampling from regions closer to the mode of the distribution. Values of $\alpha$ higher than $0$ reduce the likelihood of generating samples from low-density regions of the TT distribution. This algorithm is described in  Algorithm \ref{alg:tt_cd}. The prioritized sampling can be relaxed by setting $\alpha=0$ in the algorithm, resulting in the standard sampling procedure described in Section \ref{tt_sample_appendix}. Note that the algorithm allows a parallel implementation to quickly generate a large number of samples.

Alternatively, instead of using prioritized samples with a specified $\alpha$ value, we can choose to use a deterministic version of the algorithm. This involves selecting the top $N$ elements from the multinomial distribution $\bm{p}_k$ at iteration $k$ instead of sampling $N$ elements independently and keeping track of the history of the selected indices from the previous modes ($k-1$). This approach is suitable when the approximation error in the TT model is low and we need only one solution. Both algorithms are available in the software accompanying this article. In this article, the results are demonstrated using the stochastic version as we are interested in multiple solutions for a given optimization problem.

\begin{algorithm}[t] 
\caption{ TT-CD Sampling with Sample Prioritization}
\label{alg:tt_cd}
\begin{algorithmic}[1]
\Require{TT Blocks $\bm{\mc{P}}=(\bm{\mc{P}}^1,\ldots,\bm{\mc{P}}^d)$ corresponding to the distribution $\text{Pr}$, sample priority $\alpha \in (0,1)$ } 
\Ensure{$N$ $\alpha-$prioritized samples  $\{(x^l_1,\ldots, x^l_d)\}_{l=1}^N$ from the distribution $\text{Pr}$}

\State {$\bm{\hat{\pi}}_{d+1}$ $\gets$  $1$}
\For{$k \gets d$ to $2$}                    
    \State {$\bm{\hat{\pi}}_k = (\sum_{x_k}\bm{\mc{P}}^{k}_{:,x_{k},:}) \bm{\hat{\pi}}_{k+1}$}
\EndFor
\State {$\bm{\Phi_1} \gets \bm{1}\in \mb{R}^{N \times 1}$}
\For{$k \gets 1$ to $d$}                    
    \State {$\bm{\pi}_k({x_k}) = \bm{\mc{P}}^k_{:,x_k,:} \bm{\hat{\pi}}_{k+1}, \forall x_k$}
    \For{$l = 1,\ldots,N$}
    \State{$\bm{p}_k(x_k) = |\Phi_k(l,:)\bm{\pi}_k(x_k)|, \forall x_k$}
    \State{$\bm{p}_k \leftarrow \frac{\bm{p}_k}{\max{\bm{p}_k}} $}
    \State{$\bm{p}_k \leftarrow \bm{p}_k^{\frac{1}{1-\alpha +\epsilon}}$, where $\epsilon$ is positive and $\epsilon \approx 0$}
    \State{$\bm{p}_k(x_k) \gets \frac{\bm{p}_k(x_k)}{\sum_{x_k} \bm{p}_k(x_k)}, \forall x_k$  } (normalization) 
    \State{Sample $x^l_k$ from the multinomial distribution $\bm{p}_k$}
    \State{$\bm{\Phi}_{k+1}(l,:) = \bm{\Phi}_k(l,:) \bm{\mc{P}}^k_{:,x^l_k,:}$}
    \EndFor
\EndFor

\end{algorithmic}
\end{algorithm}




\subsection{Benchmark functions}
\label{benchmark_appendix}
We apply our framework to extended versions of some benchmark functions for numerical optimization techniques, i.e., Rosenbrock and Himmelblau functions. They are known to be notoriously difficult for gradient-based optimization techniques to find the global optima, which could be more than one.
Some of the functions also have some parameters that can change the shape of the functions. 
We consider these parameters as the task parameters, hence making the problem even more challenging. The benchmark functions are considered as the cost functions and we transform them to obtain a suitable probability density function. In addition, we also include a sinusoidal function to show that TTGO can handle a cost function with an infinite number of global optima, and a mixture of Gaussians to test the performance of TTGO on a high-dimensional multimodal function.

\begin{figure*}[t]
    \centering
    \subfigure[$\alpha=0$]{%
        \includegraphics[height=0.22\textwidth]{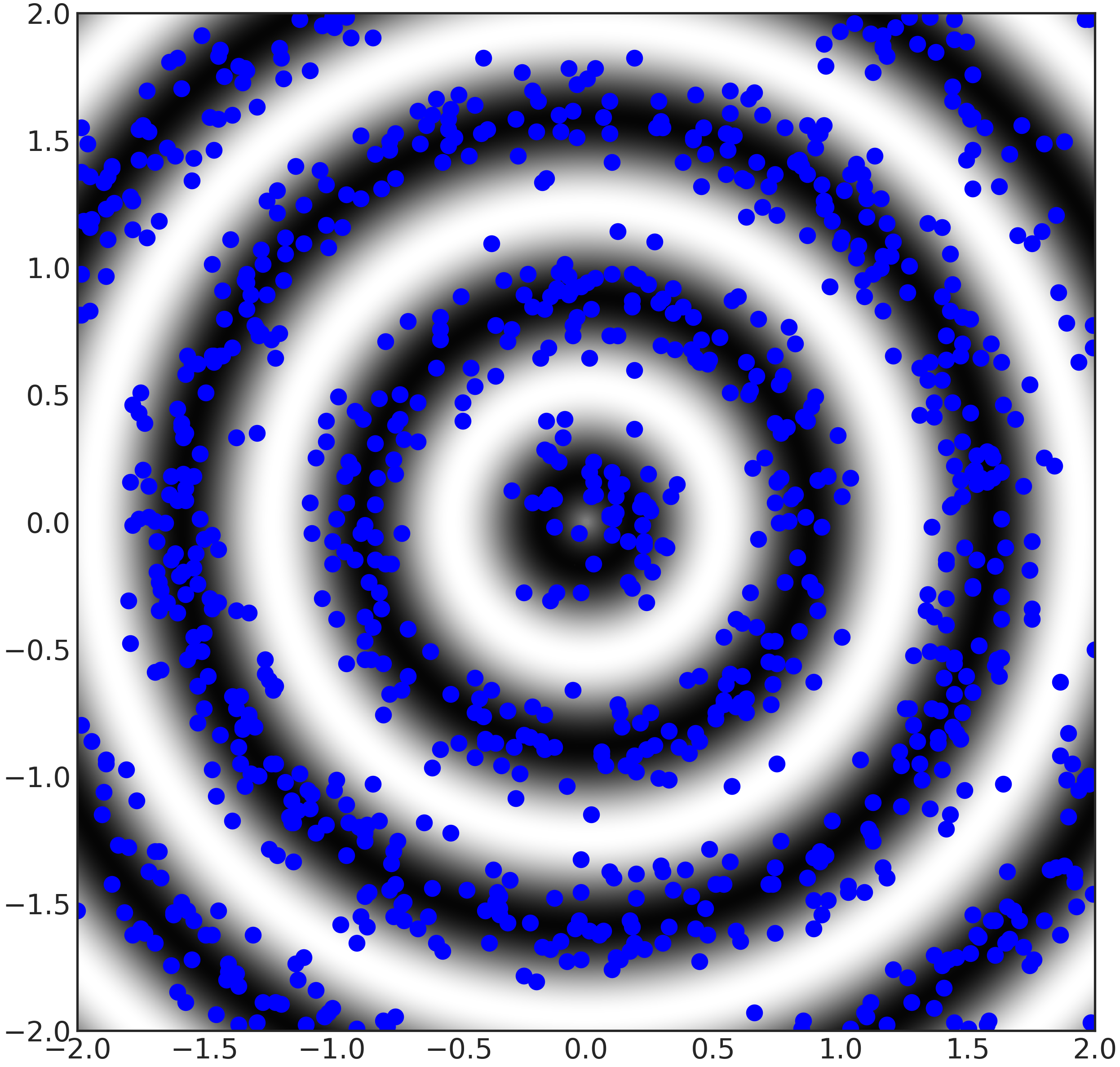}
        }
    \hfill
    \subfigure[$\alpha=0.5$]{%
        \includegraphics[height=0.22\textwidth]{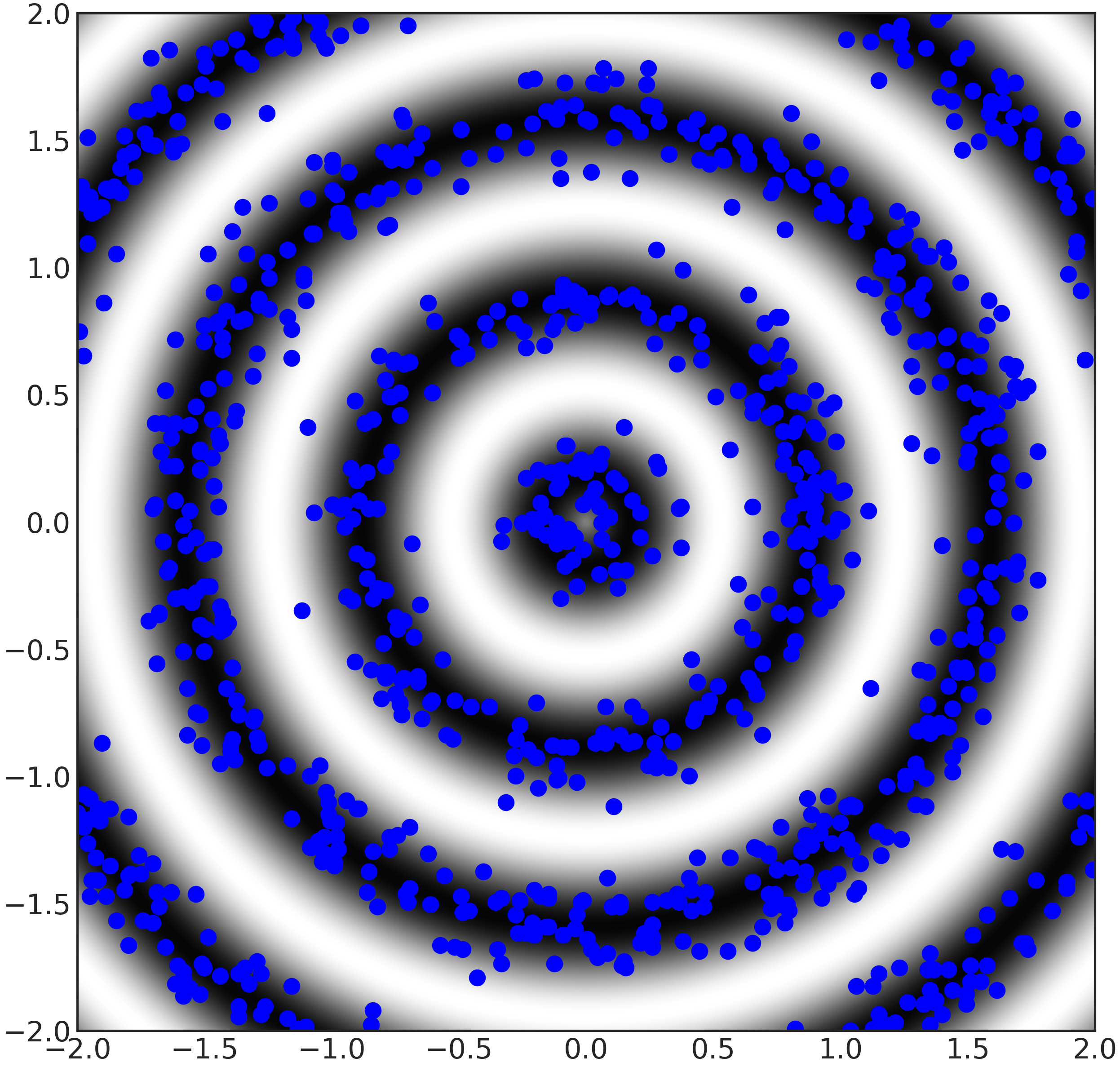}
        }
    \hfill
    \subfigure[$\alpha=0.75$]{%
        \includegraphics[height=0.22\textwidth]{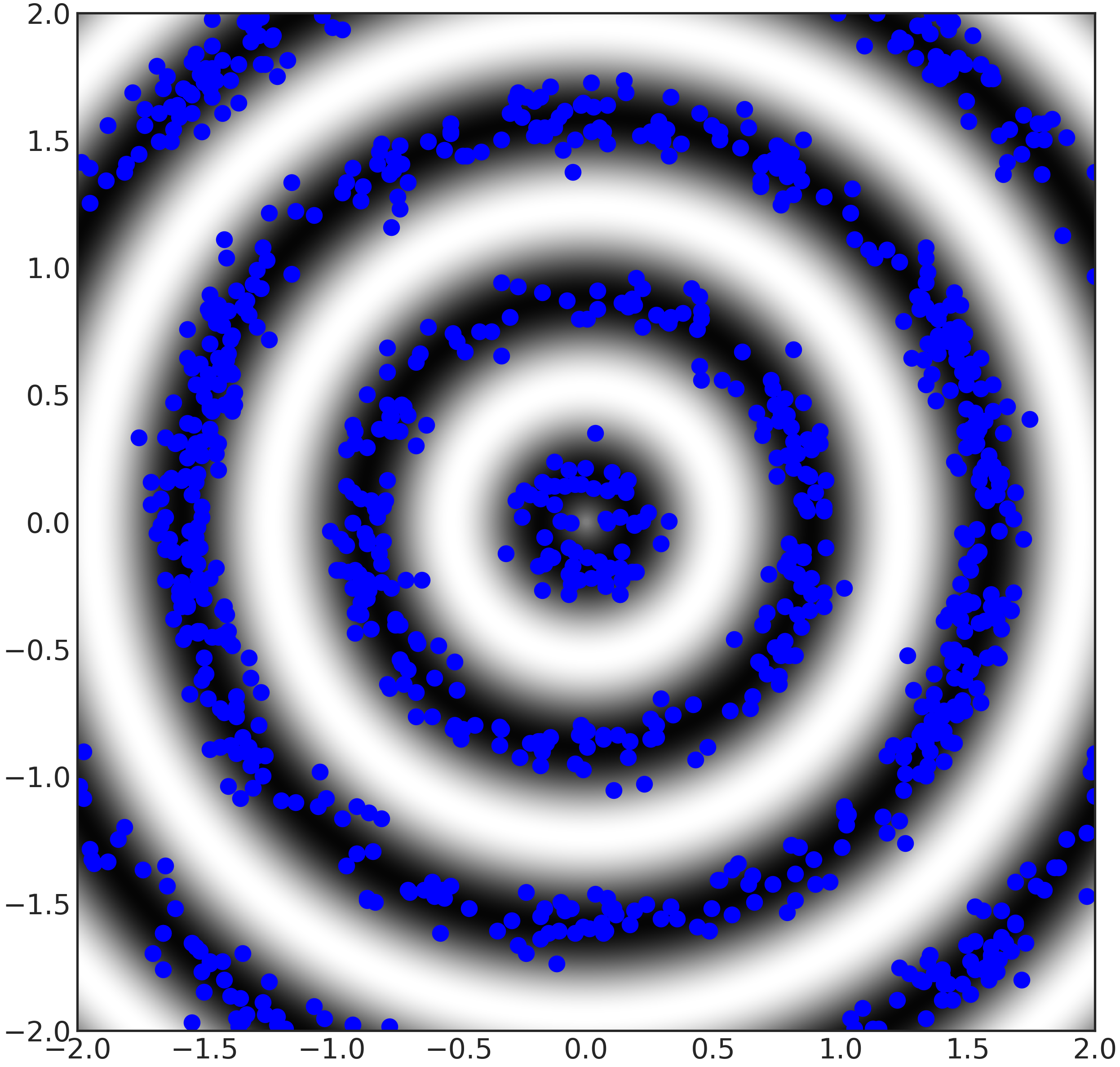}
        }
    \hfill
    \subfigure[$\alpha=0.9$]{%
        \includegraphics[height=0.22\textwidth]{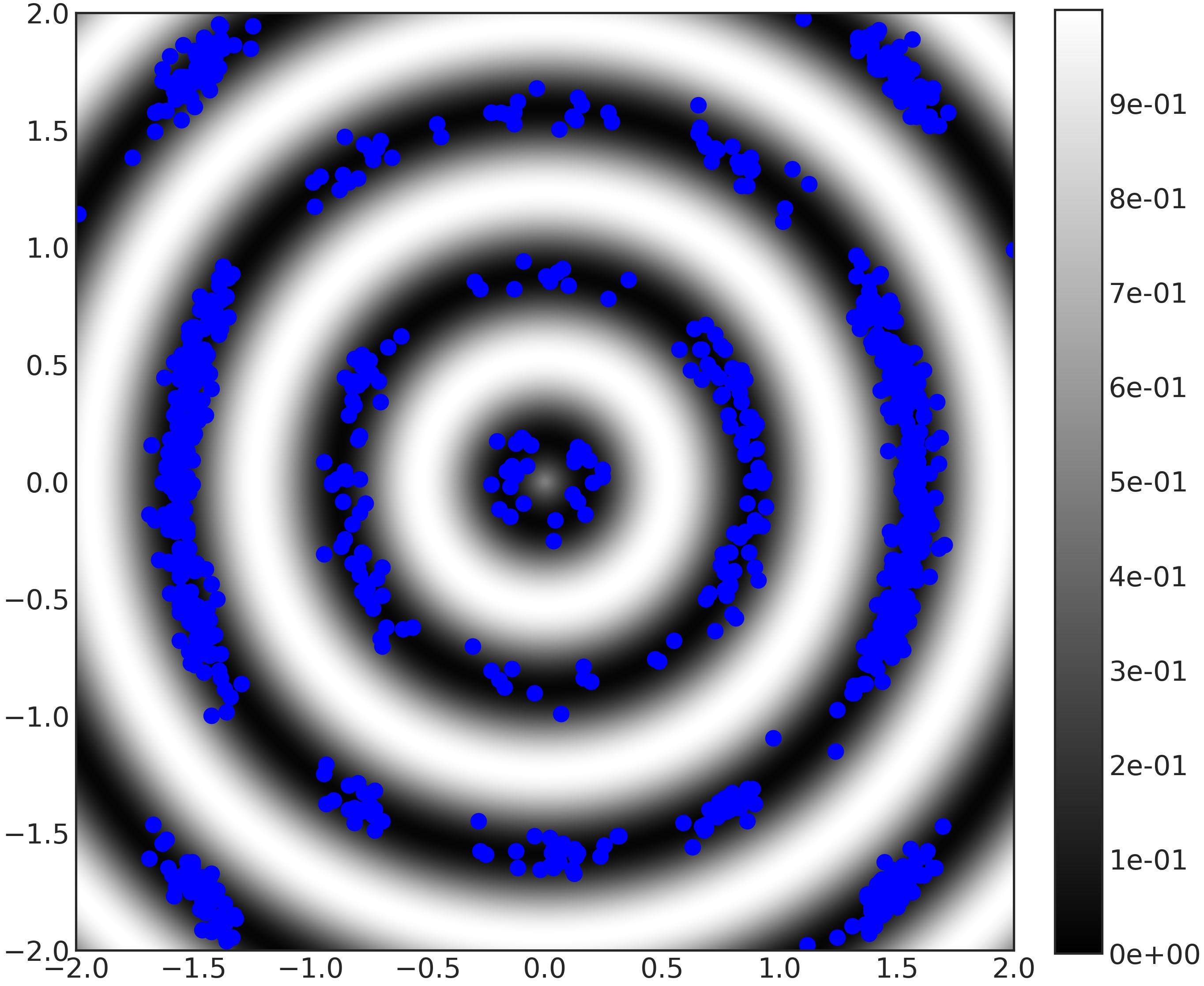}
        }
    
    \caption{$1000$ samples (shown as blue dots) from the TT distribution of a 2D sinusoidal function for different values of $\alpha$. The function has an infinite number of global optima (on the dark circles) and we see that TTGO is able to sample from these regions. As we increase $\alpha$, the samples become more concentrated on the circles.}
    \label{fig:sine_wave}
\end{figure*}

Furthermore, we also evaluate the prioritized sampling approach proposed in this article. We show how the sampling parameter $\alpha$ influences the obtained solutions. When $\alpha$ is small, the generated samples cover a wide region around many different local optima. When $\alpha$ is close to one, the obtained samples are observed to be very close to the global optima. All the results can be observed in Figure~\ref{fig:sine_wave}-~\ref{fig:gmm}, where the samples from the TT distribution (without any refinement by another solver) are shown as blue dots. The contour plot corresponds to the cost function in Figure~\ref{fig:sine_wave}-\ref{fig:himmelblaue_alpha} and the density function in Figure~\ref{fig:gmm}, where the dark region is the region with low cost (i.e., high density). 

In all of the test cases, we observe that the solutions proposed by TTGO are close to the actual optima and that the refinement using SLSQP quickly leads to global optima consistently. When there exist multiple solutions, we are also able to find them. Note that the task parameters influence the locations of the global optima, and TTGO can adapt accordingly by conditioning the model on the given task parameters. In all of the following cases, we choose a uniform discretization of the domain with the number of discretization points $n_k=500$ set for each variable to construct the TT model.

Except for the sinusoidal function, uniform sampling requires a large number of samples to reach the global optima. For the mixture of Gaussians case, it fails most of the time to get the global optima even after the refinement step. In contrast, we could consistently get the optima using TTGO with few samples. In fact, by using $\alpha$ close to 1, we could find the global optima with just one sample from the TT distribution. 

\subsubsection{Sinusoidal Function:} 

\begin{align*}
    C(\bm{x}) &= 1 - 0.5(1+\sin(4 \pi ||\bm{x}||/\sqrt{d})) \\
    P(\bm{x})&= 1-C(\bm{x}),
\end{align*}
where $\bm{x}= \bm{x}_2 = (y_1,y_2)$, $, \Omega_{\bm{x}_2} = [-2,2]^2$ with no task parameters. For this function, finding the optima is not a difficult problem. However, as the cost function has uncountably many global optima (on the circles separated by one period of the sinusoidal function), we use it to test the approximation power of TT-model and check the multimodality in the TTGO samples. As we can see in Figure~\ref{fig:sine_wave} for $d=2$, the samples from the TT model mainly come from the modes corresponding to the optima and the nearby region with cost values comparable to the optimal cost. At $\alpha=0$, we can still observe a few samples in the white area (low density region), and as we increase $\alpha$, the samples become more concentrated in the dark area, i.e., high-density region.

\begin{figure*}[t]
    \centering
    \subfigure[$(a,b)=(1,100)$]{%
        \includegraphics[height=0.22\textwidth]{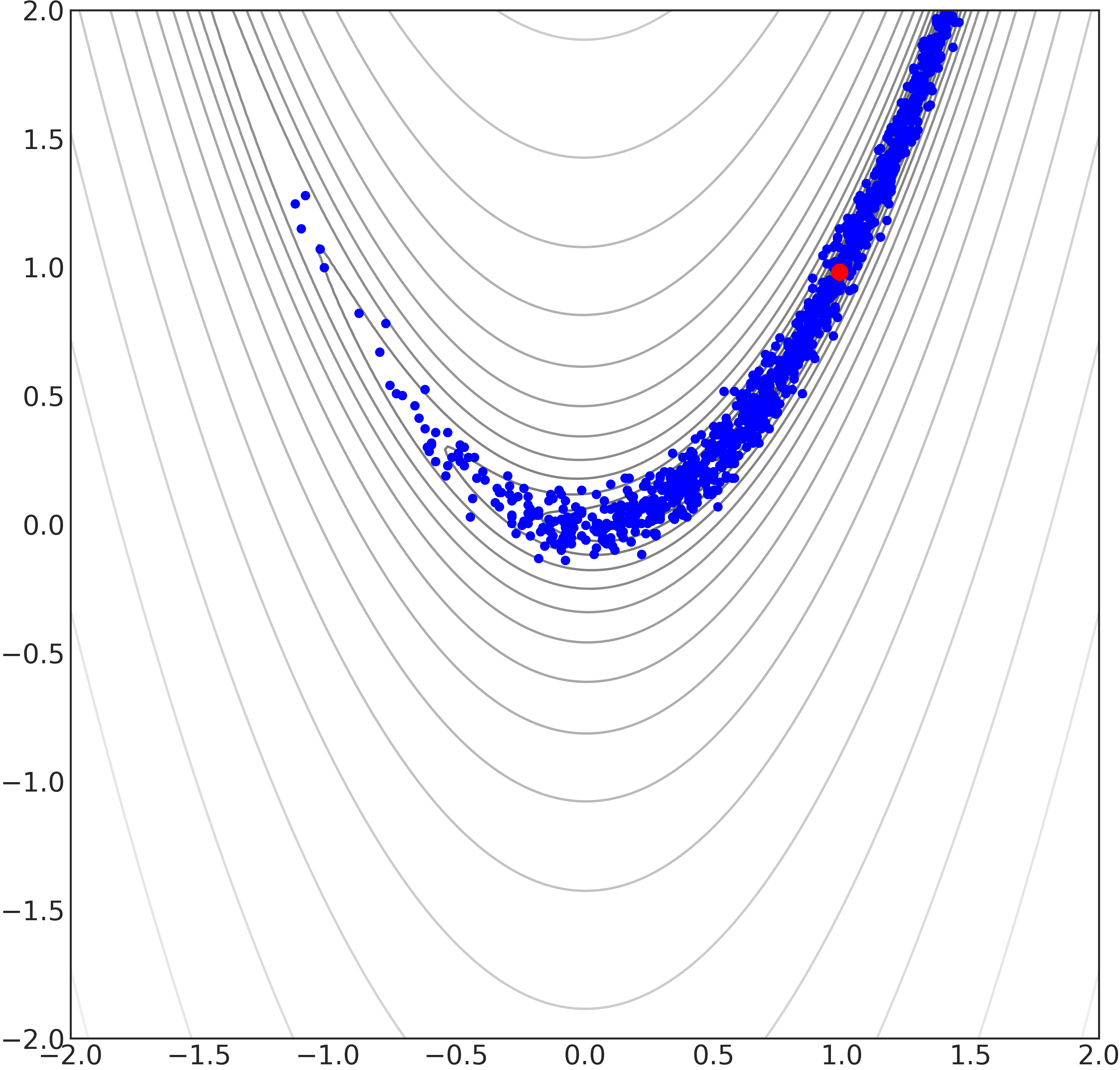}
        }
    \hfill
    \subfigure[$(a,b)=(0,140)$]{%
        \includegraphics[height=0.22\textwidth]{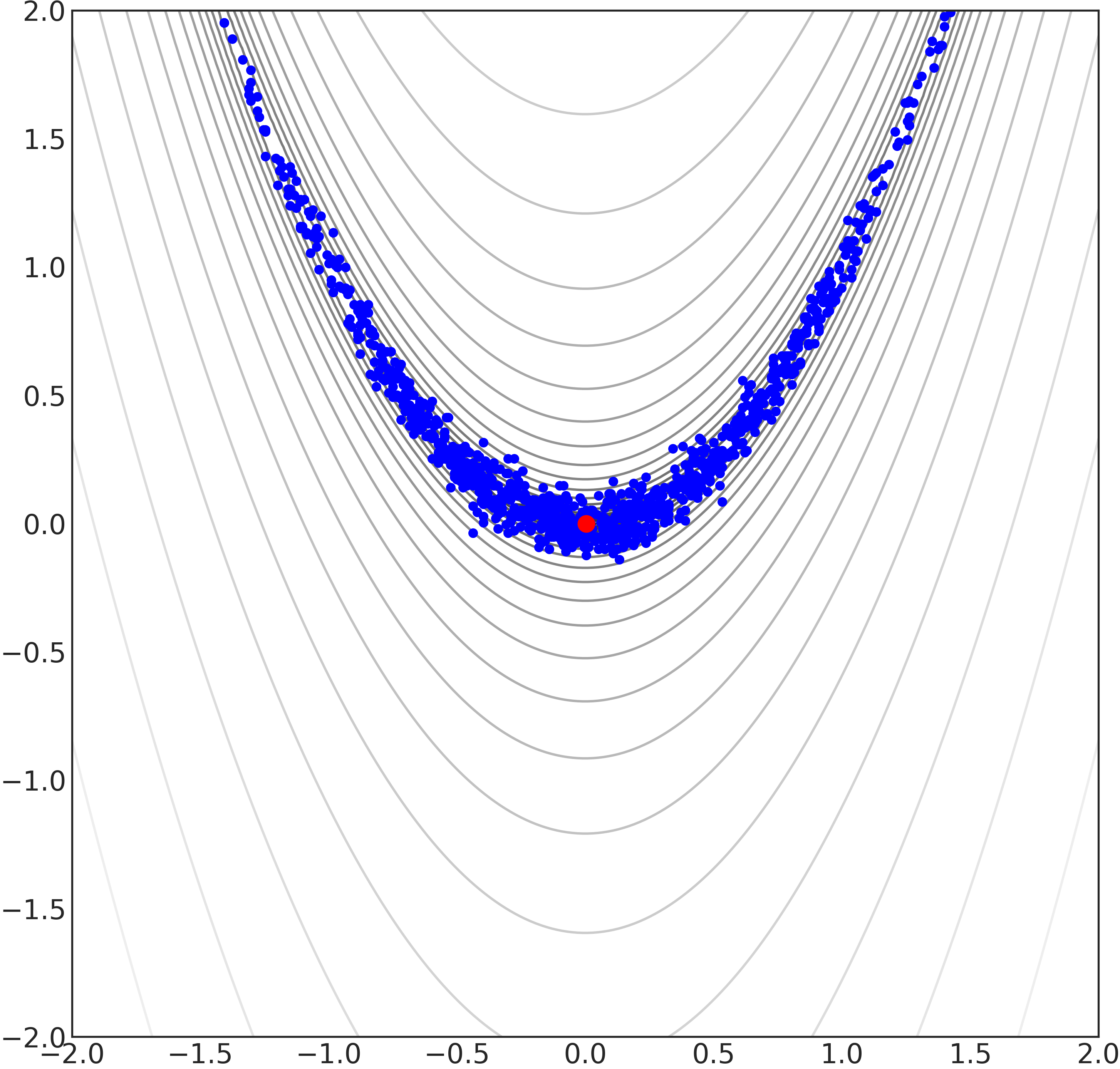}
        }
    \hfill
    \subfigure[$(a,b)=(0.5,60)$]{%
        \includegraphics[height=0.22\textwidth]{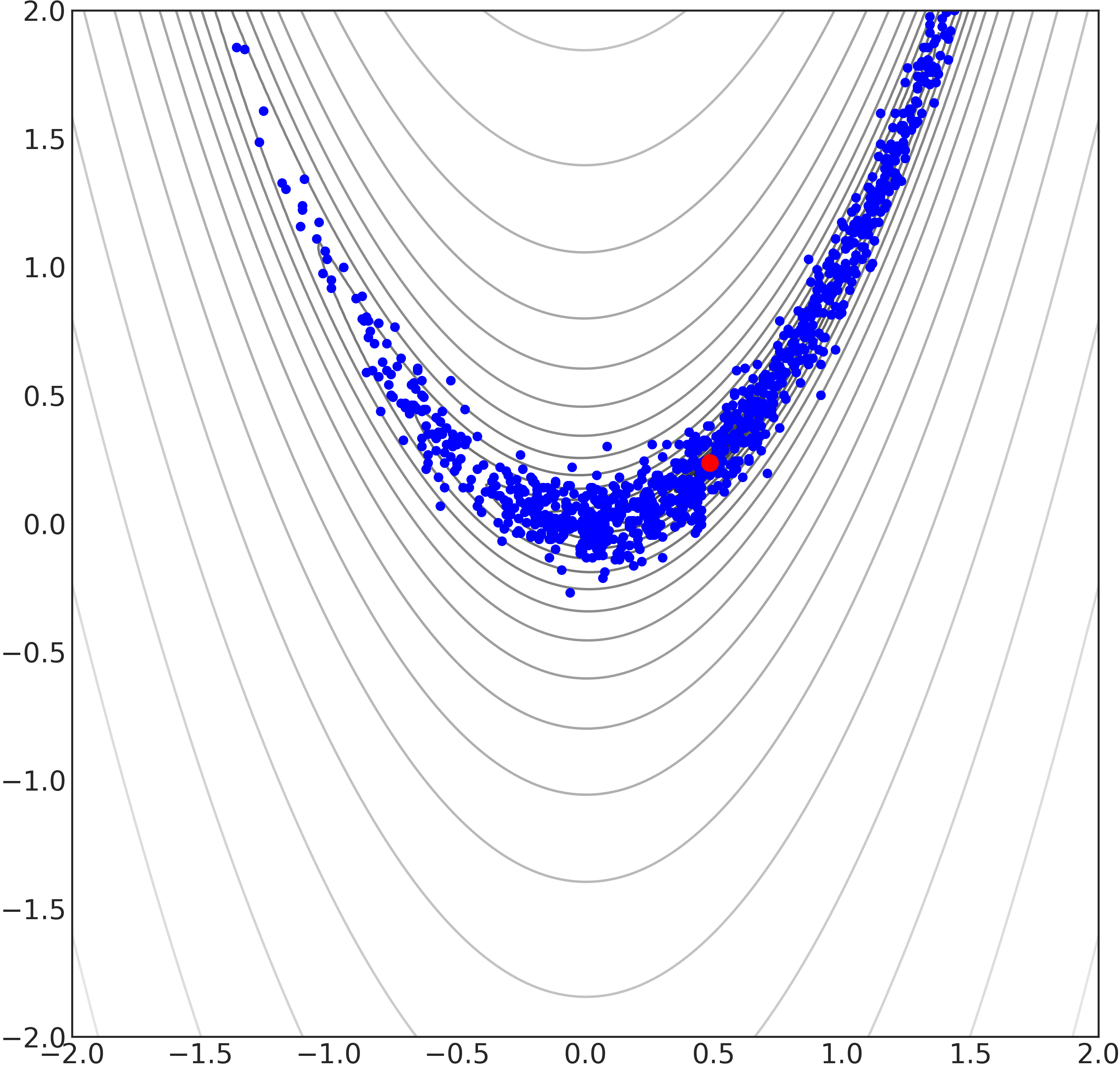}
        }
    \hfill
    \subfigure[$(a,b)=(-1,100)$]{%
        \includegraphics[height=0.22\textwidth]{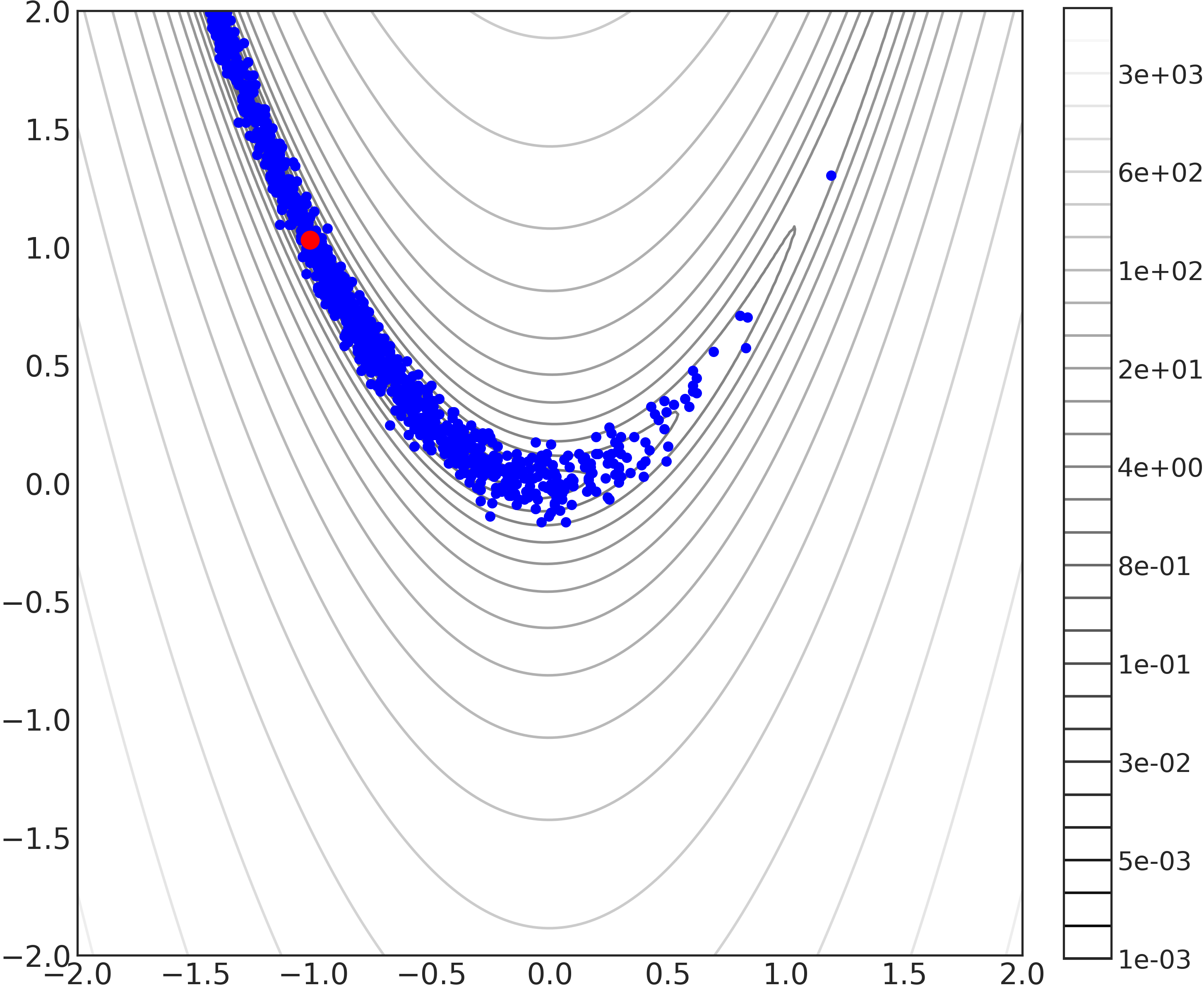}
        }
    
    \caption{$1000$ samples from the conditional TT distribution of a Rosenbrock function for various choices of the task parameters $(a,b)$ and $\alpha=0$. The function has a unique global optimum at $(a,a^2)$ as shown in red. As the task parameters change, the global optimum moves accordingly, but TTGO is still able to sample from the high-density regions.}
    \label{fig:rosenbrock_task}
\end{figure*}

\begin{figure*}[t]
    \centering
    \subfigure[$\alpha=0$]{%
        \includegraphics[height=0.22\textwidth]{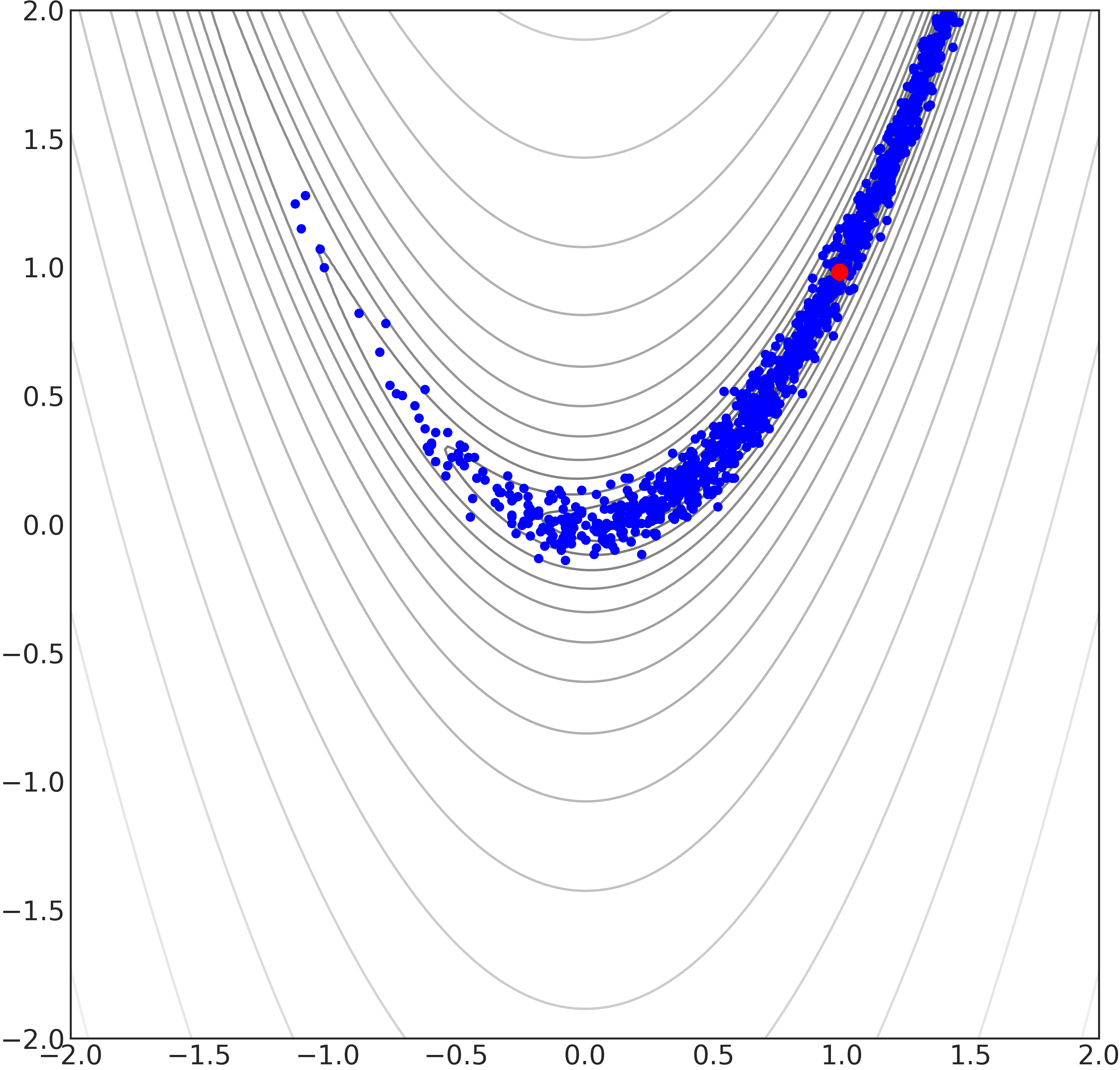}
        }
    \hfill
    \subfigure[$\alpha=0.5$]{%
        \includegraphics[height=0.22\textwidth]{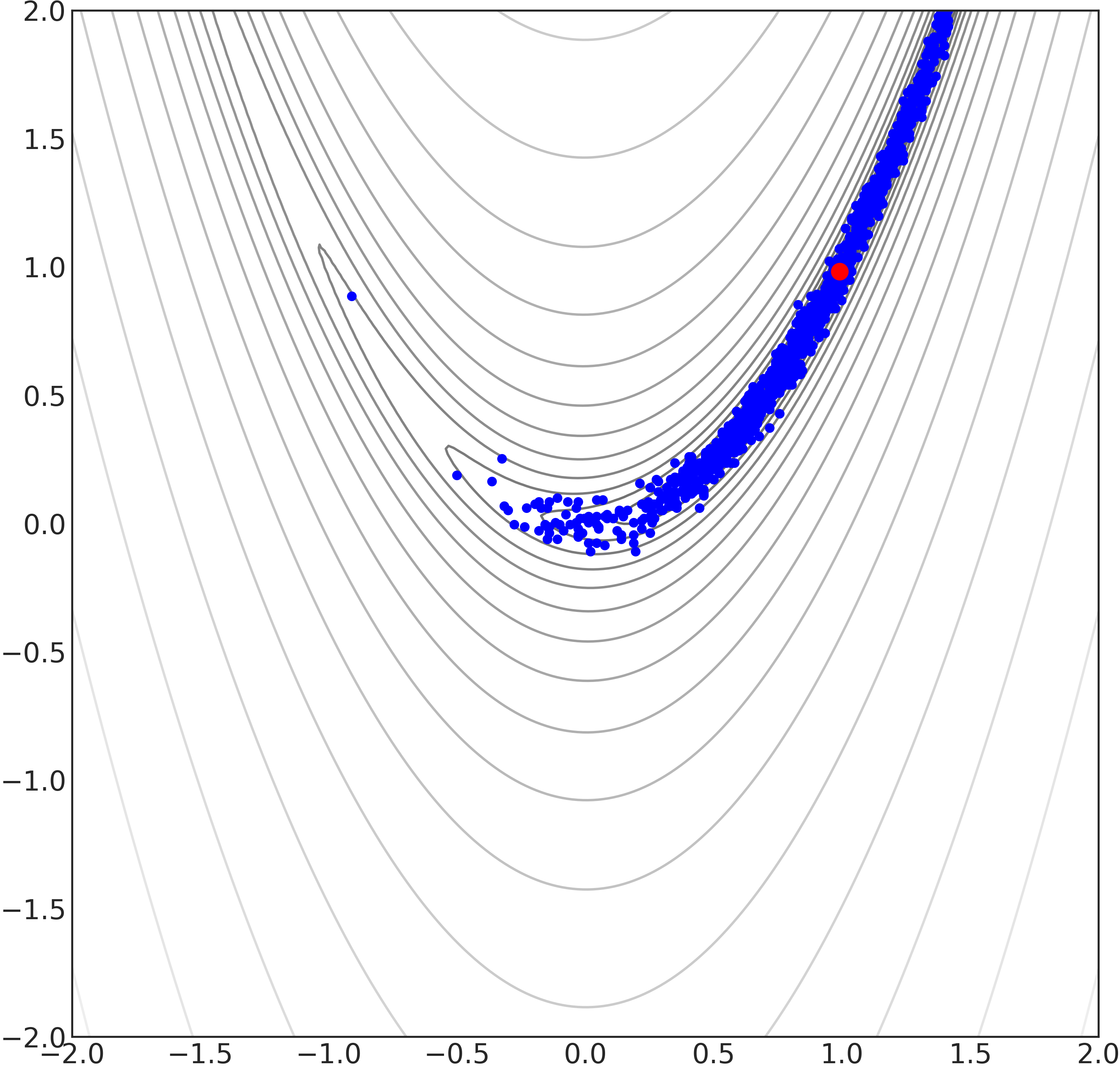}
        }
    \hfill
    \subfigure[$\alpha=0.75$]{%
        \includegraphics[height=0.22\textwidth]{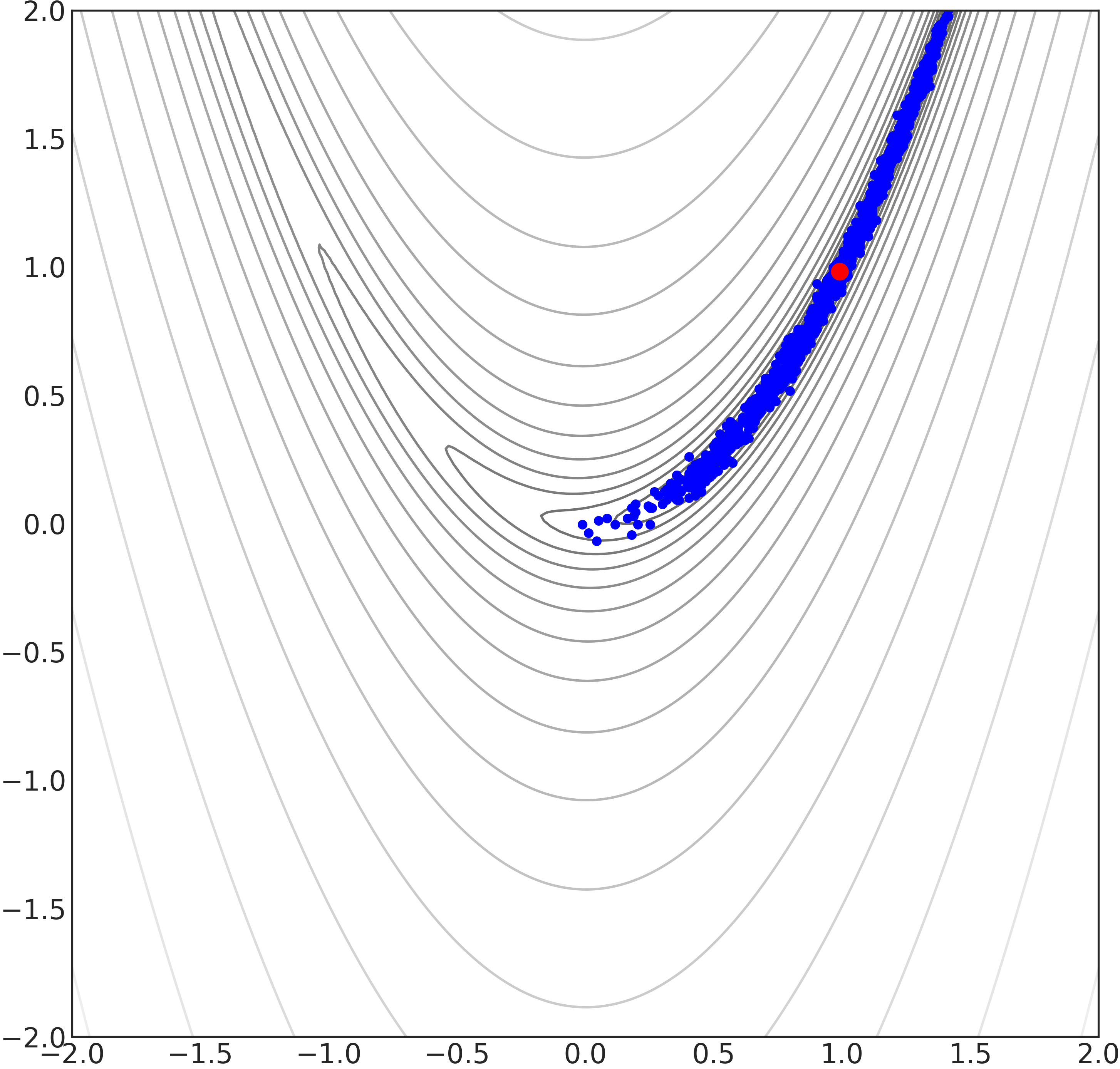}
        }
    \hfill
    \subfigure[$\alpha=0.9$]{%
        \includegraphics[height=0.22\textwidth]{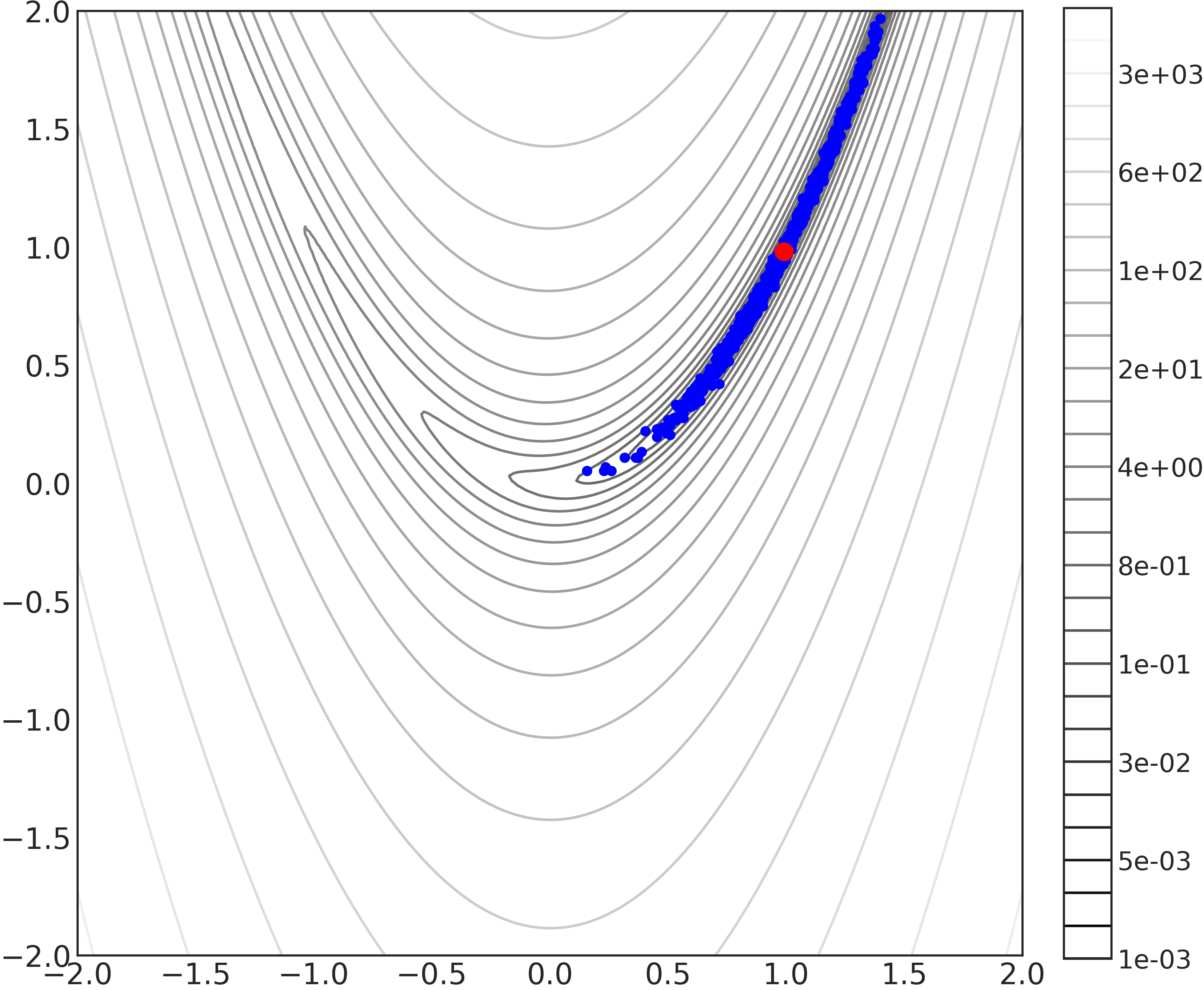}
        }
    
    \caption{$1000$ samples from the conditional TT distribution of a Rosenbrock function with the task parameters $a=1, b=100$ and various values of $\alpha$. As $\alpha$ increases, the samples become more concentrated around the global optimum (as shown in red).}
    \label{fig:rosenbrock_alpha}
\end{figure*}

\begin{figure*}[t]
    \centering
    \subfigure[$(a,b)=(3,3)$]{%
        \includegraphics[height=0.224\textwidth]{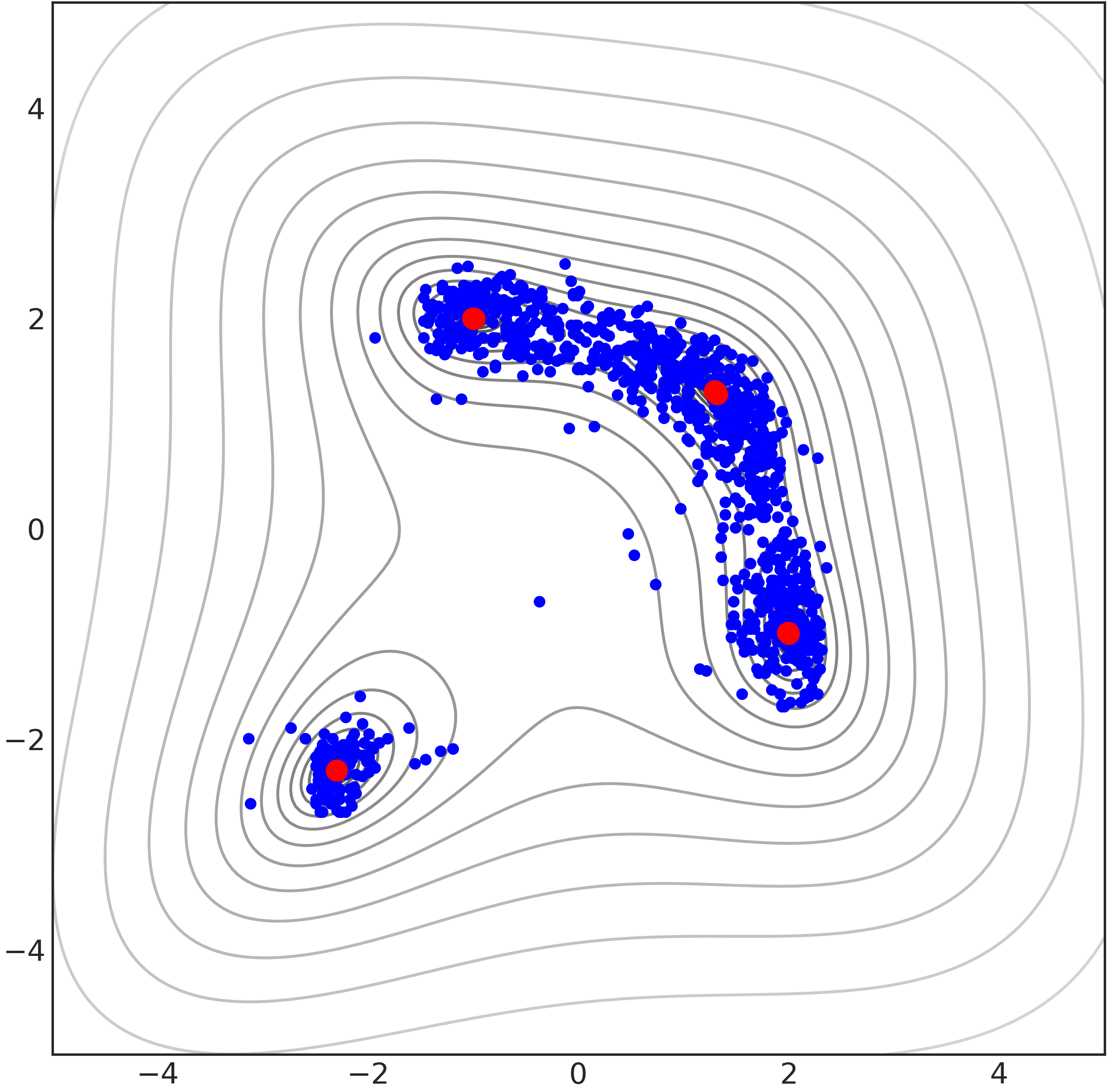}
        }\hfill
    \subfigure[$(a,b)=(3,14)$]{%
        \includegraphics[height=0.224\textwidth]{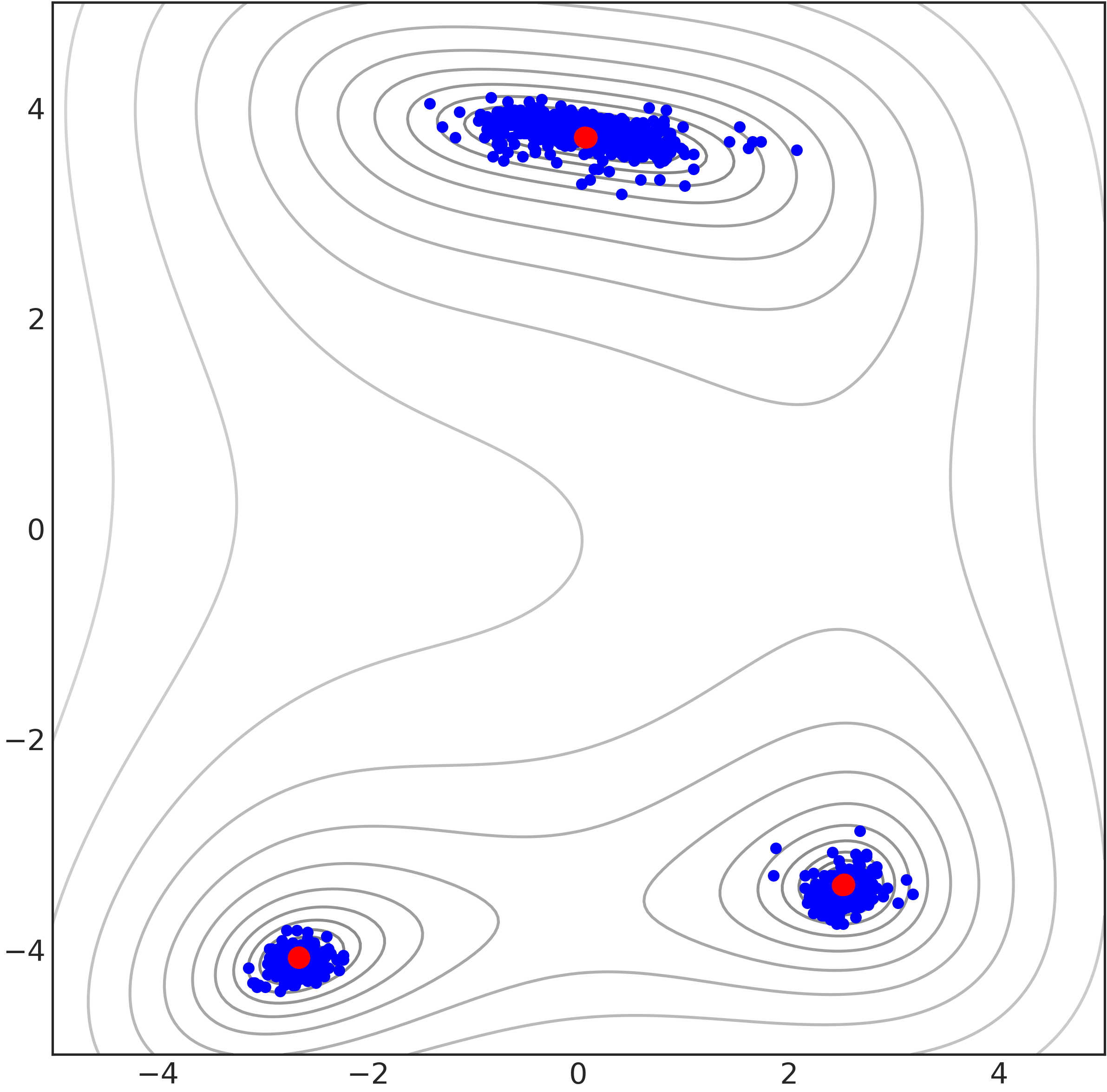}
        }\hfill
    \subfigure[$(a,b)=(7,11)$]{%
        \includegraphics[height=0.224\textwidth]{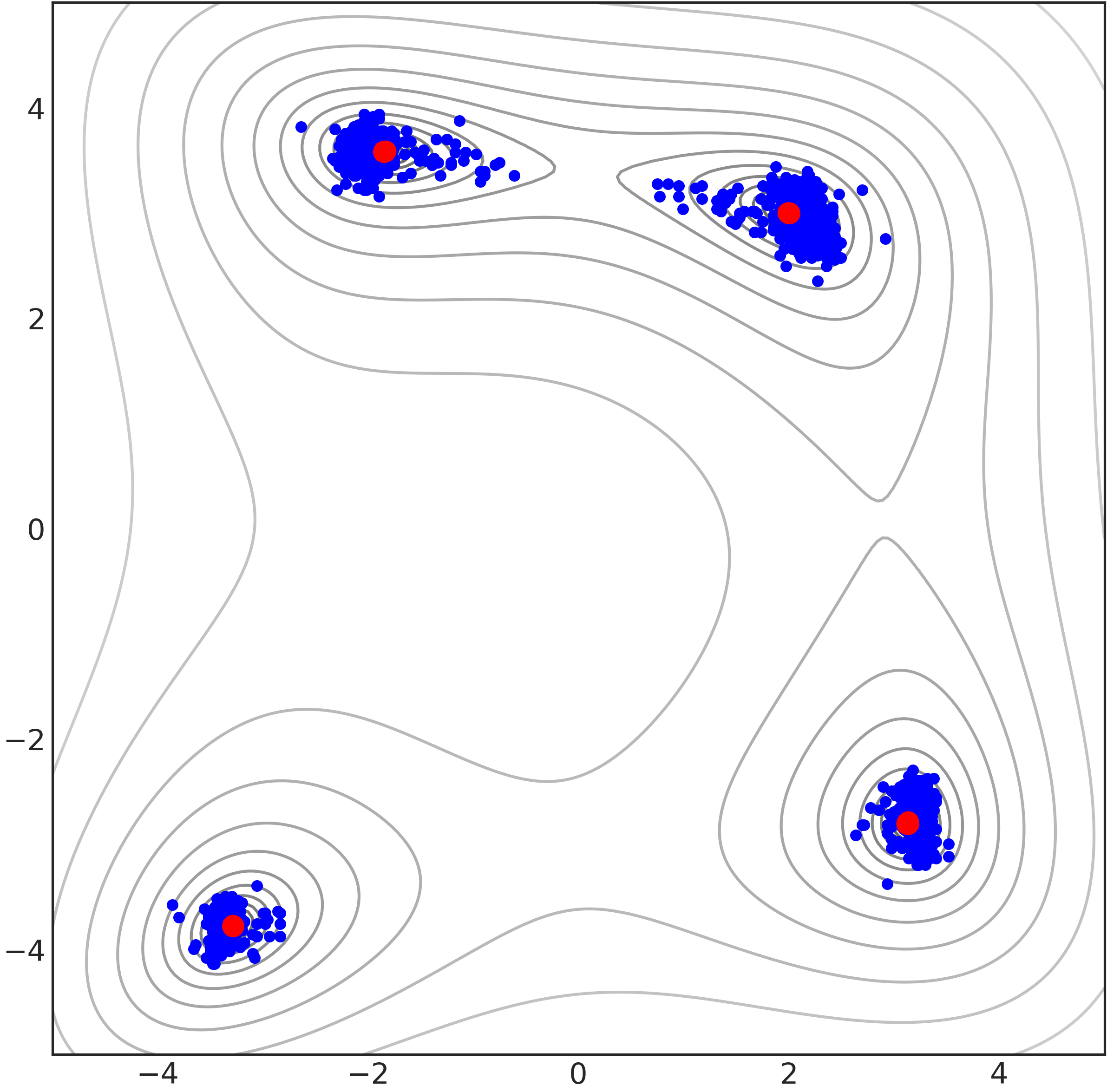}
        }\hfill
    \subfigure[$(a,b)=(13,5)$]{%
        \includegraphics[height=0.225\textwidth]{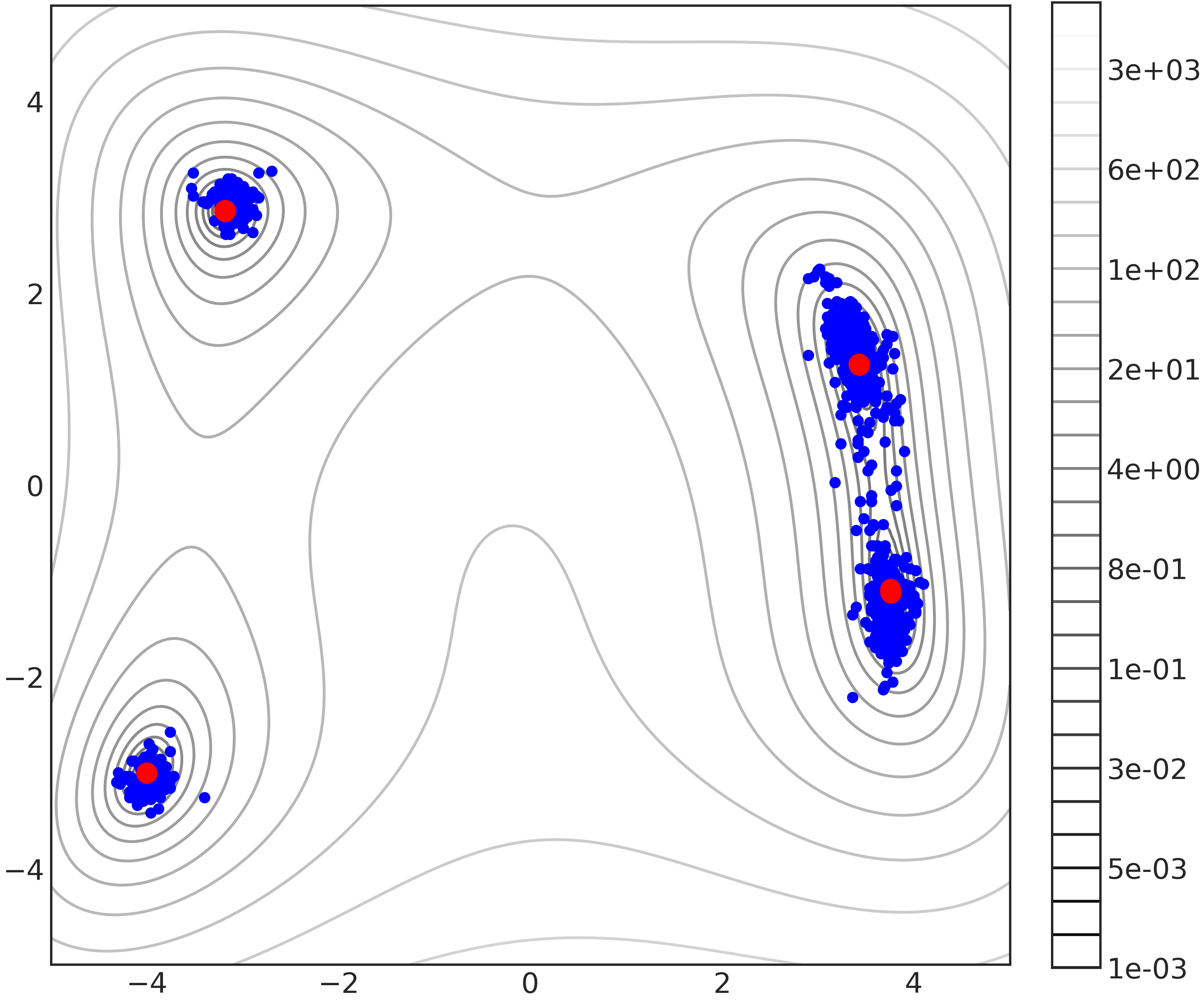}
        }
    
    \caption{$1000$ samples from the conditional TT distribution of a 2D Himmelblau function for various choices of the task parameters $(a,b)$ and $\alpha=0$. The location of the multiple global optima (in red) depend on the task parameters, but TTGO is able to generate the samples from the high-density regions.}
    \label{fig:himmelblaue_tasks}
\end{figure*}

\begin{figure*}[t]
    \centering
    \subfigure[$\alpha=0$]{%
        \includegraphics[height=0.22\textwidth]{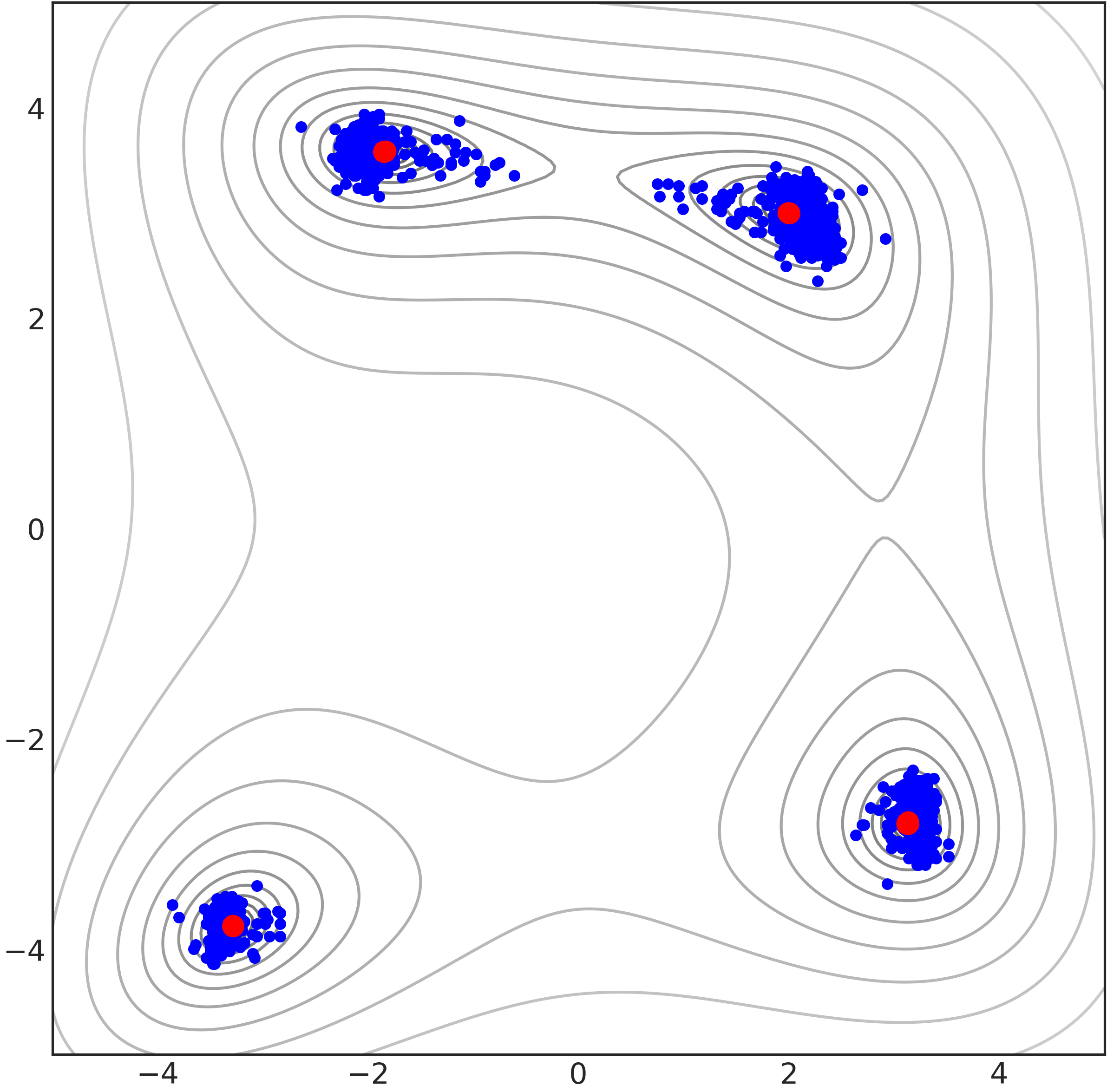}
        }
    \hfill
    \subfigure[$\alpha=0.5$]{%
        \includegraphics[height=0.22\textwidth]{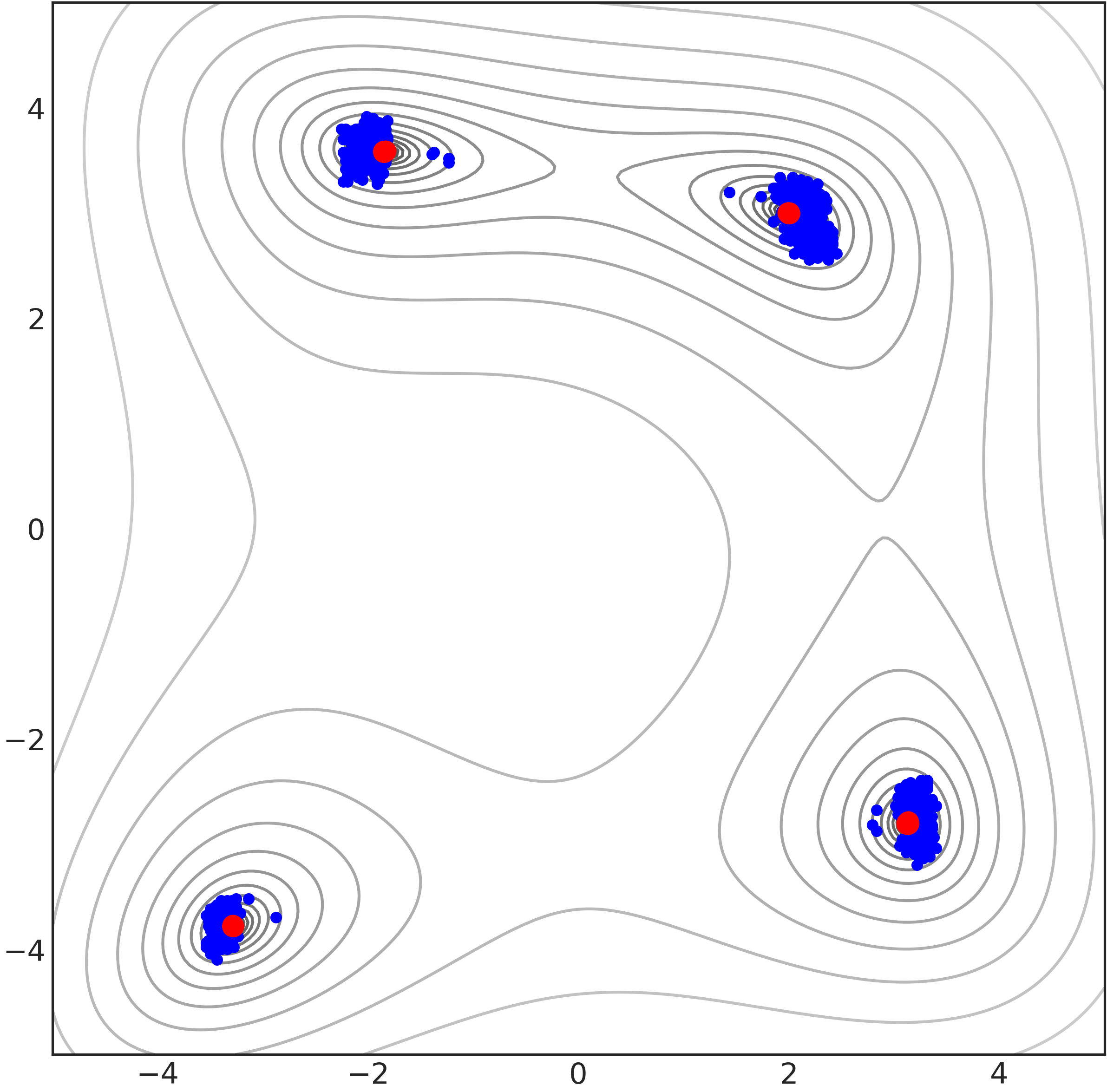}
        }
    \hfill
    \subfigure[$\alpha=0.9$]{%
        \includegraphics[height=0.22\textwidth]{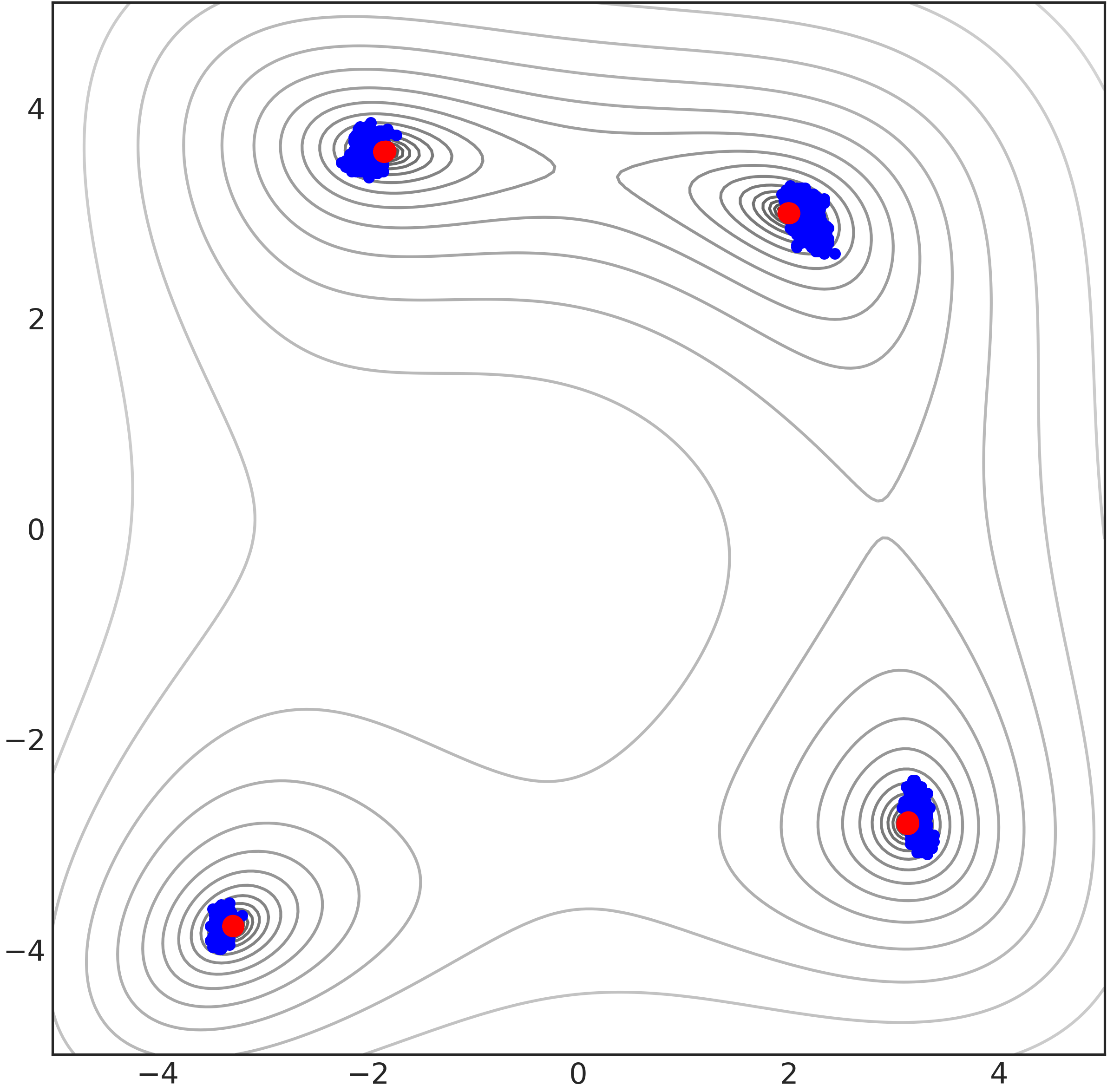}
        }
    \hfill
    \subfigure[$\alpha=1$]{%
        \includegraphics[height=0.22\textwidth]{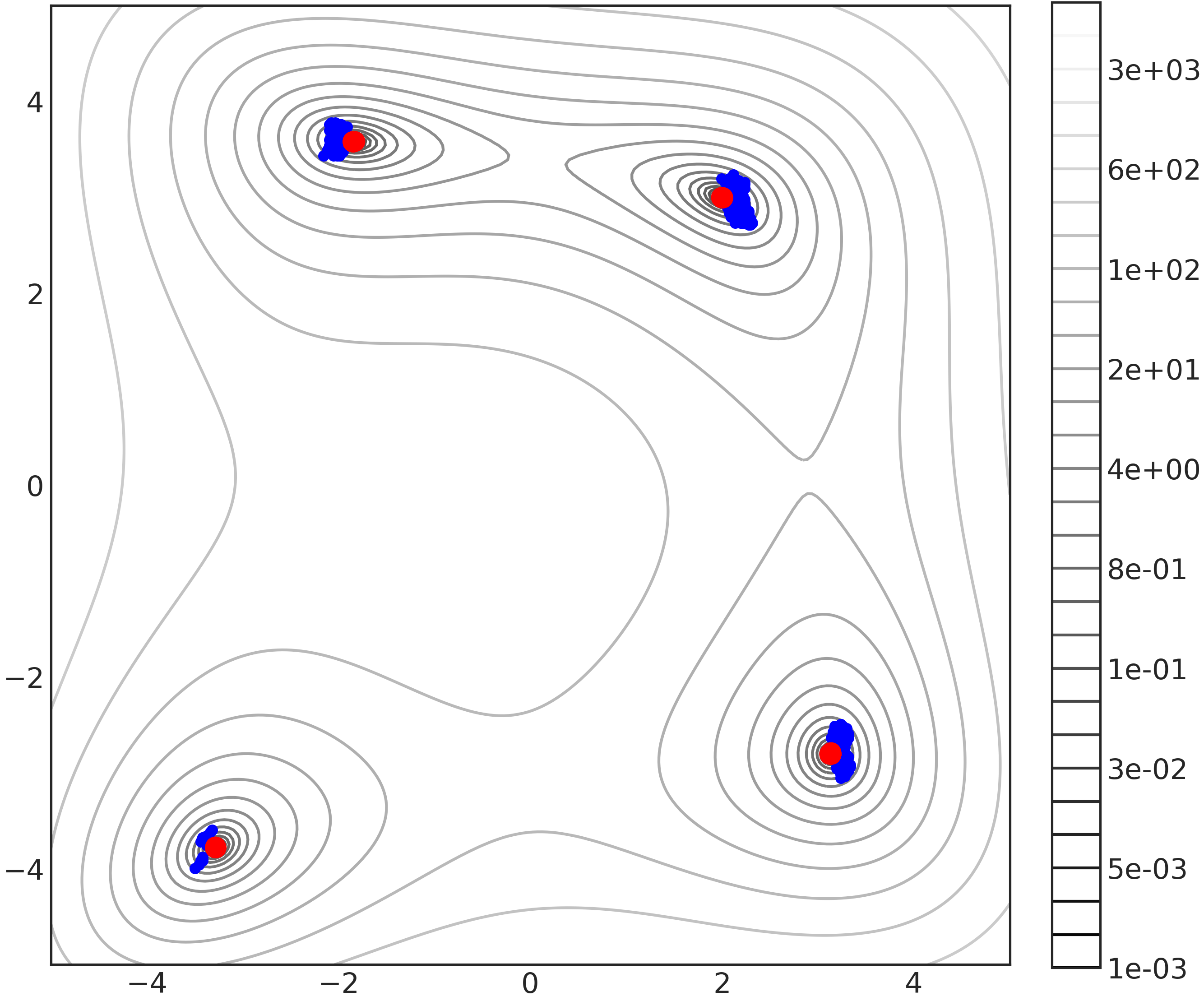}
        }
    
    \caption{$1000$ samples from the conditional TT distribution of a 2D Himmelblau function with task parameters $a=7, b=11$ for various values of $\alpha$. As $\alpha$ increases, the samples become more concentrated around the global optima.}
    \label{fig:himmelblaue_alpha}
\end{figure*}

\subsubsection{Rosenbrock Function:}

\begin{align*}
    C(a,b,y_1,\ldots,y_{d_1}) &=\sum_{k=1}^{d_2/2}(y_{2k-1}-a)^2+b (y_{2k-1}-y^2_{2k})^2  \\
    P(\bm{x}) &= \exp(-C(\bm{x})^2),
\end{align*}
where $\bm{x}=(\bm{x}_1,\bm{x}_2)$, $\bm{x}_1 = (a,b)$, $\bm{x}_2=(y_1,\ldots,y_{d_2})$, $\Omega_{\bm{x}_1} = [-1.5,1.5]\times[50,150]$, $\Omega_{\bm{x}_2}=[-2,2]^{d_2}$. The function is similar to a banana distribution which is quite difficult to approximate. The cost function $C(\bm{x})$ for a specified $(a,b)$ has a unique global minima at $(a,a^2,\ldots,a,a^2)$. However, if we do not initialize the solution from the parabolic valley area (see Figure~\ref{fig:rosenbrock_task}), a gradient-based solver will have difficulty in  converging to the global optima quickly. We can see from Figure~\ref{fig:rosenbrock_task} that TTGO samples are concentrated around this region, allowing most of them to reach the global optima after refinement. In fact, by increasing the $\alpha$, the TTGO samples are already very close to the global optima (as shown in red). 

Figure~\ref{fig:rosenbrock_alpha} shows how the task parameters $\bm{x}_1=(a,b)$ change the shape of the function with respect to $\bm{x}_2$ and consequently the location of the global optima. After the offline training, we condition our TT model on these task parameters and sample from the conditional distribution $\text{Pr}(\bm{x}_2|\bm{x}_1=(a,b))$. We can see in this figure that TTGO can adapt to the new task parameters easily, as the samples are concentrated around the new global optima. 

We also test TTGO performance on Rosenbrock functions for $d_2$ up to $30$ and find that it can find the global optima consistently. We show in the figures the results for the 2D case, which are easier to visualize.

\subsubsection{Himmelblau's function:}

\begin{align*}
    C(a,b,y_1,y_2) &= (y_1^2+y_2-a)^2+(y_1+y_2^2-b)^2 \\
    P(\bm{x}) &= \exp(-C(\bm{x})^2),
\end{align*}
where $\bm{x}=(\bm{x}_1,\bm{x}_2)$, $\bm{x}_1 = (a,b)$, $\bm{x}_2=(y_1,y_2)$,  $P(\bm{x}) = \exp(-C(\bm{x}))$, $\Omega_{\bm{x}_1} = [0,15]^2$, $\Omega_{\bm{x}_2}=[-5,5]^2$. The cost function $C(a,b,y_1,y_2)$ for a given $(a,b)$ has multiple distinct global optima and many local optima. The samples from the TT distribution $\text{Pr}(\bm{x}_2|\bm{x}_1=(a,b))$ are shown in Figure~\ref{fig:himmelblaue_tasks}--\ref{fig:himmelblaue_alpha} for different choice of task parameters and the prioritized sampling parameters $\alpha$. We can see that TTGO can generate samples from all of the modes consistently according to the task parameters.

\subsubsection{Mixture of Gaussians:}
$$P(\bm{x})=\sum_{j=1}^{J} \alpha_j \exp(-\beta_j ||\bm{x}-\bm{a}_j||^2),$$

We use an unnormalized mixture of Gaussian functions to define the probability function $P(\bm{x})$ to test our framework for high-dimensional multimodal functions. For verification, we design the mixture components so that we know the global optima \emph{a priori} by carefully choosing the centers, mixture coefficients and variances. We test it for various values for the number of mixtures $J$, $\beta \in [1,1000]$ and the dimension $d \in (2,\ldots,50)$ of $\bm{x}$. We choose $\bm{x}=(\bm{x}_1,\bm{x}_2)$ with $\Omega_{\bm{x}} = [-2,2]^d$ for various choices of values and dimension of $\bm{x}$. As TTGO does not differentiate between task parameters and optimization variables internally, we could consider various possibilities to segment $\bm{x}$ into $(\bm{x}_1, \bm{x}_2)$ as task parameters and decision variables. We tested this problem for $d<100$, and our approach could consistently find the optima with less than $100$ samples from the TT-model, for arbitrary choice of variables being conditioned as task parameters. In contrast, finding the optima using Newton-type optimization with random initialization is highly unlikely for $\beta_j > 1$ and $d>10$, even after considering millions of samples from uniform distribution for initialization. 

\begin{figure*}[t]
    \centering
    \subfigure[$\alpha=0$]{%
        \includegraphics[height=0.22\textwidth]{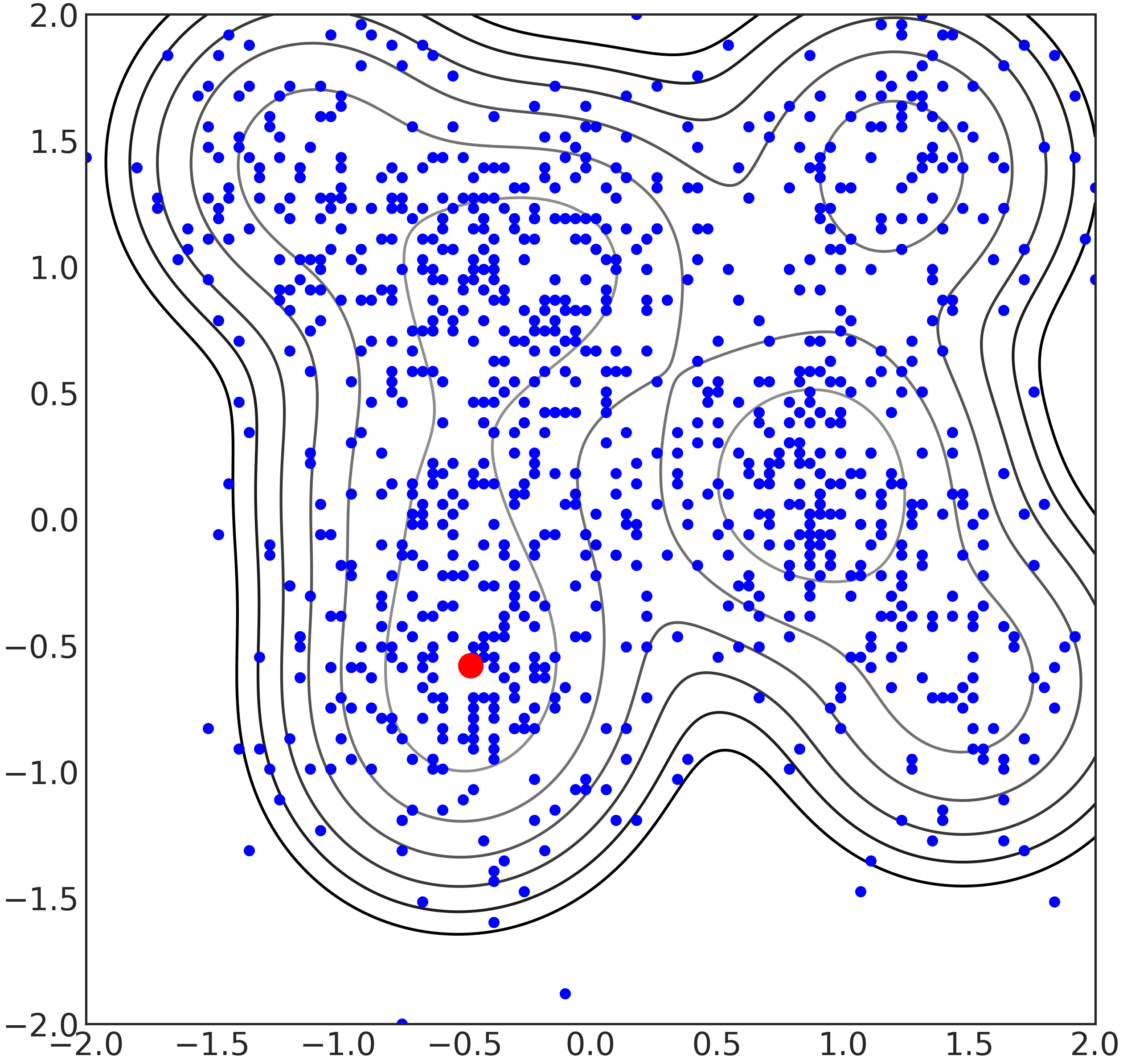}
        }
    \hfill
    \subfigure[$\alpha=0.5$]{%
        \includegraphics[height=0.22\textwidth]{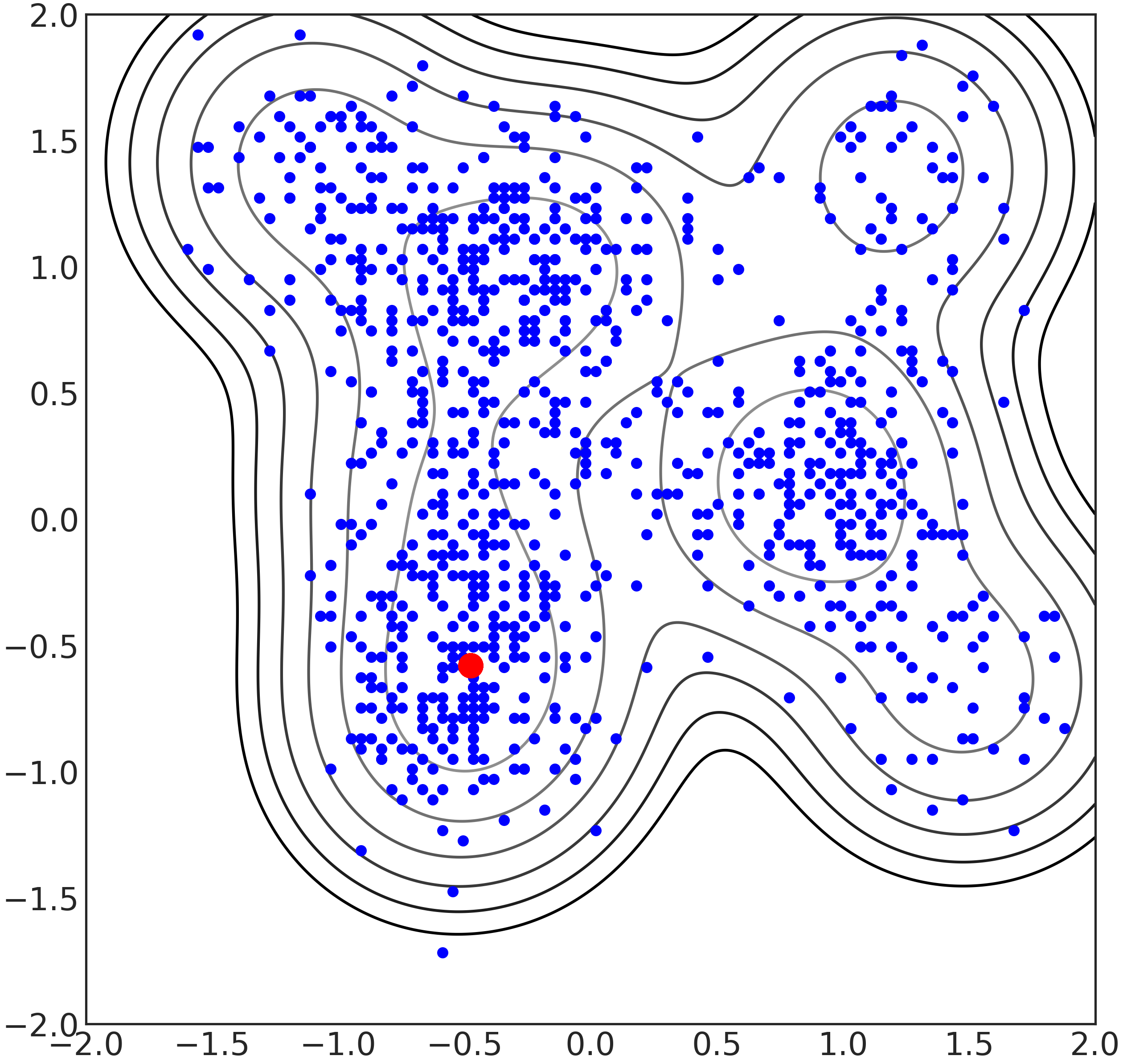}
        }
    \hfill
    \subfigure[$\alpha=0.75$]{%
        \includegraphics[height=0.22\textwidth]{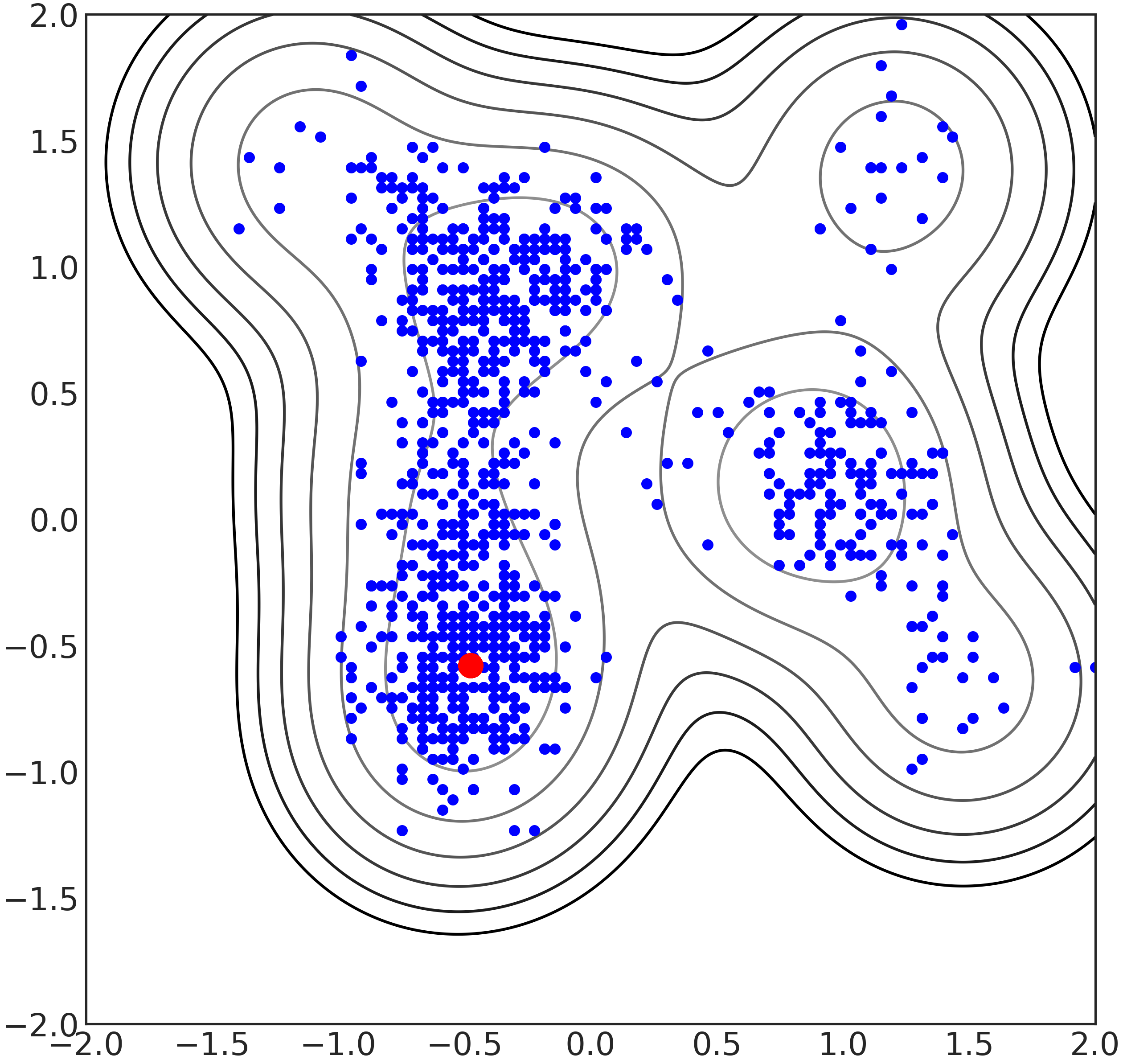}
        }
    \hfill
    \subfigure[$\alpha=0.9$]{%
        \includegraphics[height=0.22\textwidth]{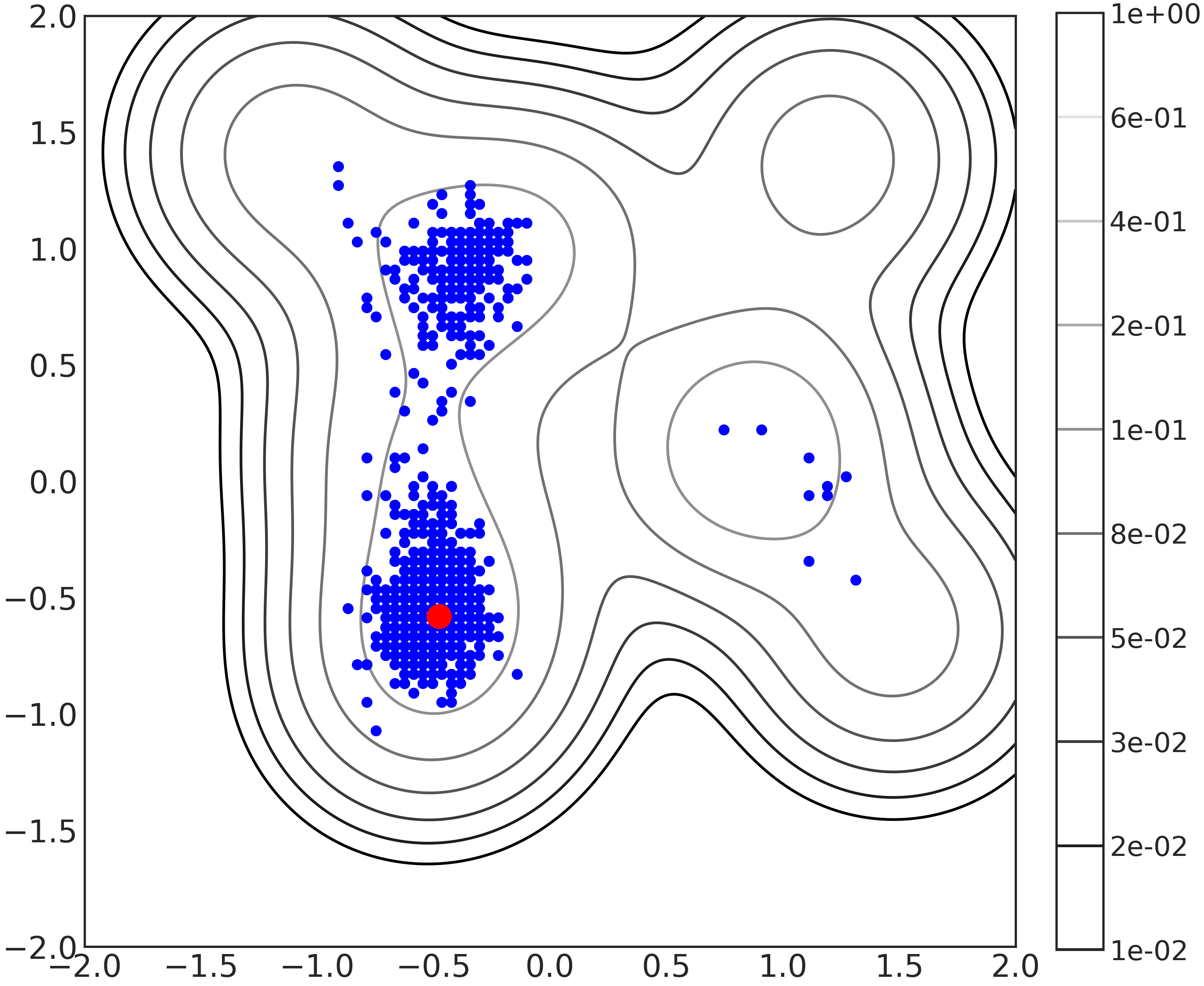}
        }
    
    \caption{$1000$ samples from the conditional TT distribution of a mixture of Gaussians with $J=10$, $d=50$, $\beta_j=175$, and various values of $\alpha$. For visualization, we choose the the first $d-2$ coordinates of $\mu_j$ to be the same for all $j$ and choose the task-parameters to be the first $d-2$ coordinate of the centers. This density function has one global optimum (in red) and some other modes that are comparable to the global optimum. As $\alpha$ increases, the samples become more concentrated around the mode with the highest density.}
    \label{fig:gmm}
\end{figure*}

Figure~\ref{fig:gmm} shows one particular example with $J=10$, $\beta_j=175$ and $d=50$. To visualize, we choose $\bm{x}_1 \in \mb{R}^{d-2}$ and $\bm{x}_2 \in \mb{R}^2$, and we generate 1000 samples from the conditional TT distribution $\text{Pr}(\bm{x}_2|\bm{x}_1)$. With low values of $\alpha$, the samples are generated around all the different modes, but as $\alpha$ is increased, the samples become more concentrated around the mode with the highest probability.

\subsection{Inverse Kinematics Formulation}
\label{ik_formulation}

The cost function for the inverse kinematics problem in Section~\ref{ik_7dof} is given by
\begin{equation}
    C(\bm{x}) = \frac{1}{3}  \Big(\frac{C_{p}(\bm{\theta},\bm{p}_d)}{\beta_{p}} + \frac{C_\text{obst}(\bm{\theta})}{\beta_\text{obst}} + \frac{C_\text{orient}(\bm{\theta})}{\beta_\text{orient}}\Big), 
\end{equation} 

\noindent where $\bm{x}=(\bm{x}_1,\bm{x}_2)$ and:
\begin{itemize}

    \item $C_{p}(\bm{\theta},\bm{p}_d) = \|\bm{p}_d-\bm{p}(\bm{\theta})\|$, Euclidean distance of the end effector position from the desired position.
    \item  $C_\text{obst}(\bm{\theta})$ represents the obstacle cost based on the Signed Distance Function (SDF). The links are approximated as a set of spheres (as done in CHOMP), and we use the SDF to compute the distance from each sphere to the nearest obstacle.
    \item $C_\text{orient}(\bm{\theta})$ represents the cost on the orientation of the end-effector. In our application, we specify a desired orientation of the end-effector, given by quaternion $\bm{q}_d$, while allowing a rotation around the axis of rotation $\bm{v}_d$ which corresponds to the z-axis of the world frame. This constraints the gripper orientation to be horizontal while allowing rotation around the z-axis. This is suitable for picking cylindrical objects from a shelf. The cost is then  $C_\text{orient}(\bm{\theta}) = 1 - <\bm{v}(\bm{\theta}), \bm{v}_d>^2$ where $\bm{v}(\bm{\theta})$ represents the screw axis (computed from the quaternion) of the actual end-effector frame w.r.t.~the desired frame. Alternatively, if the application demands a variation in the desired orientation, one could use the pose $(\bm{p}_d,\bm{q}_d)$ directly as the task parameter. 
    \item $\beta_p, \beta_\text{obst}, \beta_\text{orient}$ are scaling factors for each cost. Intuitively, they represent the acceptable value for each cost. We use $\beta_p=0.05$, $\beta_\text{obst}=0.01$ , and $\beta_\text{orient}=0.2$ for the orientation.
\end{itemize}

For the IK problem of the 6-DoF UR10 robot, there is no obstacle cost, and the orientation is specified to be identity (corresponding to upward-facing end-effector orientation) without any free axis of rotation.

\subsection{Motion Planning Formulation}
\label{appendix_mp}

For both the reaching and the pick-and-place tasks, the cost function, $\bm{x}=(\bm{x}_1,\bm{x}_2)$,  is given by
\begin{equation}
\footnotesize
    C(\bm{x}) = \frac{1}{4}  \Big(\frac{C_{p}(\bm{x})}{\beta_{p}} + \frac{C_\text{obst}(\bm{x})}{\beta_\text{obst}} +
     \frac{C_\text{orient}(\bm{x})}{\beta_\text{orient}} + \frac{C_\text{control}(\bm{x})}{\beta_\text{control}} \Big) 
\end{equation} 
with the following objectives:
\begin{itemize}
    \item $C_p(\bm{x})$ represents the cost on the end effector position(s) from the target location(s).
    \item  $C_\text{obst}(\bm{x})$ represents the cost incurred from the obstacles computed using SDF as in Section~\ref{ik_7dof} but accumulated for the whole motion.
    \item $C_\text{orient}(\bm{x})$ represents the cost on the orientation of the end effector at the target location(s). 
    \item $C_\text{control}(\bm{x})$ represents the cost of the length of the joint angle trajectory and the length of the end effector trajectory.
    \item $\beta_p, \beta_\text{obst}, \beta_\text{orient}, \beta_\text{control}$ are scaling factors for each cost. Intuitively, they represent the acceptable nominal cost value for each cost. We use $\beta_p = 0.05$, $\beta_\text{obst}=0.1$, $\beta_\text{orient}=0.2$, $\beta_\text{control}=2$. 
\end{itemize}

We consider the initial configuration of the manipulator to be fixed (we can relax this condition by considering the initial configuration as a task parameter). In the reaching task, the objective is to reach an end effector target location on the shelf. In the pick-and-place task, the objective is to reach a target on the shelf to pick an object, then move to another target above the box to place the object, and finally move back to the initial configuration.

Instead of considering the target in configuration space, we focus on Cartesian space, which presents unique challenges for optimization-based motion planning solvers. Reaching a target in configuration space typically yields a clear gradient to the solver, whereas reaching a Cartesian target poses a more difficult optimization problem due to the larger solution space, as the target may correspond to multiple configurations. A gradient-based solver will attempt to reach the target with the configuration that is closest to the initial configuration, particularly if initialized with a stationary trajectory at the initial configuration. However, if this solution is infeasible, it is challenging for the solver to find a different solution with a final configuration significantly different from the initial one, unless initialized well. One alternative approach is to use inverse kinematics (IK) to identify several possible final configurations, then use motion planning solvers to reach those configurations. However, it is not easy to select good configurations as the target, as it is difficult to determine whether a specific configuration is reachable from the initial configuration. Moreover, even when a solution is found, it may be highly suboptimal.

Our approach involves tackling both the IK problem and motion planning problem concurrently, with decision variables comprising the robot configuration(s) for the Cartesian target(s) and the joint angle trajectory needed to achieve those configurations. While optimizing both simultaneously can be challenging, our TTGO formulation enables us to obtain multiple solutions. To simplify the problem's complexity, we use motion primitives to represent the joint angle trajectory, as outlined in Appendix~\ref{appendix_primitives}. By utilizing our motion primitives formulation and given the initial and final configurations, we ensure that the movement always starts from the initial configuration and ends at the final configuration while complying with joint limitations.

Consider an $m$-DoF manipulator. The configuration of the manipulator can be represented using the joint angles $\bm{\theta}=(\theta_1,\ldots,\theta_{m}) \in \mb{R}^{m}$. We can assume that the domain of the joint angles is bounded by a rectangular domain $\Omega_{\bm{\theta}} = \times_{i=1}^{m} [\theta_{{\min}_i}, \theta_{{\max}_i}]$. We represent the trajectory evolution in terms of the phase of the motion, i.e., $t \in (0,1)$ with $t=1$ representing the end of the motion.

\subsection{Motion Primitives}
\label{appendix_primitives}
In our motion planning formulation, we generate motions using a basis function representation that satisfies the boundary conditions (with respect to phase/time) and the limits of the trajectory (the magnitude) while maintaining zero velocity at the boundary. Suppose we are given a choice of basis functions $\bm{\phi}= (\phi_k)^J_{j=1}, \phi_j(t) \in \mb{R}, \forall t \in [0,1]$. For example, we could use radial basis functions $\phi_j(t) = \exp(-\gamma (t-\mu_j)^2)$ with $\mu_j \in [0,1], \gamma \in \mb{R}^+ $. We define a trajectory using a weighted combination of these basis functions as $\hat{\tau}(t) = \sum_{j=1}^J w_j \phi_j(t)$. We transform this trajectory so that the boundary conditions and joint limits are satisfied.

Given the trajectory $\hat{\tau}(t), t\in [0,1]$,  and the boundary conditions $\tau(0)=\tau_0, \tau(1)=\tau_1$ and the limits $\tau_{\min} \le \tau(t) \le \tau_{\max}$, we can transform $\hat{\tau}(t)$ to obtain a trajectory $\tau(t) = \Psi(\hat{\tau}(t),\tau_0,\tau_1, \tau_{\min},\tau_{\max})$ such that $\tau(0) = \tau_0$, $\tau(1)=\tau_1$ and $\tau_{\min} \le \tau(t) \le \tau_{\max}$.  We define the transformation $\Psi$ as follows:
\begin{enumerate}
    \item Input: $\hat{\tau}, \tau_0, \tau_1, \tau_{\min}, \tau_{\max}$
    \item Discretize the time interval $[0,1]$ uniformly to obtain $\{t_i\}_{i=0}^N$ so that $dt = t_{i+1}-t_i$, $t\in \{t_i\}_{i=0}^N$.
    \item Define $\hat{z}(t) = \hat{\tau}(t) + \tau_0 -\hat{\tau}(0) + t (\tau_1-\tau_0 + \hat{\tau}(0)-\hat{\tau}(1)) $, which satisfies the specified boundary conditions.
    \item Clip the trajectory within the joint limits to obtain $z(t) = \text{clip}(\hat{z}(t), \tau_{\min}, \tau_{\max})$. The clipping will result in non-smoothness.
    \item Smoothen the trajectory $z(t)$ to obtain the desired trajectory $\tau(t)$: To do this, we append the trajectory $z(t)$ with the same values as initial value in the beginning  and with the final value at the end. Then we can apply a moving average filter over the trajectory. This creates the desired smooth trajectory $\tau(t)$ that has zero velocity at the boundary.

\end{enumerate}

\begin{figure}[ht]
    \centering
     \includegraphics[width=0.95\linewidth]{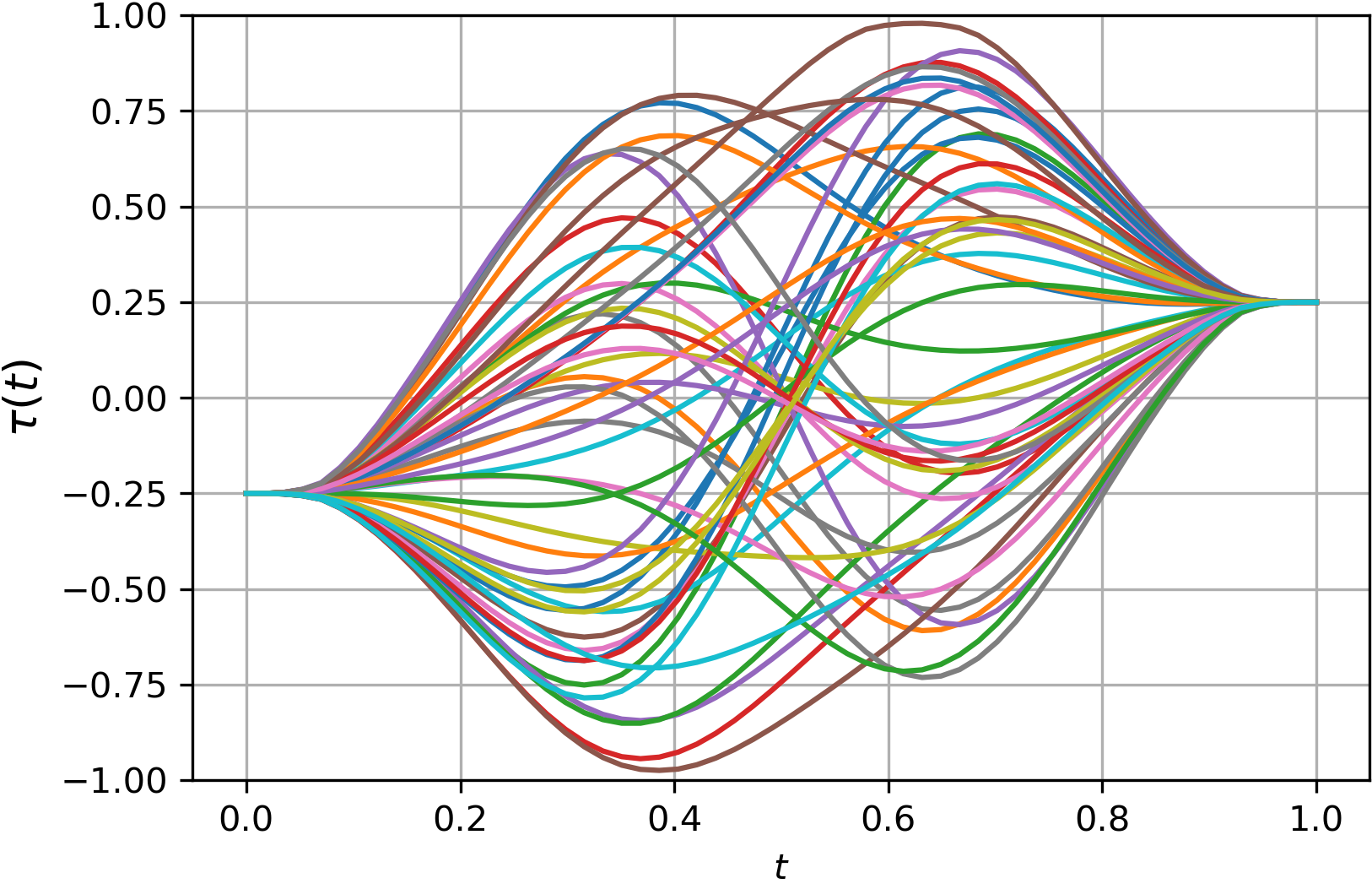}
     \caption{A distribution of $50$ smooth trajectories generated by transforming trajectories generated by using two radial basis functions with weights chosen uniformly in the range $[-1,1]$. The transformations are done to maintain a boundary condition $\tau_0 = -0.25, \tau_1=0.25$ and the limits $\tau_{\min}=-1, \tau_{\max}=1$.}
     \label{fig:mp}
\end{figure}

This way we can generate smooth motion while satisfying the boundary conditions and the joint limits, and maintain zero velocity at the boundary (see Figure \ref{fig:mp}).

\subsection{Computation Time}
\label{comp_time}
The computation time of TTGO can be divided into \emph{offline computation}, i.e., the time to construct the TT model $\bm{\mc{P}}$, and \emph{online computation}, i.e., the time to condition the TT model on the given task parameters and to sample. 

The offline training uses an NVIDIA GEFORCE RTX 3090 GPU with 24GB memory, while the sampling time evaluation is performed on an AMD Ryzen 7 4800U laptop.

The offline computation time depends on the number of TT-Cross iterations, the maximum rank $r$, and the discretization (i.e., how many elements along each dimension of the tensor). The number of function evaluations has $\mc{O}(ndr^2)$ complexity hence linear in terms of the number of dimensions and the number of discretization points. The computation time of a single cost function also has a significant influence on the TTGO computation time. However, we used parallel implementation with GPU that allows us to construct all of the models in our applications in an unsupervised manner (using TT-Cross) in less than one hour. 

The rank $r$ and the number of iterations of TT-Cross also determine the variety in the  solutions proposed by TTGO. If the application does not demand multiple solutions, we can keep the maximum allowable rank of the TT model and the number of iterations of TT-Cross to be very low, which results in a significant saving in offline computation time and the sampling time in the online phase. However, for the experiments in this article, we kept the rank $r$ to be reasonably large (about $r=60$ for IK and motion planning problems with manipulators) so that we could obtain a variety of solutions from TTGO.

For most of the 2D benchmark functions, it takes less than $0.01$s to obtain the TT model. For the high-dimensional mixture of Gaussians and Rosenbrock functions with $d<30$, we could obtain good enough TT-models in less than $60$s. It takes about $30$s for the inverse kinematics problem with the Franka Emika robot, which corresponds to $30$ iterations of TT-Cross. Finally, the target reaching task takes around $10$ minutes while the pick-and-place task takes around $1$ hour. The motion planning computation time is relatively slower due to the time for computing a single cost function since we compute the obstacle cost at small time intervals. It can be made faster by considering the continuous collision cost as done in TrajOpt~\citep{schulman2014motion}, since it allows us to use coarser time discretization for evaluating the collision cost, resulting in a faster evaluation of the cost function.

For the online computation time, the conditioning time is insignificant as it is very fast, so we focus on the sampling time. Unlike the TT model construction, the sampling time does not depend on the cost function and only depends on the size of the tensor. The computation complexity is $\mc{O}(ndr^2)$. Results of sampling time evaluation with the different number of samples averaged over 100 tests are given in Figure~\ref{fig:sampling_times}. We show the results for $d=7$ and $d=70$, roughly corresponding to the IK and the pick-and-place task, respectively. We can see that due to the parallel implementation, generating 1000 samples is not much different compared to generating 1 sample. For the IK problem, generating 1 sample takes around 1-3ms, which is comparable to the solving time of a standard IK solver. For the pick-and-place task, generating 1 sample takes around 15ms, much faster than a typical computation time for motion planning (typically in the order of 1s).

\end{document}